\documentclass[english,twocolumn,amsmath,amssymb,nofootinbib,superscriptaddress,longbibliography]{revtex4-2}
\usepackage[T1]{fontenc}
\usepackage{babel}
\usepackage{array}
\usepackage{calc}
\usepackage{amsmath}
\usepackage{graphicx}
\usepackage[unicode=true,pdfusetitle,
 bookmarks=true,bookmarksnumbered=false,bookmarksopen=false,
 breaklinks=false,pdfborder={0 0 0},pdfborderstyle={},backref=false,colorlinks=false]
 {hyperref}

\makeatletter

\providecommand{\tabularnewline}{\\}

\makeatother

\begin{document}
\global\long\def\tr{\mathrm{tr}\,}%

\title{Coding schemes in neural networks learning classification tasks}
\author{Alexander van Meegen}
\email{avanmeegen@fas.harvard.edu}

\address{Center for Brain Science, Harvard University, Cambridge, MA 02138}
\author{Haim Sompolinsky}
\email{hsompolinsky@mcb.harvard.edu}

\address{Center for Brain Science, Harvard University, Cambridge, MA 02138}
\address{Edmond and Lily Safra Center for Brain Sciences, Hebrew University,
Jerusalem 9190501, Israel}
\begin{abstract}
Neural networks posses the crucial ability to generate meaningful
representations of task-dependent features. Indeed, with appropriate
scaling, supervised learning in neural networks can result in strong,
task-dependent feature learning. However, the nature of the emergent
representations, which we call the `coding scheme', is still unclear.
To understand the emergent coding scheme, we investigate fully-connected,
wide neural networks learning classification tasks using the Bayesian
framework where learning shapes the posterior distribution of the
network weights. Consistent with previous findings, our analysis of
the feature learning regime (also known as `non-lazy', `rich', or
`mean-field' regime) shows that the networks acquire strong, data-dependent
features. Surprisingly, the nature of the internal representations
depends crucially on the neuronal nonlinearity. In linear networks,
an analog coding scheme of the task emerges. Despite the strong representations,
the mean predictor is identical to the lazy case. In nonlinear networks,
spontaneous symmetry breaking leads to either redundant or sparse
coding schemes. Our findings highlight how network properties such
as scaling of weights and neuronal nonlinearity can profoundly influence
the emergent representations.
\end{abstract}
\maketitle

\section{Introduction}

The remarkable empirical success of deep learning stands in strong
contrast to the theoretical understanding of trained neural networks.
Although every single detail of a neural network is accessible and
the task is known, it is still very much an open question how the
neurons in the network manage to collaboratively solve the task. While
deep learning provides exciting new perspectives on this problem,
it is also at the heart of more than a century of research in neuroscience.

Two key aspects are representation learning and generalization. It
is widely appreciated that neural networks are able to learn useful
representations from training data \citep{10.1109/TPAMI.2013.50,Lecun2015deep,Goodfellow-et-al-2016}.
But from a theoretical point of view, fundamental questions about
representation learning remain wide open: Which features are extracted
from the data? And how are those features represented by the neurons
in the network? Furthermore, neural networks are able to generalize
even if they are deeply in the overparameterized regime \citep{zhang2017understanding,10.1145/3446776,doi:10.1073/pnas.1903070116,Nakkiran2020Deep,Belkin_2021,shwartzziv2024just}
where the weight space contains a subspace---the solution space---within
which the network perfectly fits the training data. This raises a
fundamental question about the effect of implicit and explicit regularization:
Which regularization biases the solution space towards weights that
generalize well?

We here investigate the properties of the solution space using
the Bayesian framework where the posterior distribution of the weights
determines how the solution space is sampled \citep{Mackay2003,Bahri20_501}.
For theoretical investigations of this weight posterior, the size
of the network is taken to infinity. Crucially, the scaling of the
network and task parameters, as the network size is taken to infinity,
has a profound impact: neural networks can operate in different regimes
depending on this scaling. Here, the relevant scales are the width
of the network $N$ and the size of the training data set $P$. A
widely used scaling limit is the infinite width limit where $N$ is
taken to infinity while $P$ remains finite \citep{Neal96,Williams96_ae5e3ce4,Lee18,Matthews18,NEURIPS2019_5e69fda3,PhysRevE.104.064301,Segadlo_2022,pmlr-v162-hron22a}.
A scaling limit closer to empirically trained networks takes both
$N$ and $P$ to infinity at fixed ratio $\alpha=P/N$ \citep{Li21_031059,Naveh21_NeurIPS,ZavatoneVeth22_064118,pmlr-v202-cui23b,doi:10.1073/pnas.2301345120,Pacelli2023,fischer2024critical}.

In addition to scaling of $P$ and $N$, the scaling of of the final
network output with $N$ is important because it has a strong effect
on representation learning \citep{NEURIPS2019_ae614c55,pmlr-v125-woodworth20a,Geiger_2020}
(illustrated in Fig.~\ref{fig:overview}A). If the readout scales
its inputs with $1/\sqrt{N}$ the network operates in the `lazy' regime.
Lazy networks rely predominantly on random features and representation
learning is only a $1/N$ correction both for finite $P$ \citep{pmlr-v107-yaida20a,Dyer2020Asymptotics,ZavatoneVeth21_NeurIPS_II,Segadlo_2022,Roberts22}
and fixed $\alpha$ \citep{Li21_031059}; in the latter case it nonetheless
affects the predictor variance \citep{Li21_031059,doi:10.1073/pnas.2301345120,Pacelli2023}.
If the readout scales its inputs with $1/N$ the network operates
in the `non-lazy' (also called mean-field or rich) regime and learns
strong, task-dependent representations \citep{Mei18_E7665,NEURIPS2018_a1afc58c,NEURIPS2018_196f5641,SIRIGNANO20201820,pmlr-v139-yang21c,bordelon2022selfconsistent,bordelon2023dynamics}
(illustrated in Fig.~\ref{fig:overview}C). However, the nature of
the solution, in particular how the learned features are represented
by the neurons, remains unclear. Work based on a partial differential
equations for the weights \citep{Mei18_E7665,NEURIPS2018_a1afc58c,NEURIPS2018_196f5641,SIRIGNANO20201820}
only provides insight in effectively low dimensional settings; investigations
based on kernel methods \citep{pmlr-v139-yang21c,bordelon2022selfconsistent,bordelon2023dynamics}
average out important structures at the single neuron level.

In this paper, we develop a theory for the weight posterior of non-lazy
networks in the limit $N,P\to\infty$. We show that the representations
in non-lazy networks trained for classification tasks exhibit a remarkable
structure in the form of coding schemes, where distinct groups of
neurons are characterized by the subset of classes that activate them.
Another central result is that the nature of the learned representation
strongly depends on the type of neuronal nonlinearity (illustrated
in Fig.~\ref{fig:overview}D). Owing to the influence of the nonlinearity,
we consider three nonlinearities one after the other: linear, sigmoidal,
and ReLU leading to analog, redundant, and sparse coding schemes,
respectively. For each nonlinearity we first consider the learned
representations on training inputs in the simple setting of a toy
task and then investigate learned representations on training and
test inputs and generalization on MNIST and CIFAR10. The paper concludes
with a brief summary and discussion of our results.

\section{Results}

\subsection{Setup}

\begin{figure*}
\begin{centering}
\includegraphics[width=1.5\columnwidth]{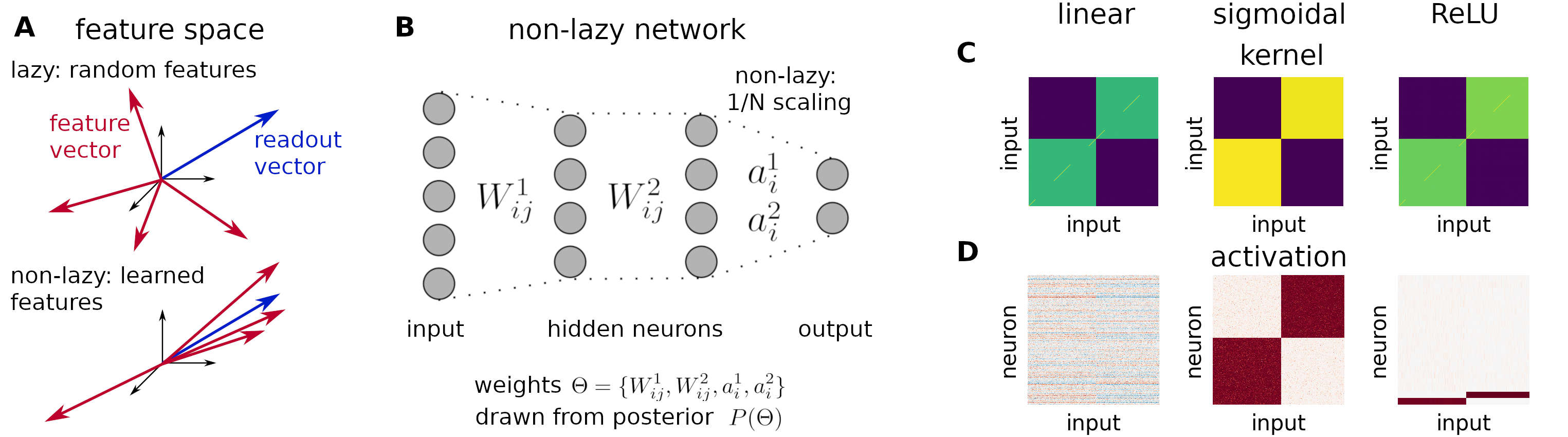}
\par\end{centering}
\caption{(\textbf{A}) Sketch of the last layer feature space in the lazy regime,
where the predominantly random features are almost orthogonal to the
readout weight vector ($O(\sqrt{N})$ overlap), and in the non-lazy
regime, where the learned features are aligned to the readout weight
vector ($O(N)$ overlap). (\textbf{B}) Fully connected network with
two hidden layers ($L=2$) and two outputs ($m=2$). Readouts are
scaled by $1/N$ instead of $1/\sqrt{N}$ in non-lazy networks after
training, i.e., under the posterior, which enforces the network to
learn strong representations. (\textbf{C}) Posterior-averaged kernels
of non-lazy networks with a single hidden layer for random binary
classification of orthogonal data. Nonlinearity from left to right:
linear, sigmoidal, and ReLU. (\textbf{D}) Activations for a given
weight sample from the networks shown in (C); for ReLU only the $20$
most active neurons are shown. Parameters: $N=P=500$, $N_{0}=510$,
classes assigned with probability $1/2$, targets $y_{+}=1$ and $y_{-}=-1$.
\label{fig:overview}}
\end{figure*}

We denote the output of a fully-connected feedforward networks with
$L$ hidden layers and $m$ outputs (Fig.~\ref{fig:overview}B) by
\begin{align}
f_{r}(x;\Theta) & =\frac{1}{N}\sum_{i=1}^{N}a_{i}^{r}\phi[z_{i}^{L}(x)],\quad r=1,\dots,m,\label{eq:network}
\end{align}
where $\phi(z)$ denotes the neuronal nonlinearity, $z_{i}^{L}(x)$
the last layer preactivation, and $x$ is an arbitrary $N_{0}$-dimensional
input. The preactivations are $z_{i}^{\ell}(x)=\frac{1}{\sqrt{N}}\sum_{j=1}^{N}W_{ij}^{\ell}\phi[z_{j}^{\ell-1}(x)]$
in the hidden layers $\ell=2,\dots,L$ and $z_{i}^{1}(x)=\frac{1}{\sqrt{N_{0}}}\sum_{j=1}^{N_{0}}W_{ij}^{1}x_{j}$
in the first layer. The activations are $\phi[z_{i}^{\ell}(x)]$ and
we assume for simplicity that the width of all hidden layers is $N$.
Importantly, we scale the output $f_{r}(x;\Theta)$ with $1/N$ such
that the network is in the non-lazy regime but the hidden layer preactivations
$z_{i}(x)$ with $1/\sqrt{N}$ or $1/\sqrt{N_{0}}$ (scaling $f_{r}(x;\Theta)$
with $1/\sqrt{N}$ would instead corresponds to the lazy regime).

The trainable parameters of the networks are the readout and hidden
weights, which we collectively denote by $\Theta$. The networks are
trained using empirical training data $\mathcal{D}=\{(x_{\mu},y_{\mu})\}_{\mu=1}^{P}$
containing $P$ inputs $x_{\mu}$ of dimension $N_{0}$ and $P$ targets
$y$ of dimension $m$. The $N_{0}\times P$ matrix containing all
training inputs is denoted by $X$; the $P\times m$ matrix containing
all training targets is denoted by $Y$.

\subsubsection{Weight Posterior}

The weight posterior is \citep{Li21_031059}
\begin{equation}
P(\Theta)=\frac{1}{Z}\exp\big[-\beta\mathcal{L}(\Theta)+\log P_{0}(\Theta)\big]\label{eq:posterior}
\end{equation}
where $Z=\int dP_{0}(\Theta)\,\exp[-\beta\mathcal{L}(\Theta)]$ is
the partition function, $\mathcal{L}(\Theta)=\frac{1}{2}\sum_{r=1}^{m}\sum_{\mu=1}^{P}[y_{\mu}^{r}-f_{r}(x_{\mu};\Theta)]^{2}$
a mean-squared error loss, and $P_{0}(\Theta)$ an i.i.d.~zero-mean
Gaussian prior with prior variances $\sigma_{a}^{2}$, $\sigma_{\ell}^{2}$
for readout and hidden weights, respectively. Temperature $T=1/\beta$
controls the relative importance of the quadratic loss and the prior.
In the overparameterized regime, the limit $T\to0$ restricts the
posterior to the solution space where the network interpolates the
training data. For simplicity, we set $\sigma_{\ell}=1$ in the main
text. Throughout, we denote expectations w.r.t.~the weight posterior
by $\langle\cdot\rangle_{\Theta}$.

The interplay between loss and prior shapes the solutions found by
the networks. Importantly, on the solution space the prior alone determines
the posterior probability. Thus, the prior plays a central role for
regularization which complements the implicit regularization due to
the non-lazy scaling.

\textbf{Scaling limit:} We consider the limit $N,N_{0},P\to\infty$
while the number of readout units as well as the number of layers
remain finite, i.e., $m,L=O(1)$. Because the gradient of $\mathcal{L}(\Theta)$
is small due to the non-lazy scaling of the readout, the noise introduced
by the temperature needs to be scaled down as well, requiring the
scaling $T\rightarrow T/N$ (see supplement). In the remainder of
the manuscript $T$ denotes the rescaled temperature which therefore
remains of $O(1)$ as $N$ increases. In the present work, we focus
on the regime where the training loss is essentially zero, and therefore
consider the limit $T\to0$ for the theory; to generate empirical
samples from the weight posterior we use a small but non-vanishing
rescaled temperature.

Under the posterior, the non-lazy scaling leads to large readout weights:
the posterior norm per neuron grows with $P$, thereby compensating
for the non-lazy scaling. To avoid this undoing of the non-lazy scaling,
we scale down $\sigma_{a}^{2}$ with $P$, guaranteeing that the posterior
norm of the readout weights per neuron is $O(1)$. The precise scaling
depends on the depth and the type of nonlinearity (see supplement).

In total there are three differences between (\ref{eq:posterior})
and the corresponding posterior in the lazy regime \citep{Li21_031059}:
1) The non-lazy scaling of the readout in (\ref{eq:network}) with
$1/N$ instead of $1/\sqrt{N}$. 2) The scaling of the temperature
with $1/N$. 3) The prior variance of the readouts weights $\sigma_{a}^{2}$
needs to be scaled down with $P$.

Due to the symmetry of fully connected networks under permutations
of the neurons, the statistics of the posterior weights must be invariant
under permutations. However, for large $N$ and $P$ this symmetry
may be broken at low $T$ so that the posterior consists of permutation
broken branches which are disconnected from each other in weight space.
A major finding of our work is the emergence of this spontaneous symmetry
breaking and its far reaching consequences for the nature of the learned
neural representations.

\subsubsection{Kernel and Coding Schemes}

Learned representations are frequently investigated using the kernel
\citep[e.g.,][]{JMLR:v13:cortes12a,pmlr-v97-kornblith19a,Li21_031059,bordelon2022selfconsistent}
which measures the overlap between the neurons' activations on pairs
of inputs: 
\begin{equation}
K_{\ell}(x_{1},x_{2};\Theta)=\frac{1}{N}\sum_{i=1}^{N}\phi[z_{i}^{\ell}(x_{1})]\phi[z_{i}^{\ell}(x_{2})]\label{eq:def_kernel}
\end{equation}
where the inputs $x_{1}$, $x_{2}$ can be either from the training
or the test set.  We denote the posterior averaged kernel by $K_{\ell}(x_{1},x_{2})=\langle K_{\ell}(x_{1},x_{2};\Theta)\rangle_{\Theta}$;
the posterior averaged $P\times P$ kernel matrix on all training
inputs is denoted by $K_{\ell}$ without arguments. In addition to
capturing the learned representations, the kernel is a central observable
because it determines the statistics of the output function in wide
networks \citep{Neal96,Williams96_ae5e3ce4,Lee18,Matthews18}.

Crucially, the kernels disregards how the learned features are represented
by the neurons. Consider, for example, binary classification in two
extreme cases: (1) each class activates a single neuron and all remaining
neurons are inactive; (2) each class activates half of the neurons
and the remaining half of the neurons are inactive. In both scenarios
the kernels agree up to an overall scaling factor despite the drastic
difference in the underlying representations (illustrated in Fig.~\ref{fig:overview}(C,D)).

\begin{table}
\begin{centering}
\begin{tabular*}{1\columnwidth}{@{\extracolsep{\fill}}>{\centering}p{0.24\columnwidth}>{\raggedright}p{0.74\columnwidth}}
\noalign{\vskip1mm}
Term & Definition\tabularnewline
\hline 
\noalign{\vskip1mm}
code & The subset of classes that activate a neuron.\tabularnewline
\noalign{\vskip1mm}
coding scheme & The collection of codes implemented by a neuronal population.\tabularnewline
\noalign{\vskip1mm}
sparse coding & Only a small subset of neurons exhibit codes.\tabularnewline
\noalign{\vskip1mm}
redundant coding & All the codes in the scheme are shared by a large subset of neurons.\tabularnewline
\noalign{\vskip1mm}
analog coding & All neurons respond to all classes but with different strength.\tabularnewline
\end{tabular*}
\par\end{centering}
\caption{Glossary of neural coding terminology. \label{tab:glossary}}
\end{table}

The central goal of our work is to understand the nature of the representations.
As illustrated above, this means that the theory needs to go beyond
kernels. To this end, we use the notion of a neural code which we
define as the subset of classes that activate a given neuron (see
Table \ref{tab:glossary}). At the population level, this leads to
a ``coding scheme'': the collection of codes implemented by the
neurons in the population. As we will show the coding schemes encountered
in our theory are of the three types (illustrated in Fig.~\ref{fig:overview}(D)):
1) Sparse coding where only a small subset of neurons exhibit codes
(as in the first scenario above). 2) Redundant coding where all the
codes in the scheme are shared by a large subset of neurons (as in
the second scenario above). 3) Analog coding where all neurons respond
to all classes but with different strength.

As illustrated in Fig.~\ref{fig:overview}(D), the nature of the
learned representations varies crucially with the choice of the activation
function of the hidden layer, $\phi(z)$, although the kernel matrix
is similar in all cases. Specifically, in the linear case, $\phi(z)=z$,
the coding is analog, in the sigmoidal case, $\phi(z)=\frac{1}{2}[1+\mathrm{erf}\ensuremath{(\sqrt{\pi}z/2})])$,
the coding is redundant, and in the case of ReLU, $\phi(z)=\max(0,z)$,
the coding is sparse.

\subsection{Linear Networks}

To gain insight into the neural code and its relation to the nonlinearity
we develop a mean field theory that is exact in the above scaling
limit. We begin with linear networks $\phi(z)=z$ because the theory
is analytically solvable by adapting the techniques developed in \citep{Li21_031059}
to non-lazy networks. In order to prevent growth of the norm of the
readout weights with $P$ we set $\sigma_{a}^{2}=1/P^{1/L}$ (see
supplement). The theoretical results are in the zero temperature limit.

\begin{figure}
\includegraphics[width=1\columnwidth]{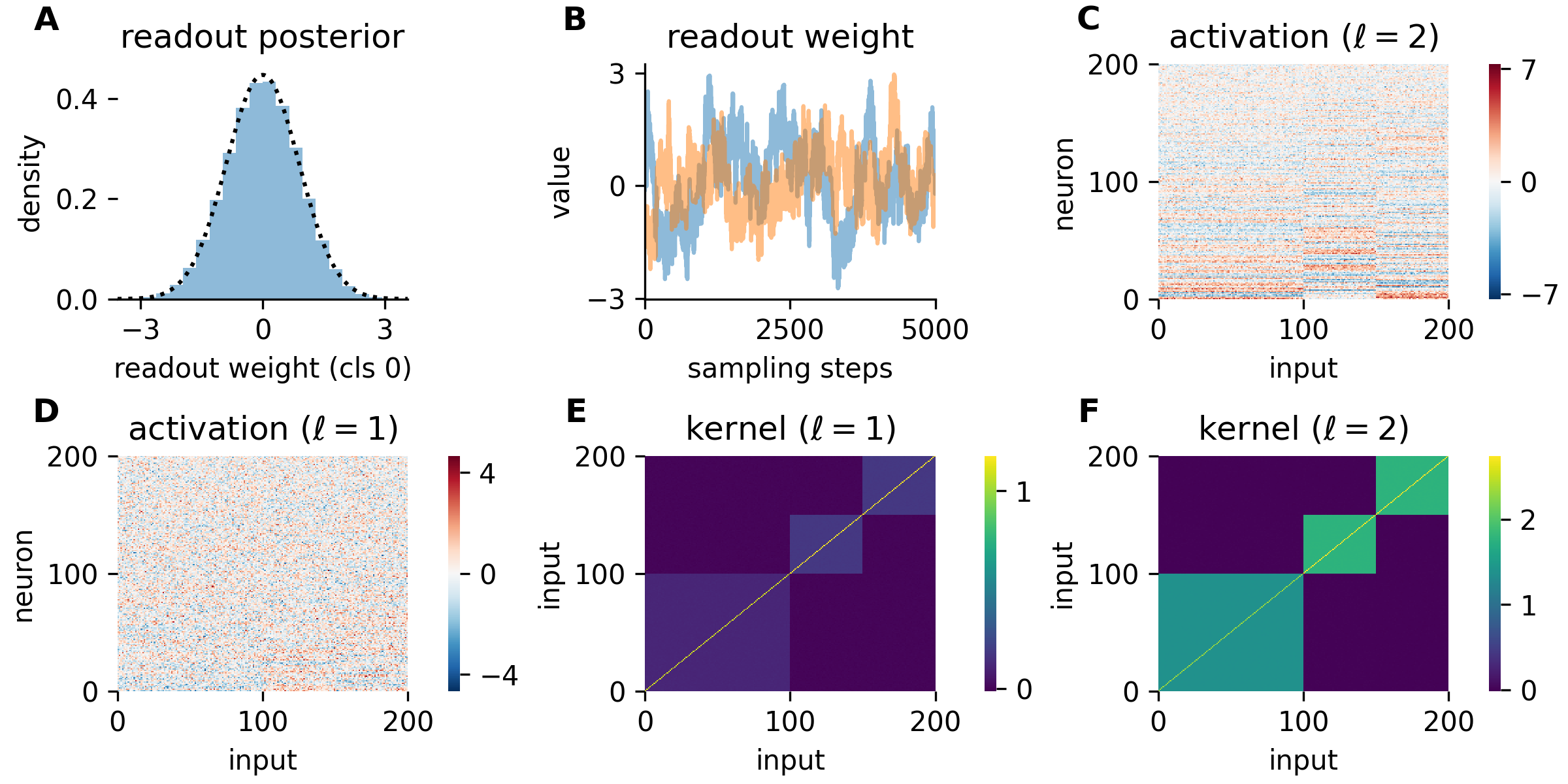}

\caption{Analog coding in two hidden layer linear networks on random classification
task. (\textbf{A}) Readout weight posterior on first class; theoretical
distribution (\ref{eq:a_posterior_linear}) as black dashed line.
(\textbf{B}) Sampled readout weights for first class and two arbitrary
neurons. (\textbf{C},\textbf{D}) Activations of all neurons on all
training inputs for a given weight sample in the second (C) and first
(D) hidden layer. Neurons sorted by mean squared activity; inputs
are sorted by class. (\textbf{E},\textbf{F}) Kernel on training data
from sampling in the first (E) and second (F) hidden layer. Parameters:
$N=P=200$, $N_{0}=220$, classes assigned with fixed ratios $[1/2,1/4,1/4]$,
targets $y_{+}=1$ and $y_{-}=0$. \label{fig:linear_code}}
\end{figure}

\subsubsection{Training}

We start with the learned representations on the training set. A
key result of the theory is that the joint posterior of the readout
weights and the activations factorizes across neurons and layers,
$\prod_{i=1}^{N}[P(a_{i})P(z_{i}^{L}\,|\,a_{i})]\prod_{\ell=1}^{L-1}[\prod_{j=1}^{N}P(z_{j}^{\ell})]$,
where $a_{i}$ denotes the $m$-dimensional vector containing the
readout weights for neuron $i$ and $z_{i}^{\ell}$ denotes the $P$-dimensional
vector containing the activations of neuron $i$ on all training inputs.

The single neuron readout posterior is Gaussian,
\begin{equation}
P(a_{i})=\mathcal{N}(a_{i}\,|\,0,U),\label{eq:a_posterior_linear}
\end{equation}
where $U$ is a $m\times m$ matrix determined by $U^{L+1}=\sigma_{a}^{2L}Y^{\top}K_{0}^{-1}Y$
where $K_{0}=\frac{1}{N_{0}}X^{\top}X$ is the input kernel (see supplement).
Gaussianity of the sampled readout weights is shown in Fig.~\ref{fig:linear_code}(A)
using a network with two hidden layers and a toy task (mutually orthogonal
inputs, randomly assigned classes, and one-hot targets $y_{\pm}$)
with three classes which are present in the data with ratios $[1/2,1/4,1/4]$.
Fig.~\ref{fig:linear_code}(B) shows that the full posterior is explored
during sampling.

The coding scheme is determined by the conditional single neuron
posterior of the last layer activations given the readout weights
which is also Gaussian, $P(z_{i}^{L}\,|\,a_{i})=\mathcal{N}(z_{i}^{L}\,|\,YU^{-1}a_{i},K_{L-1})$.
Its mean $YU^{-1}a_{i}$ codes for the training targets with a strength
proportional to the readout weights. Since the readout weights are
Gaussian, this leads to an analog coding of the task. A sample of
the analog coding scheme is shown in Fig.~\ref{fig:linear_code}(C)
for the two layer network on the toy task. 

For the previous layers $\ell<L$, the single neuron posterior is
independent of $a_{i}$ and the other layers' activations, and is
Gaussian: $P(z_{i}^{\ell})=\mathcal{N}(z_{i}^{\ell}\,|\,0,K_{\ell})$
for $\ell=1,\dots,L-1$. In all layers, the activation's covariance
is determined by the kernel
\begin{align}
K_{\ell} & =K_{0}+\sigma_{a}^{2(L-\ell)}YU^{-(L-\ell+1)}Y^{\top},\quad\ell=1,\dots,L.\label{eq:kernel_linear}
\end{align}
The kernel decomposes into a contribution $K_{0}$ due to the prior
and a learned low rank contribution $\sigma_{a}^{2(L-\ell)}YU^{-(L-\ell+1)}Y^{\top}$.
The strength of the learned part increases across layers, $\sigma_{a}^{2(L-\ell)}=1/P^{1-\frac{\ell}{L}}$.
In the last layer it is $O(1)$, indicating again that the last hidden
layer learns strong features. This is in strong contrast to the lazy
case where the learned part is suppressed by $1/N$ in all layers
\citep{Li21_031059}. We show a sample of the first layer activations
in Fig.~\ref{fig:linear_code}(D) for the two layer network and the
toy task. In contrast to the last layer activations, the learned structure
is hardly apparent in the first layer. This is also reflected in the
first layer kernel (Fig.~\ref{fig:linear_code}(E)) where the learned
block structure is weak compared to the last layer kernel (Fig.~\ref{fig:linear_code}(F)).

\begin{figure}
\includegraphics[width=1\columnwidth]{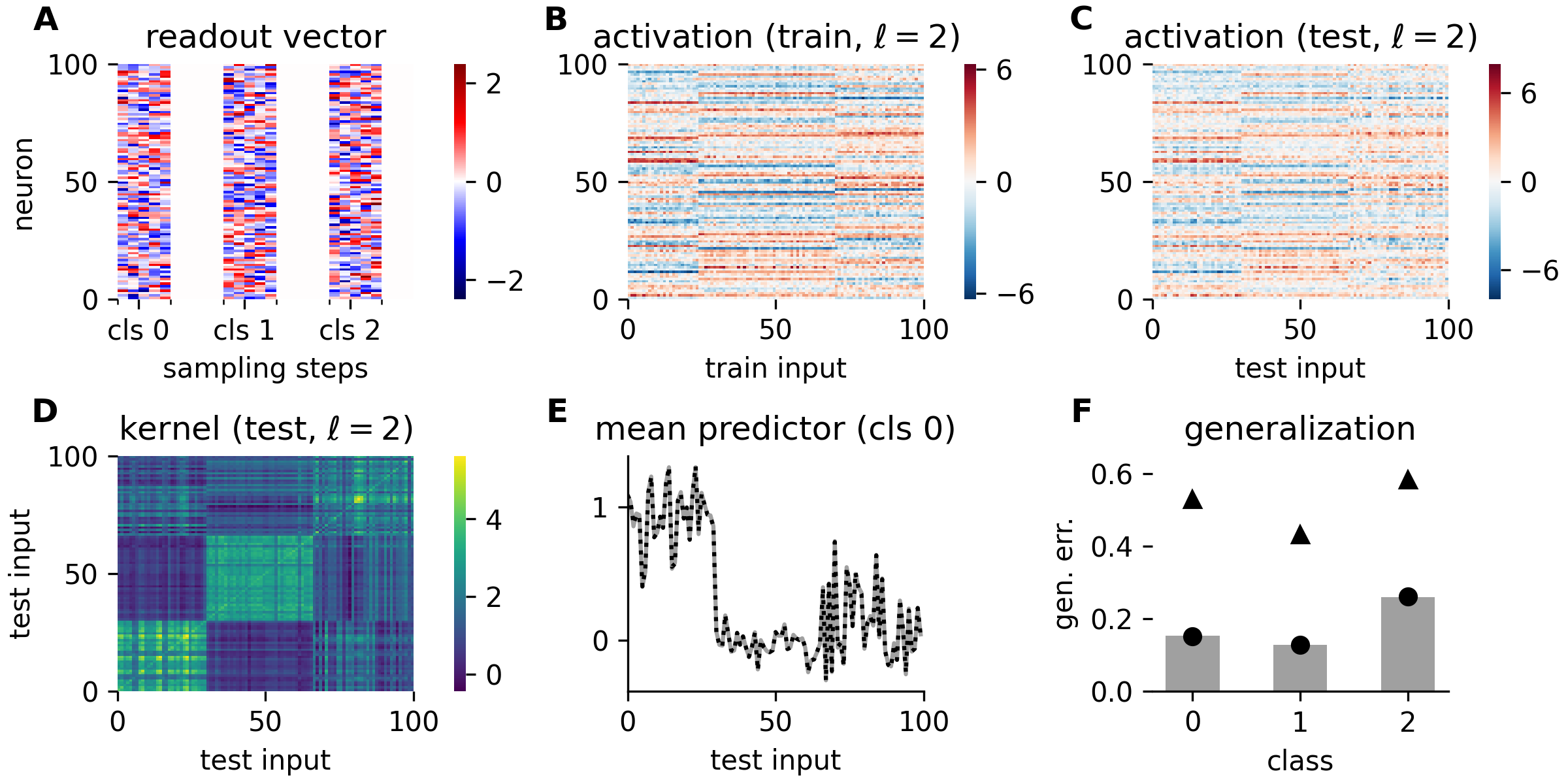}

\caption{Coding scheme and generalization of two hidden layer linear networks
on MNIST. (\textbf{A}) Samples of the readout weights of all three
classes. (\textbf{B},\textbf{C}) Activations of all neurons on all
training (B) and $100$ test (C) inputs for a given weight sample.
(\textbf{D}) Kernel on $100$ test inputs from sampling. (\textbf{E})
Mean predictor for class 0 from sampling (gray) and theory (eq.~(\ref{eq:predictor_linear}),
black dashed). (\textbf{F}) Generalization error for each class averaged
over $P_{*}=1,000$ test inputs from sampling (gray bars), theory
(Eq.~(\ref{eq:predictor_linear}), black circles), and GP theory
(back triangles). Parameters: $N=P=100$, $N_{0}=784$, classes 0,
1, 2 assigned randomly with probability $1/3$, targets $y_{+}=1$
and $y_{-}=0$. \label{fig:linear_mnist}}
\end{figure}

\subsubsection{Generalization}

For generalization, we include the $P_{*}$ dimensional vector of
test activations $z_{i}^{\ell}(x^{*})\equiv z_{i}^{\ell,*}$ on a
set of test inputs $x_{\mu}^{*}$, $\mu=1,\dots,P_{*}$, in the joint
posterior. The joint posterior still factorizes across neurons and
layers into single neuron posteriors $P(z_{i}^{L,*}\,|\,z_{i}^{L})P(z_{i}^{L}\,|\,a_{i})P(a_{i})$
in the last layer and $P(z_{i}^{\ell,*}\,|\,z_{i}^{\ell})P(z_{i}^{\ell})$
in the previous layers $\ell<L$. Like the other single neuron posteriors,
$P(z_{i}^{\ell,*}\,|\,z_{i}^{\ell})$ is Gaussian for all layers $\ell=1,\dots,L$
(see supplement).

The kernel for a pair of arbitrary inputs $x_{1},x_{2}$ is $K_{\ell}(x_{1},x_{2})=\kappa_{0}(x_{1},x_{2})+\sigma_{a}^{2(L-\ell)}f(x_{1})^{\top}U^{-(L-\ell+1)}f(x_{2})$
with the input kernel $\kappa_{0}(x_{1},x_{2})=\frac{1}{N_{0}}x_{1}^{\top}x_{2}$
and the mean predictor $f(x)=\langle f(x;\Theta)\rangle_{\Theta}$
which evaluates to
\begin{equation}
f(x)=Y^{\top}K_{0}^{-1}k_{0}(x)\label{eq:predictor_linear}
\end{equation}
where $k_{0}(x)=\frac{1}{N_{0}}X^{\top}x$ and $K_{0}=\frac{1}{N_{0}}X^{\top}X$.
The test kernel is identical to the training kernel (\ref{eq:kernel_linear})
except that the training targets are replaced by the predictor. As
in the training kernel, the learned part is low rank and becomes more
prominent across layers, reaching $O(1)$ in the last layer. Despite
the strong learned representations, the mean predictor is identical
to the lazy case \citep{Li21_031059} and the GP limit (see supplement).
In contrast to the lazy case, the variance of the predictor can be
neglected (see supplement).

We apply the theory to classification of the first three digits of
MNIST. The readout weights are Gaussian and change during sampling
(Fig.~\ref{fig:linear_mnist}(A)) and the last layer training activations
display a clear analog coding of the task (Fig.~\ref{fig:linear_mnist}(B)).
Thus, the nature of the solution is captured by an analog coding scheme,
as in the toy task. On test inputs, the last layer activations still
display the analog coding scheme (Fig.~\ref{fig:linear_mnist}(C)),
which shows that the learned representations generalize. This is confirmed
by the test kernel which has a block structure corresponding to the
task (Fig.~\ref{fig:linear_mnist}(D)). Also the mean predictor generalizes
well (Fig.~\ref{fig:linear_mnist}(E)) and the class-wise generalization
error $\varepsilon_{r}=\frac{1}{P_{*}}\sum_{\mu=1}^{P_{*}}\langle[y_{\mu}^{r,*}-f_{r}(x_{\mu}^{*};\Theta)]^{2}\rangle_{\Theta}$
is reduced compared to the GP theory (Fig.~\ref{fig:linear_mnist}(F)).
Since the mean predictors are identical, the reduced generalization
error is exclusively an effect of the reduced variance of the predictor
$\langle\delta f_{r}(x;\Theta)^{2}\rangle_{\Theta}$ due to the bias
variance decomposition $\langle[y^{r}-f_{r}(x;\Theta)]^{2}\rangle_{\Theta}=[y^{r}-f_{r}(x)]^{2}+\langle\delta f_{r}(x;\Theta)^{2}\rangle_{\Theta}$.

\subsection{Sigmoidal Networks}

For the first nonlinear case, we consider nonnegative sigmoidal networks,
$\phi(z)=\frac{1}{2}[1+\mathrm{erf}\ensuremath{(\sqrt{\pi}z/2})]$.
To avoid growth of the readout weights with $P$ we set $\sigma_{a}^{2}=1/P$
for arbitrary depth.

\begin{figure}
\includegraphics[width=1\columnwidth]{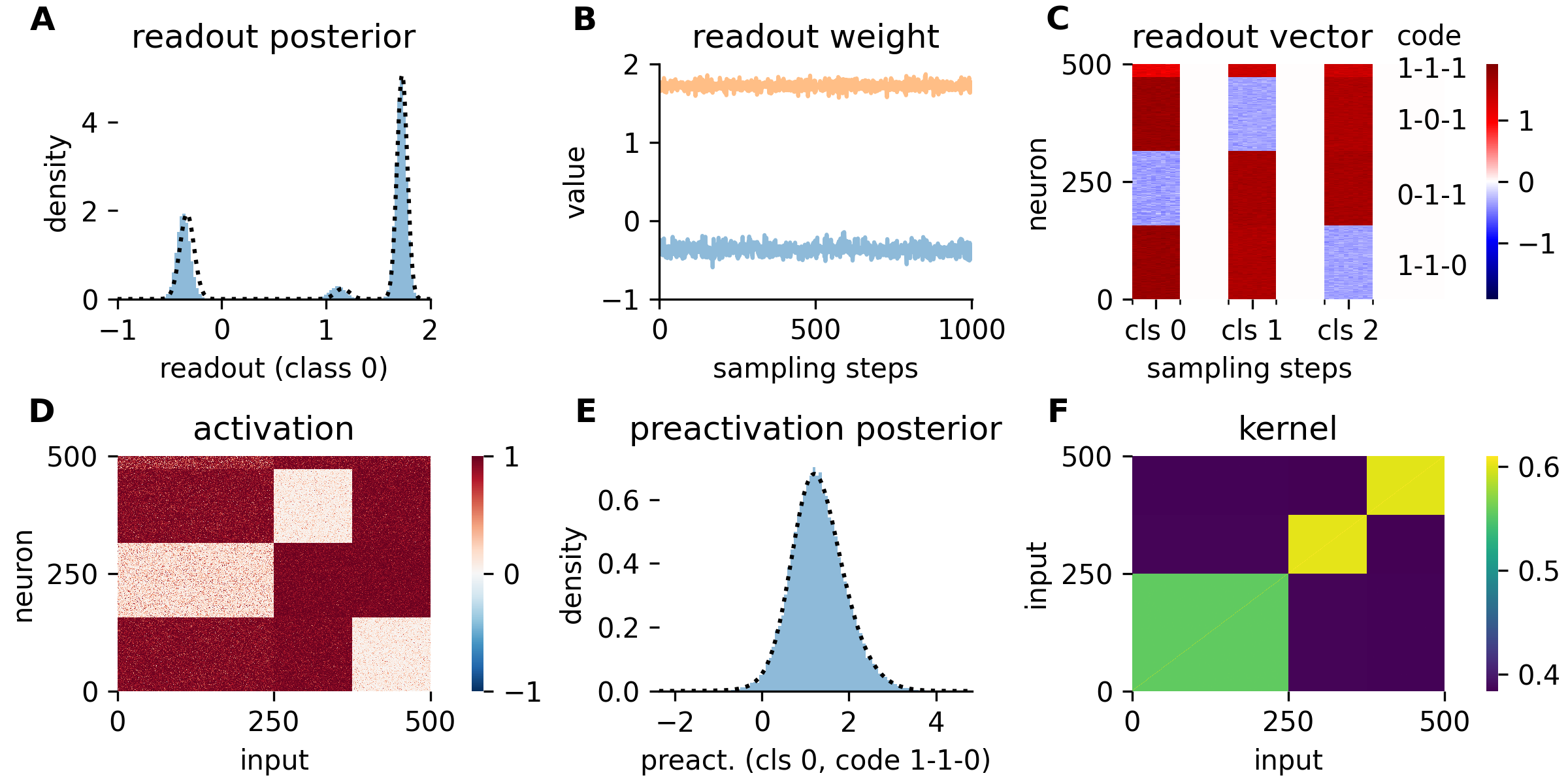}

\caption{Redundant coding in single layer sigmoidal networks on random classification
task. (\textbf{A}) Distribution of readout weights for first class
across neurons and samples (blue); theoretical distribution (black
dashed) includes finite $P$ corrections (see supplement). (\textbf{B})
Sampled readout weights for first class and two arbitrary neurons.
(\textbf{C}) Samples of the readout weights of all three classes.
(\textbf{D}) Activations of all neurons on all training inputs for
a given weight sample. (\textbf{E}) Distribution of training activations
of neurons with 1-1-0 code on inputs from class 0 across neurons for
a given weight sample (blue) and according to theory (black dashed).
(\textbf{F}) Kernel on training data from sampling. Parameters: $N=P=500$,
$N_{0}=520$, classes assigned with fixed ratios $[1/2,1/4,1/4]$,
targets $y_{+}=1$ and $y_{-}=1/2$. \label{fig:sigmoidal_code}}
\end{figure}

\subsubsection{Training}

As in the linear case, the joint posterior of readout weights and
preactivations factorizes across layers and neurons, $\prod_{i=1}^{N}[P(a_{i})P(z_{i}^{L}\,|\,a_{i})]\prod_{\ell=1}^{L-1}[\prod_{j=1}^{N}P(z_{j}^{\ell})]$.
In contrast to the linear case, the single neuron posteriors are no
longer Gaussian.

In particular the readout posterior is drastically different: it
is a weighted sum of Dirac deltas
\begin{align}
P(a_{i}) & =\sum_{\gamma=1}^{n}P_{\gamma}\delta(a_{i}-a_{\gamma})\label{eq:a_posterior_sigmoidal}
\end{align}
where the weights $P_{\gamma}$ and the $m$-dimensional vectors $a_{\gamma}$
are determined jointly for all $\gamma=1,\dots,n$ by a set of self-consistency
equations which depend on the training set (see supplement). This
means that the vector of readout weights is redundant: each of the
$N$ neurons has one of $n$ possible readout weights $a_{\gamma}$
and the distribution of readout weights across neurons is given by
$P_{\gamma}$.

We show the readout posterior for a single layer network on the same
toy task with three classes used above in Fig.~\ref{fig:sigmoidal_code}.
Across neurons, the posterior splits into disconnected branches which
are accurately captured by the theory (Fig.~\ref{fig:sigmoidal_code}(A);
here the marginal distribution of the readout weights for the first
class is shown). For given neurons, sampling is restricted to one
of the branches of the posterior (Fig.~\ref{fig:sigmoidal_code}(B)).
Considering all readouts and all neurons simultaneously, the redundant
structure of the readout weights becomes apparent (Fig.~\ref{fig:sigmoidal_code}(C)).

The readout weights $a_{i}$ determine the last layer preactivation
$z_{i}^{L}$ through the single neuron conditional distribution $P(z_{i}^{L}\,|\,a_{i})$.
Due to the redundant structure of the readout posterior, a redundant
coding scheme emerges in which, on average, a fraction $P_{\gamma}$
of the neurons exhibit identical preactivation posteriors $P(z_{i}^{L}\,|\,a_{\gamma})$.
Each of the corresponding activations has a pronounced mean that reflects
the structure of the task but also significant fluctuations around
the mean which do not vanish in the limit of large $N$ and $P$ (see
supplement).

For the single layer network on the toy task, the redundant coding
of the task is immediately apparent (Fig.~\ref{fig:sigmoidal_code}(D))
but also the remaining fluctuations are clearly visible (Fig.~\ref{fig:sigmoidal_code}(E)).
In this example, there are four codes (Fig.~\ref{fig:sigmoidal_code}(C,D)):
1-1-0, 0-1-1, 1-0-1, and 1-1-1. Note that the presence of code 1-1-1,
which is implemented only by a small fraction of neurons, is accurately
captured by the theory (Fig.~\ref{fig:sigmoidal_code}(A) middle
peak). The relation between the neurons' activations and their readout
weights is straightforward: for neurons with code 1-1-0, the readout
weights are positive for classes 0 and 1 and negative for class 2
(Fig.~\ref{fig:sigmoidal_code}(C)), and vice versa for the other
codes. Due to the symmetry of the last two classes, which account
for $1/4$ of the data each, the readout weights for codes 1-0-1 and
1-1-0 are identical such that in Fig.~\ref{fig:sigmoidal_code}(A)
only three peaks are visible but the height of the last peak is doubled.
Clearly, the coding scheme with four codes is not the unique solution.
Indeed, the theory admits other coding schemes which are, however,
unlikely to occur (see supplement).

\begin{figure}
\includegraphics[width=1\columnwidth]{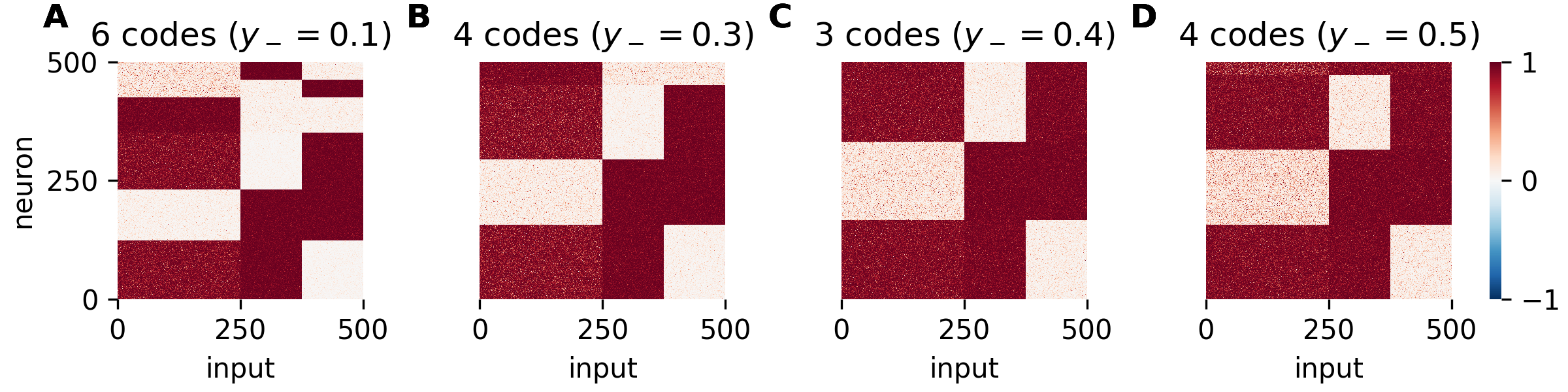}

\caption{Coding scheme transitions in single layer sigmoidal networks on random
classification task. (\textbf{A}-\textbf{D}) Activations of all neurons
on all training inputs for a given weight sample with changing training
target $y_{-}$. Parameters as in Fig.~\ref{fig:sigmoidal_code}
except that $y_{-}$ changes as indicated in the figure. \label{fig:sigmoidal_code_changes}}
\end{figure}

Interestingly, there are transitions in the coding scheme depending
on the targets $y_{\mu}^{r}$. In Fig.~\ref{fig:sigmoidal_code_changes}
we show the emerging code using the same setup as in Fig.~\ref{fig:sigmoidal_code}
except that the one-hot encoding of the targets $y_{-}$ increases
from $y_{-}=0.1$ to $y_{-}=0.5$ (which was used in Fig.~\ref{fig:sigmoidal_code}),
keeping $y_{+}=1$ as before. At $y_{-}=0.1$, the coding scheme contains
all six permutations of the 1-0-0 and 1-1-0 codes (Fig.~\ref{fig:sigmoidal_code_changes}(A)).
At $y_{-}=0.3$ the codes 0-1-0 and 0-0-1 disappeared (Fig.~\ref{fig:sigmoidal_code_changes}(B))
and at $y_{-}=0.4$ also the code 1-0-0 is lost (Fig.~\ref{fig:sigmoidal_code_changes}(C)).
Finally, at $y_{-}=0.5$ the new code 1-1-1 emerged (Fig.~\ref{fig:sigmoidal_code_changes}(D)).
Keeping $y_{-}=0.5$ but increasing temperature gradually decreases
the strength of the coding scheme until it disappears above a critical
temperature (see supplement).

The last layer kernel on training inputs is $K_{L}=\sum_{\gamma=1}^{n}P_{\gamma}\langle\phi(z^{L})\phi(z^{L})^{\top}\rangle_{z^{L}|a_{\gamma}}$.
Due to the pronounced mean of the activations on the toy task, the
kernel possesses a dominant low rank component. For the single layer
network and the toy task, this is shown in Fig.~\ref{fig:sigmoidal_code}(F).

A fundamental consequence of the theory is that neuron permutation
symmetry is broken in the last layer: due to the disconnected structure
of the readout posterior (\ref{eq:a_posterior_sigmoidal}), two readout
weights from different branches cannot change their identity during
sampling (Fig.~\ref{fig:sigmoidal_code}(B)). Furthermore, the readout
weights freeze to their value $a_{\gamma}$, in strong contrast to
the linear case (Fig.~\ref{fig:linear_code}(B)). Since the readout
weights determine the code through $P(z_{i}^{L}\,|\,a_{\gamma})$,
freezing of the readout weights implies freezing of the code, despite
the non vanishing fluctuations of preactivations (Fig.~\ref{fig:sigmoidal_code}(E)).

\begin{figure}
\includegraphics[width=1\columnwidth]{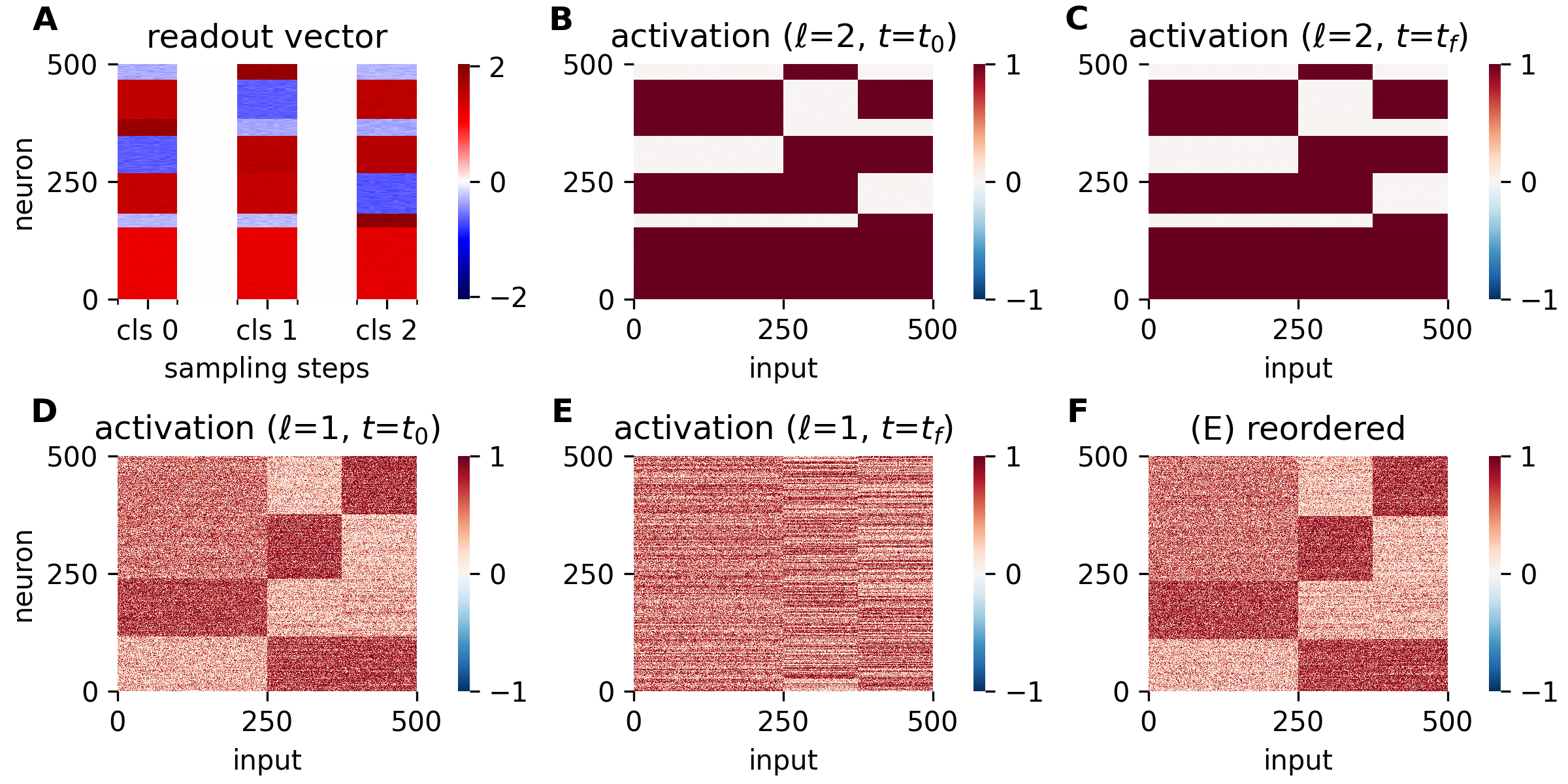}

\caption{Redundant coding in two layer sigmoidal networks on random classification
task. (\textbf{A}) Samples of the readout weights of all three classes.
(\textbf{B},\textbf{C}) Activations of all second layer neurons on
all training inputs using first (B) and last (C) weight sample. (\textbf{D},\textbf{E})
Activations of all first layer neurons on all training inputs using
first (D) and last (E) weight sample. (\textbf{F}) Activations from
(E) but neurons are reordered. Parameters: $N=P=500$, $N_{0}=520$,
classes assigned with fixed ratios $[1/2,1/4,1/4]$, targets $y_{+}=1$
and $y_{-}=1/2$. \label{fig:sigmoidal_code_multilayer}}
\end{figure}

For the lower layers $\ell<L$, the posterior $P(z_{i}^{\ell})$
is multimodal with each mode corresponding to a code (see supplement).
In contrast to the last layer, the code is not frozen since the preactivations
are independent from the readout weights. We show the activations
of a two layer network on the toy task in Fig.~\ref{fig:sigmoidal_code_multilayer}.
The readout weights are redundant and frozen (Fig.~\ref{fig:sigmoidal_code_multilayer}(A))
and the last layer activations show a prominent redundant coding of
the task (Fig.~\ref{fig:sigmoidal_code_multilayer}(B)) which does
not change during sampling (Fig.~\ref{fig:sigmoidal_code_multilayer}(C)).
In stark contrast, the first layer activations show a redundant coding
(Fig.~\ref{fig:sigmoidal_code_multilayer}(D)) which is not preserved
during sampling (Fig.~\ref{fig:sigmoidal_code_multilayer}(E)). This
is a fundamental difference to the last layer where the code is frozen.
However, the structure of the code is preserved: after reordering
the neurons the solution from the last weight sample is identical
to the solution from the first weight sample (Fig.~\ref{fig:sigmoidal_code_multilayer}(F)).

\begin{figure}
\includegraphics[width=1\columnwidth]{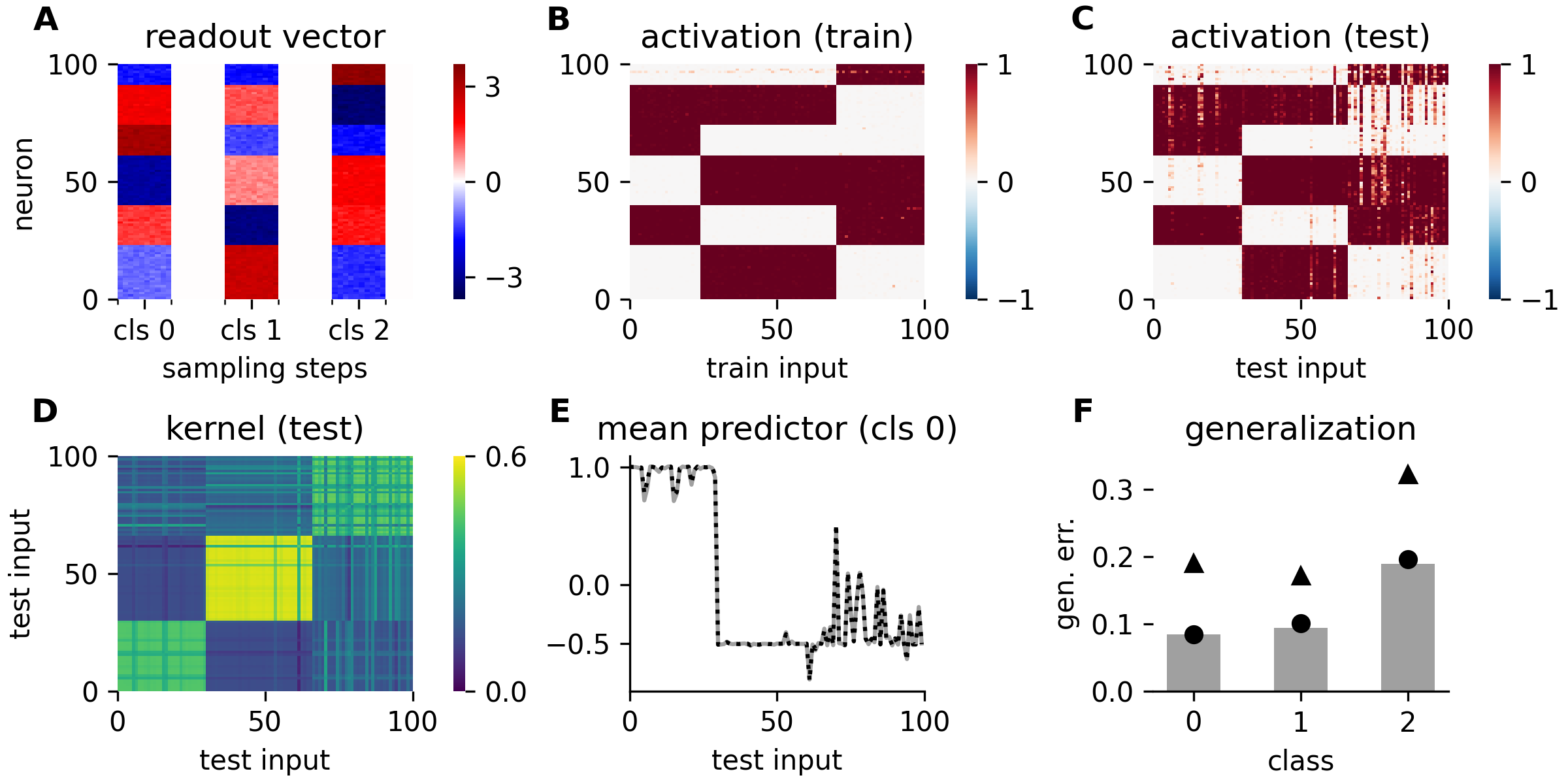}

\caption{Coding scheme and generalization of one hidden layer sigmoidal networks
on MNIST. (\textbf{A}) Samples of all three readout weights for all
neurons. (\textbf{B},\textbf{C}) Activations of all neurons on all
training (B) and $100$ test (C) inputs for a given weight sample.
(\textbf{D}) Kernel on $100$ test inputs from sampling. (\textbf{E})
Mean predictor for class 0 from sampling (gray) and theory (black
dashed) based on Eq.~(\ref{eq:predictor_sigmoidal}) neglecting the
preactivation fluctuations (see supplement). (\textbf{F}) Generalization
error for each class averaged over $1,000$ test inputs from sampling
(gray bars), theory (black circles), and GP theory (back triangles).
Parameters: $N=P=100$, $N_{0}=784$, classes 0, 1, 2 assigned randomly
with probability $1/3$, targets $y_{+}=1$ and $y_{-}=-1/2$. \label{fig:sigmoidal_mnist}}
\end{figure}

\subsubsection{Generalization}

For generalization, we include the preactivations on test inputs
$z_{i}^{\ell,*}$. As in the linear case, the posterior factorizes
across neurons and layers and the test preactivations only depend
on the training preactivations through $P(z_{i}^{\ell,*}\,|\,z_{i}^{\ell})$,
$\ell=1,\dots,L$, which is Gaussian also for nonlinear networks (see
supplement). The resulting mean predictor on a test input $x$ is
(see supplement)
\begin{align}
f_{r}(x) & =\sum_{\gamma=1}^{n}P_{\gamma}a_{\gamma}^{r}\langle\langle\phi[z^{L}(x)]\rangle_{z^{L}(x)|z^{L}}\rangle_{z^{L}|a_{\gamma}}\label{eq:predictor_sigmoidal}
\end{align}
where $P_{\gamma}$ and $a_{\gamma}$ are determined using the training
data set, see (\ref{eq:a_posterior_sigmoidal}).   As in the linear
case, the variance of the predictor can be neglected (see supplement).

We apply the theory to classification of the first three digits in
MNIST with a single hidden layer network. For the theory, we employ
the additional approximation of neglecting the fluctuations of the
preactivations conditioned on the readout weights. The readout weights
are redundant and frozen (Fig.~\ref{fig:sigmoidal_mnist}(A)) and
the training activations display a clear redundant coding scheme (Fig.~\ref{fig:sigmoidal_mnist}(B)).
On test inputs, the redundant coding scheme persists with deviations
on certain test inputs (Fig.~\ref{fig:sigmoidal_mnist}(C)), leading
to a clear block structure in the kernel on test data (Fig.~\ref{fig:sigmoidal_mnist}(D)).
The mean predictor is close to the test targets except on particularly
difficult test inputs and accurately captured by the theory (Fig.~\ref{fig:sigmoidal_mnist}(E));
the resulting generalization error is smaller than its counterpart
based on the GP theory (Fig.~\ref{fig:sigmoidal_mnist}(F)). Similar
to linear networks, reduction of the generalization error is mainly
driven by the reduced variance of the predictor, although in this
case also the non-lazy mean predictor performs slightly better.

Using randomly projected MNIST to enable sampling in the regime $P>N_{0}$
leads to similar results---a redundant coding scheme on training
inputs and good generalization (see supplement). In contrast, using
the first three classes of CIFAR10 instead of MNIST while keeping
$P=100$ still leads to a redundant coding on the training inputs
but the coding scheme is lost on test inputs, indicating that the
representations do not generalize well, which is confirmed by a high
generalization error (see supplement). Increasing the data set size
to all ten classes and the full training set with $P=50,000$ inputs
and using a $L=2$ hidden layer network with $N=1,000$ neurons improves
the generalization to an overall accuracy of $0.45$; in this case
the last layer training activations show a redundant coding scheme
on training inputs which generalize to varying degree to test inputs
(see supplement).

\subsection{ReLU Networks}

Last, we consider ReLU networks, $\phi(z)=\max(0,z)$. Here we set
$\sigma_{a}^{2}=1/P^{1/L}$ as in the linear case owing to the homogeneity
of ReLU. For the theory we consider only single hidden layer networks
$L=1$.

\begin{figure}
\includegraphics[width=1\columnwidth]{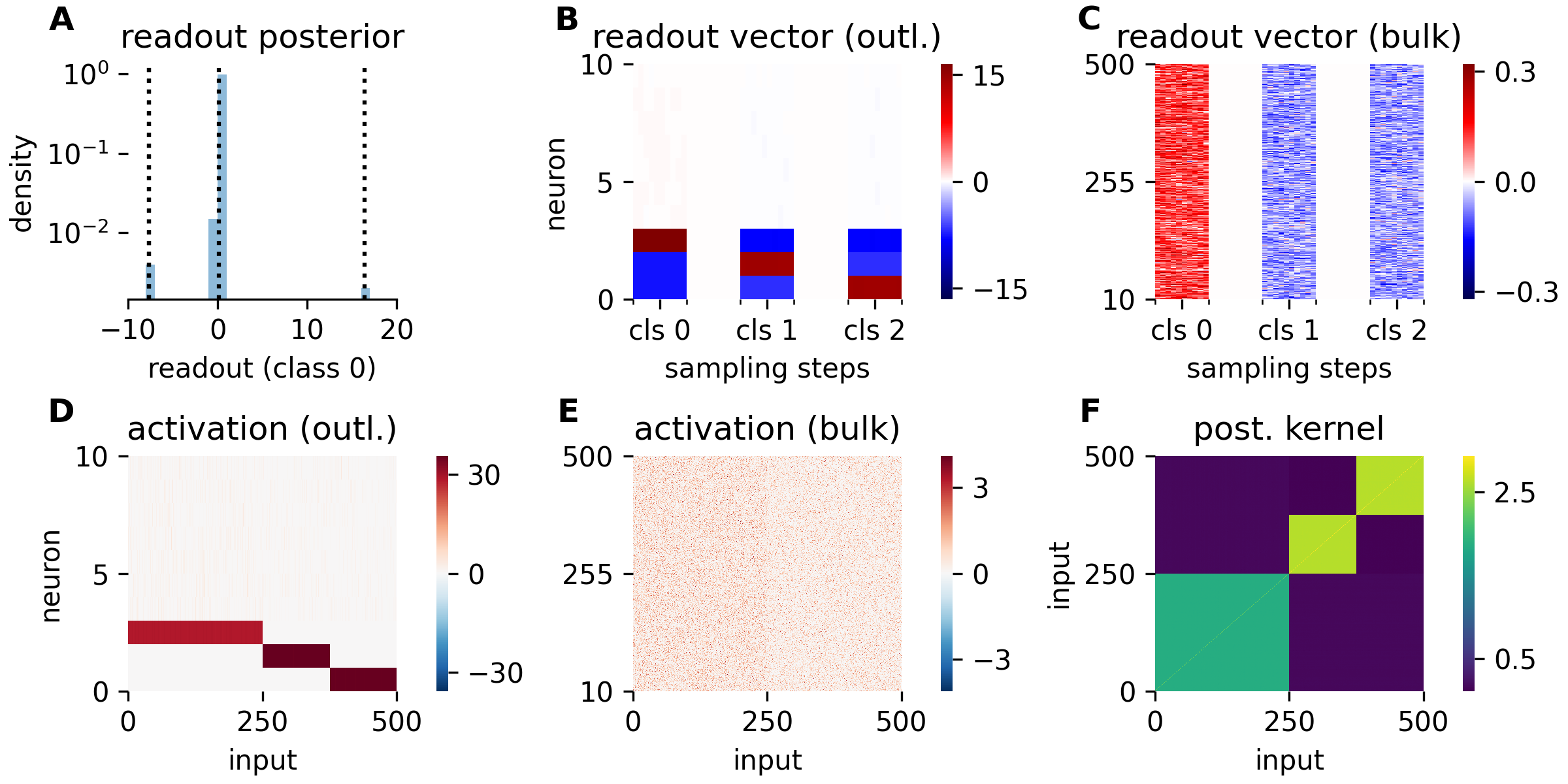}

\caption{Sparse coding in single layer ReLU networks on random classification
task. (\textbf{A}) Readout posterior of first class (blue) and theory
(black dashed). (\textbf{B},\textbf{C}) Samples of the readout weights
of all classes of the most active (B) and the remaining (C) neurons.
(\textbf{D},\textbf{E}) Activations of the most active (D) and the
remaining (E) neurons on all training inputs for a given weight sample.
(\textbf{F}) Kernel on training data from sampling. Parameters: $N=P=500$,
$N_{0}=520$, classes assigned with fixed ratios $[1/2,1/4,1/4]$,
targets $y_{+}=1$ and $y_{-}=-1/2$. \label{fig:relu_code}}
\end{figure}

\subsubsection{Training}

We start with the representations on training data. The nature of
the posterior is fundamentally different compared to the previous
cases: it factorizes into a small number $n=O(1)$ of outlier neurons
and a remaining bulk of neurons, $\prod_{i=1}^{n}[P(a_{i})P(z_{i}\,|\,a_{i})]\prod_{i=n+1}^{N}[P(a_{i})P(z_{i}\,|\,a_{i})]$.

The readout weight posteriors of the outlier neurons $i=1,\dots,n$
are Dirac deltas at $O(\sqrt{N})$ values, $P_{i}(a_{i})=\delta(a_{i}-\sqrt{N}\bar{a}_{i})$.
For the bulk neurons $i=n+1,\dots,N$, the readout weight posterior
is also a Dirac delta but the values are $O(1)$, $P(a_{i})=\delta(a_{i}-a_{0})$.
The values of $a_{0}$ and $\bar{a}_{i}$ are determined self-consistently
with the training data set (see supplement). The readout weight posterior
on the toy task is shown in Fig.~\ref{fig:relu_code}(A-C). In this
case there are $n=3$ outliers, corresponding to the $m=3$ classes.
Considering the distribution across neurons highlights the difference
in magnitude between the bulk and the outliers and demonstrates the
match between theory and sampling (Fig.~\ref{fig:relu_code}(A)).
In the plot, only two outliers are visible because the two negative
outliers have identical value due to the symmetry of the last two
classes, leading to a single peak with twice with a doubled height.
During sampling, both the outlier (Fig.~\ref{fig:relu_code}(B))
and the bulk (Fig.~\ref{fig:relu_code}(C)) readout weights are frozen
to their respective values.

The last layer representations are determined by the conditional
distribution of the preactivations. For the outliers, this distribution
is a Dirac delta at $O(\sqrt{N})$ values, $P(z_{i}\,|\,\sqrt{N}\bar{a}_{i})=\delta(z_{i}-\sqrt{N}\bar{z}_{i})$.
Thus, for the $n=O(1)$ outlier neurons the $O(\sqrt{N})$ activations
and readout weights jointly overcome the non-lazy scaling. All $N-n$
neurons of the bulk are identically distributed according to a non-Gaussian
posterior $P(z_{i}\,|\,a_{0})$. In particular, this means that the
neurons from the bulk share the same code. However, if there are multiple
classes $m>1$, or a single class with positive and negative targets,
a single code is not sufficient to solve the task. Thus, the outliers
must carry the task relevant information leading to a sparse coding
scheme. On the toy task, the outliers code for a single class each
(Fig.~\ref{fig:relu_code}(D)) while the bulk is largely task agnostic
(Fig.~\ref{fig:relu_code}(E)). The relation between readout weights
and activations is straightforward: readout weights for the outlier
neurons are positive for the coded class and negative otherwise (Fig.~\ref{fig:relu_code}(B,D)).

The kernel comprises contributions due to the outliers and the bulk:
$K=\sum_{i=1}^{n}\phi(\bar{z}_{i})\phi(\bar{z}_{i}^{\top})+\langle\phi(z_{0})\phi(z_{0}^{\top})\rangle_{z_{0}|a_{0}}$.
The contribution due to the outliers is $O(1)$ because their activations
$z_{i}=\sqrt{N}\bar{z}_{i}$ are $O(\sqrt{N})$ which cancels the
$1/N$ in front of the kernel (\ref{eq:def_kernel}). On the toy task,
the outlier contribution leads to a dominant low rank component (Fig.~\ref{fig:relu_code}(F)).

\begin{figure}
\includegraphics[width=1\columnwidth]{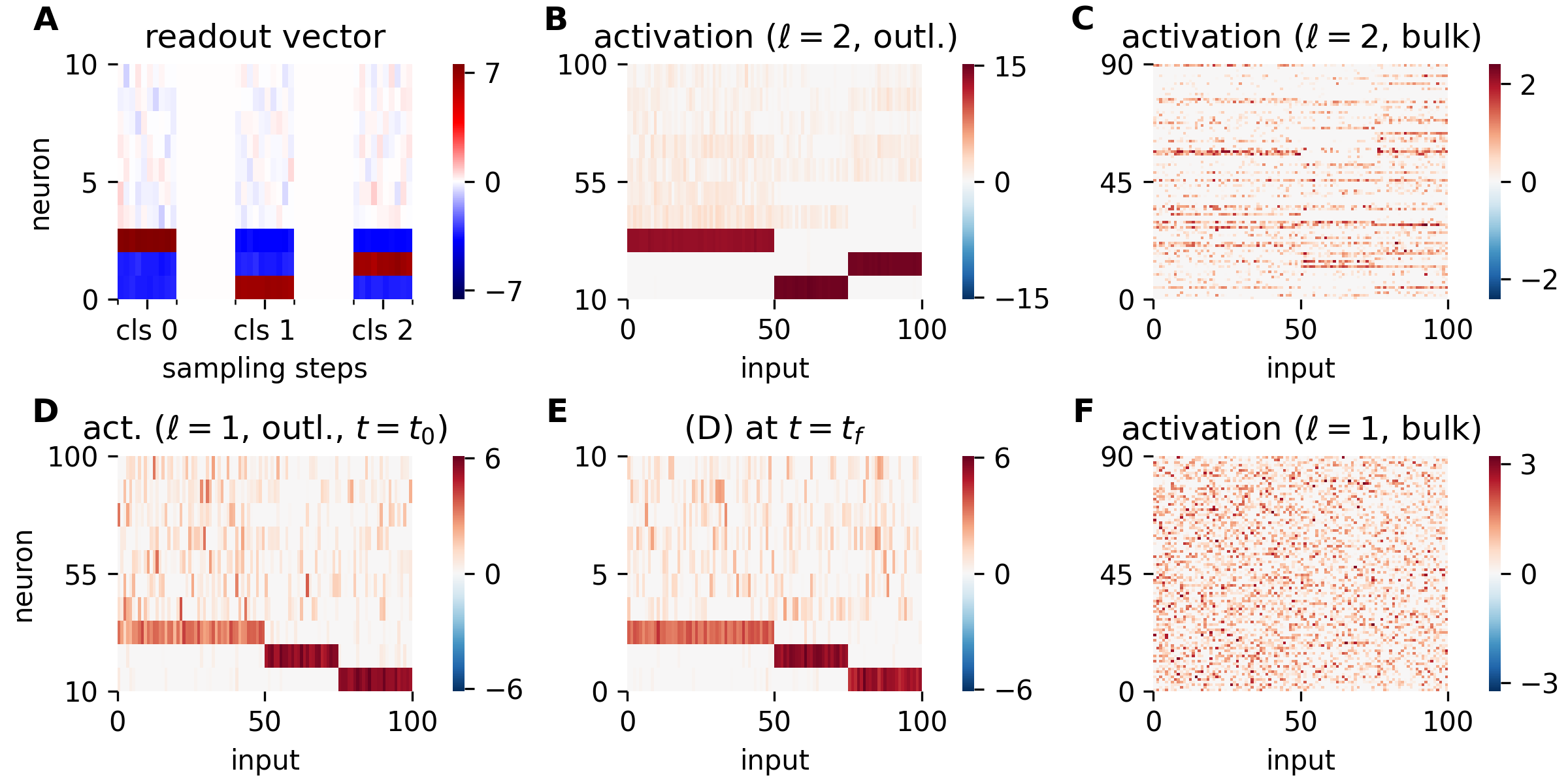}

\caption{Sparse coding in two layer ReLU networks on random classification
task. (\textbf{A}) Samples of the readout weights of all classes for
the most active neurons. (\textbf{B},\textbf{C}) Activations of the
most active (B) and the remaining (C) neurons in the second layer
on all training inputs for a given weight sample. (\textbf{D},\textbf{E},\textbf{F})
Activations of the most active (D,E) and the remaining (F) neurons
in the first layer using the first (D,F) or the last (E) weight sample.
Parameters: $N=P=100$, $N_{0}=120$, classes assigned with fixed
ratios $[1/2,1/4,1/4]$, targets $y_{+}=1$ and $y_{-}=-1/2$. \label{fig:relu_code_multilayer}}
\end{figure}

We investigate deeper networks numerically. For two hidden layers
($L=2$) on the toy task, the readout weights contain three outliers
(Fig.~\ref{fig:relu_code_multilayer}(A)) and the last layer shows
a sparse coding of the task with three outlier neurons coding for
one class each (Fig.~\ref{fig:relu_code_multilayer}(B)). A difference
to the single layer case is that the bulk also shows a weak coding
for the task (Fig.~\ref{fig:relu_code_multilayer}(C)). In the first
layer there are again three outliers coding for one class each (Fig.~\ref{fig:relu_code_multilayer}(D)).
Unlike in sigmoidal networks the codes are frozen, i.e., the outliers
do not change identity or switch with the bulk (Fig.~\ref{fig:relu_code_multilayer}(E)).
Finally, the bulk in the first layer is task agnostic (Fig.~\ref{fig:relu_code_multilayer}(F)).

\begin{figure}
\includegraphics[width=1\columnwidth]{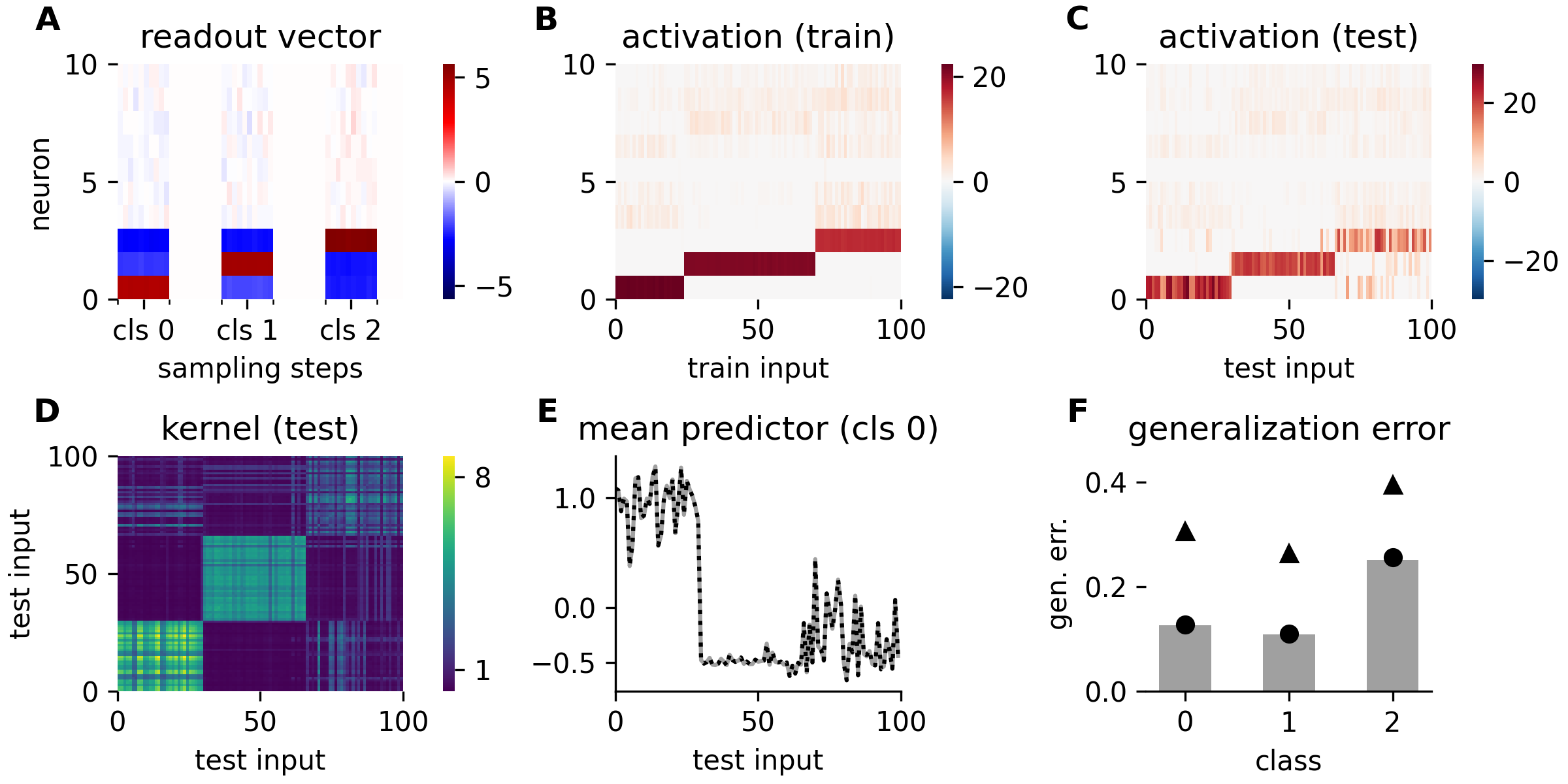}

\caption{Coding scheme and generalization of one hidden layer ReLU networks
on MNIST. (\textbf{A}) Samples of all three readout weights for most
active neurons. (\textbf{B},\textbf{C}) Activations of most active
neurons on all training (B) and $100$ test (C) inputs for a given
weight sample. (\textbf{D}) Kernel on $100$ test inputs. (\textbf{E})
Mean predictor for class 0 from sampling (gray) and theory (Eq.~(\ref{eq:predictor_relu}),
black dashed). (\textbf{F}) Generalization error for each class averaged
over $1,000$ test inputs from sampling (gray bars), theory (Eq.~(\ref{eq:predictor_sigmoidal}),
black circles), and GP theory (back triangles). Parameters: $N=P=100$,
$N_{0}=784$, classes 0, 1, 2 assigned randomly with probability $1/3$,
targets $y_{+}=1$ and $y_{-}=-1/2$. \label{fig:relu_mnist}}
\end{figure}

\subsubsection{Generalization}

For generalization we include the test preactivations $z_{i}^{*}$
on $P_{*}$ test inputs $x^{*}$. As in the previous cases, this leads
to the Gaussian conditional distribution $P(z_{i}^{*}\,|\,z_{i})$.
The predictor on a test input $x$ is
\begin{equation}
f_{r}(x)=\sum_{i=1}^{n}\bar{a}_{i}^{r}\phi(\bar{z}_{i}(x))+a_{0}^{r}\langle\langle\phi[z(x)]\rangle_{z(x)|z}\rangle_{z|a_{0}}\label{eq:predictor_relu}
\end{equation}
with $\bar{z}_{i}(x)=\bar{z}_{i}^{\top}K_{0}^{+}k_{0}(x)$ for $i=1,\dots,n$.
As for the other nonlinearities, the predictor variance can be neglected.

On the first three digits of MNIST, the readout weights are sparse
(Fig.~\ref{fig:relu_mnist}(A)) and the training (Fig.~\ref{fig:relu_mnist}(B))
as well as test (Fig.~\ref{fig:relu_mnist}(C)) activations exhibit
three outliers which code for one class each. The kernel on test data
displays a clear block structure (Fig.~\ref{fig:relu_mnist}(D))
due to the outliers. For the theory, we neglect the bulk to make it
tractable. Despite this approximation, the predictor is accurately
captured by the theory (Fig.~\ref{fig:relu_mnist}(E)). In terms
of the generalization error, the nonlazy network outperforms the GP
predictor (Fig.~\ref{fig:relu_mnist}(F)) due to the reduction of
the predictor variance.

\section{Discussion}

We developed a theory for the weight posterior of non-lazy networks
in the limit of infinite width and data set size, from which we derived
analytical expressions for the single neuron posteriors of the readout
weights and the preactivations on training and test inputs. These
single neuron posteriors revealed that the learned representations
are embedded into the network using distinct coding schemes. Furthermore,
we used the single neuron posteriors to derive the mean predictor
and the mean kernels on training and test inputs. We applied the theory
to two classification tasks: a simple toy model using orthogonal data
and random labels to highlight the coding schemes and image classification
using MNIST and CIFAR10 to investigate generalization. In both cases,
the theoretical results are in excellent agreement with empirical
samples from the weight posterior.

\paragraph{Coding Schemes}

We show that the embedding of learned representations by the neurons
exhibits a remarkable structure: a coding scheme where distinct populations
of neurons are characterized by the subset of classes that activate
them. The details of the coding scheme depend strongly on the nonlinearity:
linear networks display an analog coding scheme where all neurons
code for all classes in a graded manner (Figs.~\ref{fig:linear_code},
\ref{fig:linear_mnist}), sigmoidal networks display a redundant coding
scheme where large populations of neurons are active on specific combinations
of classes (Figs.~\ref{fig:sigmoidal_code}, \ref{fig:sigmoidal_code_multilayer},
\ref{fig:sigmoidal_mnist}), and ReLU networks display a sparse coding
scheme in which a few individual neurons are active on specific classes
while the remaining neurons are task agnostic (Figs.~\ref{fig:relu_code},
\ref{fig:relu_code_multilayer}, \ref{fig:relu_mnist}). In networks
with multiple hidden layers, the coding scheme appears in all layers
but is sharpened across layers; the coding scheme in the last layer
remains the same as in the single layer case (Figs.~\ref{fig:sigmoidal_code_multilayer},
\ref{fig:relu_code_multilayer}).

Why were the coding schemes not previously observed? Compared to
standard training protocols, there are two main differences: (1) the
sampling from the posterior, i.e., training for a long time with added
noise, and (2) the data set size dependent strength of the readout
weight variance, which corresponds to a data set size dependent regularization
using weight decay. As we show in the supplement, training without
noise also leads to sparse or redundant coding schemes. Thus, the
data set size dependent regularization seems to be crucial which is,
to the best of our knowledge, not commonly used in practice.

\paragraph{Permutation Symmetry Breaking}

The coding schemes determine the structure of a typical solution
sampled from the weight posterior, e.g., for sigmoidal networks a
solution with a redundant coding scheme. Due to the neuron permutation
symmetry of the posterior, each typical solution has permuted counterparts
which are also typical solutions. Permutation symmetry breaking occurs
if these permuted solutions are disconnected in solution space, i.e.,
if the posterior contains high barriers between the permuted solutions.

In sigmoidal and ReLU networks, permutation symmetry is broken. The
theoretical signature of this symmetry breaking is the disconnected
structure of the posterior of the readout weights where the different
branches are separated by high barriers. Numerically, symmetry breaking
is evidenced by the fact that, at equilibrium, neurons do not change
their coding identity with sampling time. We note that for the readout
weights, the symmetry broken phase is also a frozen state, namely
not only the coding structure is fixed but also there are no fluctuations
in the magnitude of each weight (Figs.~\ref{fig:sigmoidal_code},
\ref{fig:sigmoidal_code_multilayer}, \ref{fig:relu_code}, \ref{fig:relu_code_multilayer}).
In contrast, activations in the hidden layer, also constant in their
coding, do exhibit residual temporal fluctuations around a pronounced
mean (Figs.~\ref{fig:sigmoidal_code}, \ref{fig:sigmoidal_code_multilayer})
which carries the task relevant information.

Interestingly, the situation is more involved in networks with two
hidden layers. In the first layer, permutation symmetry is broken
and the neurons' code is frozen in ReLU networks (Fig.~\ref{fig:relu_code_multilayer})
but not in sigmoidal networks (Fig.~\ref{fig:sigmoidal_code_multilayer}).
In the latter case, the typical solution has a prominent coding scheme
but individual neurons switch their code during sampling (Fig.~\ref{fig:sigmoidal_code_multilayer}).

Symmetry breaking has been frequently discussed in the context of
learning in artificial neural networks, for example replica symmetry
breaking in perceptrons with binary weights \citep{Krauth1989,PhysRevA.45.6056,RevModPhys.65.499},
permutation symmetry breaking in fully connected networks \citep{PhysRevA.45.4146,PhysRevA.45.7590}
and restricted Boltzmann machines \citep{Hou_2019}, or continuous
symmetry breaking of the ``kinetic energy'' (reflecting the learning
rule) \citep{kunin2021neural,NEURIPS2021_d76d8dee}. Furthermore,
\citep{rubin2024grokking} links breaking of parity symmetry to feature
learning, albeit in a different scaling limit. The role of an intact
(not broken) permutation symmetry has been explored in the context
of the connectedness of the loss landscape \citep{brea2019weightspace,pmlr-v139-simsek21a,entezari2022the}.
Here, we establish, for the first time, a direct link between symmetry
breaking and the nature of the neural representations.

\paragraph{Neural Collapse}

The redundant coding scheme in sigmoidal network and the sparse coding
scheme in ReLU networks are closely related to the phenomenon of neural
collapse \citep{doi:10.1073/pnas.2015509117}. The two main properties
of collapse are (1) vanishing variability across inputs from the same
class of the last layer postactivations and (2) last layer postactivations
forming an equiangular tight frame (centered class means are equidistant
and equiangular with maximum angle). There are two additional properties
which, however, follow from the first two under minimal assumptions
\citep{doi:10.1073/pnas.2015509117}. We note that collapse is determined
only on training data in the original definition.

Formally, the first property of collapse is violated in the non-lazy
networks investigated here due to the non vanishing across-neuron
variability of the activations given the readout weights. However,
the mean activations conditional on the readout weights, which carry
the task relevant information, generate an equiangular tight frame
in both sigmoidal and ReLU networks. This creates an interesting link
to empirically trained networks where neural collapse has been shown
to occur under a wide range of conditions \citep{doi:10.1073/pnas.2015509117,han2022neural}.

Neglecting the non-vanishing variability, the main difference between
neural collapse and the coding schemes is that the latter impose a
more specific structure. Technically, the equiangular tight frame
characterizing neural collapse is invariant under orthogonal transformations
while the coding schemes are invariant under permutations, which is
a subset of the orthogonal transformations. This additional structure
makes the representations highly interpretable in terms of a neural
code---conversely, applying, e.g., a rotation in neuron space to
the redundant or sparse coding scheme would hide the immediately apparent
structure of the solution.

\paragraph*{Representation Learning and Generalization}

The case of non-lazy linear networks makes an interesting point about
the interplay between feature learning and generalization: although
the networks learn strong, task-dependent representations, the mean
predictor is identical to the Gaussian Process limit where the features
are random. The only difference between non-lazy networks and random
features is a reduction in the predictor variance. Thus, this provides
an explicit example where feature learning only mildly helps generalization
through reduction of the predictor variance.

More generally, in all examples the improved performance of nonlazy
networks (Figs.~\ref{fig:linear_mnist}, \ref{fig:sigmoidal_mnist},
\ref{fig:relu_mnist}) is mainly driven by the reduction of the predictor
variance; the mean predictor does not generalize significantly better
than a predictor based on random features. This shows an important
limitation of our work: in order to achieve good generalization performance,
it might be necessary to consider deeper nonlinear architectures or
additional structure in the model.

While the learned representations might not necessarily help generalization
on the trained task, they can still be useful for few shot learning
of a novel task \citep{1597116,NIPS2016_90e13578,NIPS2017_cb8da676,doi:10.1073/pnas.2200800119}.
Indeed, neural collapse has been shown to be helpful for transfer
learning if neural collapse still holds (approximately) on inputs
from the novel classes \citep{galanti2022on}. Due to the relation
between coding schemes and neural collapse, this suggests that the
learned representations investigated here are useful for downstream
task---if the nature of the solution does not change on the new inputs.
This remains to be systematically explored.

\paragraph{Lazy vs.~Non-Lazy Regime}

The definition of lazy vs.~non-lazy regime is subtle. Originally,
the lazy regime was defined by the requirement that the learned changes
in the weights affect the final output only linearly \citep{NEURIPS2018_5a4be1fa}.
This implies that the learned changes in the representations are weak
since changes in the hidden layer weights nonlinearly affect the final
output. To overcome the weak representation learning, \citep{pmlr-v139-yang21c,bordelon2022selfconsistent}
define the non-lazy regime as $O(1)$ changes of the features during
the first steps of gradient descent. This definition leads to an initialization
where the readout scales its inputs with $1/N$ \citep{bordelon2022selfconsistent}.
However, the scaling might change during training---which is not
captured by a definition at initialization.

We here define the non-lazy regime such that the readout scales its
inputs with $1/N$ after learning, i.e., under the posterior. Importantly,
we prevent that the readout weights overcome the scaling by growing
their norm, which leads to the requirement of a decreasing prior variance
of the readout weights with increasing $P$. The definition implies
that strong representations are learned: the readout weights must
be aligned with the last layer's features on all training inputs,
which is not the case for random features.

\begin{acknowledgments}
Many helpful discussions with Qianyi Li and feedback on the manuscript
by Lorenzo Tiberi, Chester Mantel, Kazuki Irie, and Haozhe Shan are
gratefully acknowledged. This research was supported by the Swartz
Foundation, the Gatsby Charitable Foundation, and in part by grant
NSF PHY-1748958 to the Kavli Institute for Theoretical Physics (KITP).
\end{acknowledgments}

\clearpage{}

\newpage{}

\appendix
\onecolumngrid

\section{Linear Networks}

\subsection{Training Posterior}

We start with the joint posterior of the readout weights and the training
preactivations. The change of variable from the $N\times N$ (or $N\times N_{0}$)
weights $W_{\ell}$ to the $N\times P$ preactivations $Z_{\ell}=\frac{1}{\sqrt{N}}W_{\ell}Z_{\ell-1}$
and $Z_{1}=\frac{1}{\sqrt{N_{0}}}W_{1}X$ replaces the prior distribution
of the weights by the prior distribution of the preactivations which
is $P_{0}(Z_{\ell})=\mathcal{N}(Z_{\ell}\,|\,0,I_{N}\otimes\sigma_{\ell}^{2}K_{\ell-1})$
with $K_{\ell}=\frac{1}{N}Z_{\ell}^{\top}Z_{\ell}$, $\ell=1,\dots,L$,
and $K_{0}=\frac{1}{N_{0}}X^{\top}X$. Throughout, we use the matrix
valued normal distribution (see, e.g., \citep{Gupta1999}) to ease
the notation, e.g., $P(Z\,|\,0,I_{N}\otimes K)$ means $Z$ is zero
mean Gaussian with correlation $\langle Z_{i\mu}Z_{j\nu}\rangle=\delta_{ij}K_{\mu\nu}$.
We can write the training posterior as
\begin{equation}
P(A,Z_{L},\dots,Z_{1})\propto\mathcal{N}\big(Y\,|\,\frac{1}{N}A^{\top}Z_{L},\frac{T}{N}I_{m}\otimes I_{P}\big)P_{0}(A)\prod_{\ell=1}^{L}P_{0}(Z_{\ell})
\end{equation}
where we already rescaled $T\to T/N$ and denoted the $N\times m$
matrix containing all readout weights by $A$.

Using the identity
\begin{equation}
\mathcal{N}\big(Y\,|\,m,u\otimes K\big)=\int dt\,e^{-i\tr t^{\top}Y+i\tr t^{\top}m-\frac{1}{2}\tr ut^{\top}Kt},\label{eq:icf_Gauss}
\end{equation}
where $t$ is a $P\times m$ matrix, decouples the first factor across
neurons. The remaining factors $P_{0}(Z_{\ell})$ are only coupled
through the kernel $K_{\ell}$. Hence, we decouple these factors by
introducing $K_{\ell}$, $\ell=1,\dots,L-1$, using Dirac deltas,
$\delta(K_{\ell}-\frac{1}{N}Z_{\ell}^{\top}Z_{\ell})=\int d\tilde{K}_{\ell}\,e^{-i\tr\tilde{K}_{\ell}^{\top}K_{\ell}+\frac{i}{N}\tr\tilde{K}_{\ell}^{\top}Z_{\ell}^{\top}Z_{\ell}}$.
The training posterior factorizes across neurons into
\begin{align}
P(A,Z_{L},\dots,Z_{1})\propto\int & dt\prod_{\ell=1}^{L-1}[dK_{\ell}d\tilde{K}_{\ell}]\,e^{\frac{T}{2}N\tr t^{\top}t-N\tr t^{\top}Y-\frac{1}{2}N\sum_{\ell=1}^{L-1}\tr\tilde{K}_{\ell}^{\top}K_{\ell}}\nonumber \\
 & \times\prod_{i=1}^{N}\Big[\underbrace{P_{0}(a_{i})\mathcal{N}(z_{i}^{L}\,|\,0,\sigma_{L}^{2}K_{L-1})e^{a_{i}^{\top}t^{\top}z_{i}^{L}}}_{\propto P(a_{i},z_{i}^{L})}\Big]\prod_{\ell=1}^{L-1}\Big[\prod_{j=1}^{N}\underbrace{\mathcal{N}(z_{j}^{\ell}\,|\,0,\sigma_{\ell}^{2}K_{\ell-1})e^{\frac{1}{2}z_{j}^{\ell\top}\tilde{K}_{\ell}z_{j}^{\ell}}}_{\propto P(z_{j}^{\ell})}\Big]
\end{align}
where we rescaled $it\to Nt$ and $i\tilde{K}_{\ell}\to\frac{1}{2}N\tilde{K}_{\ell}$.
The factorized single neuron posteriors are not yet normalized. Taking
the normalization into account leads to
\begin{equation}
P(A,Z_{L},\dots,Z_{1})\propto\int dt\prod_{\ell=1}^{L-1}[dK_{\ell}d\tilde{K}_{\ell}]\,e^{-NE(t,\{K_{\ell},\tilde{K}_{\ell}\})}\prod_{i=1}^{N}[P(a_{i},z_{i}^{L})]\prod_{\ell=1}^{L-1}[\prod_{j=1}^{N}P(z_{j}^{\ell})],
\end{equation}
\begin{align}
E(t,\{K_{\ell},\tilde{K}_{\ell}\})= & -\frac{T}{2}\tr t^{\top}t+\tr t^{\top}Y+\frac{1}{2}\sum_{\ell=1}^{L-1}\tr\tilde{K}_{\ell}^{\top}K_{\ell}-\log\int dP_{0}(a)\int d\mathcal{N}(z^{L}\,|\,0,\sigma_{L}^{2}K_{L-1})\,e^{a^{\top}t^{\top}z^{L}}\nonumber \\
 & -\sum_{\ell=1}^{L-1}\log\int d\mathcal{N}(z^{\ell}\,|\,0,\sigma_{\ell}^{2}K_{\ell-1})\,e^{\frac{1}{2}z^{\ell\top}\tilde{K}_{\ell}z^{\ell}}.
\end{align}
The proportionality constant ensures $\int dt\prod_{\ell=1}^{L-1}dK_{\ell}d\tilde{K}_{\ell}\,e^{-NE(t,\{K_{\ell},\tilde{K}_{\ell}\})}=1$.

Using a saddle point approximation \citep{fedoryuk1977saddle,Fedoryuk1989}
of the $t$, $K_{\ell}$, and $\tilde{K}_{\ell}$ integrals we arrive
at
\begin{equation}
P(A,Z_{L},\dots,Z_{1})=\prod_{i=1}^{N}[P(a_{i})P(z_{i}^{L}\,|\,a_{i})]\prod_{\ell=1}^{L-1}[\prod_{j=1}^{N}P(z_{j}^{\ell})]
\end{equation}
where we also used Bayes rule $P(a_{i},z_{i}^{L})=P(a_{i})P(z_{i}^{L}\,|\,a_{i})$.
The saddle point equations are
\begin{equation}
y_{\mu}^{r}=\langle a_{r}z_{\mu}^{L}\rangle_{a,z^{L}}+Tt_{\mu}^{r}
\end{equation}
\begin{equation}
K_{\ell}=\langle z^{\ell}z^{\ell\top}\rangle_{z^{\ell}},\quad\ell=1,\dots,L-1
\end{equation}
\begin{equation}
\tilde{K}_{\ell-1}=\frac{1}{\sigma_{\ell}^{2}}K_{\ell-1}^{-1}\langle z^{\ell}z^{\ell\top}\rangle_{z^{\ell}}K_{\ell-1}^{-1}-K_{\ell-1}^{-1},\quad\ell=2,\dots,L
\end{equation}
The single neuron posteriors are all Gaussian:
\begin{equation}
P(a)=\mathcal{N}(a\,|\,0,U)\label{eq:a_posterior_lin}
\end{equation}
\begin{equation}
P(z^{L}\,|\,a)=\mathcal{N}(z^{L}\,|\,\sigma_{L}^{2}K_{L-1}ta,\sigma_{L}^{2}K_{L-1})\label{eq:zL_posterior_lin}
\end{equation}
\begin{equation}
P(z^{\ell})=\mathcal{N}(z^{\ell}\,|\,0,[\sigma_{\ell}^{-2}K_{\ell-1}^{-1}-\tilde{K}_{\ell}]^{-1}),\quad\ell=1,\dots,L-1
\end{equation}
where we introduced the $m\times m$ covariance matrix of the readout
weights
\begin{equation}
U=\big(\sigma_{a}^{-2}I_{m}-\sigma_{L}^{2}t^{\top}K_{L-1}t\big)^{-1}.
\end{equation}

The expectations in the saddle point equations yield
\begin{equation}
\langle az^{L\top}\rangle_{a,z^{L}}=\sigma_{L}^{2}Ut^{\top}K_{L-1}
\end{equation}
\begin{equation}
\langle z^{\ell}z^{\ell\top}\rangle_{z^{\ell}}=\big(\sigma_{\ell}^{-2}K_{\ell-1}^{-1}-\tilde{K}_{\ell}\big)^{-1},\quad\ell=1,\dots,L-1
\end{equation}

We explicitly solve the saddle point equations at $T=0$ and assume
that all kernels are invertible. The first saddle point equation yields
$t=\sigma_{L}^{-2}K_{L-1}^{-1}YU^{-1}$. Computing the last layer
kernel leads to
\begin{equation}
K_{L}=\sigma_{L}^{2}K_{L-1}+YU^{-1}Y^{\top}
\end{equation}
which reduces the saddle point equation for $\tilde{K}_{L-1}$ to
\begin{equation}
\tilde{K}_{L-1}=\sigma_{L}^{-2}K_{L-1}^{-1}YU^{-1}Y^{\top}K_{L-1}^{-1}
\end{equation}
For the lower layers we combine $\tilde{K}_{\ell-1}=\frac{1}{\sigma_{\ell}^{2}}K_{\ell-1}^{-1}K_{\ell}K_{\ell-1}^{-1}-K_{\ell-1}^{-1}$
and $K_{\ell}=[\sigma_{\ell}^{-2}K_{\ell-1}^{-1}-\tilde{K}_{\ell}]^{-1}$
to $\sigma_{\ell}^{2}K_{\ell-1}=\sigma_{\ell+1}^{2}K_{\ell}K_{\ell+1}^{-1}K_{\ell}$.
For $\ell=L-1$ we obtain
\begin{equation}
\sigma_{L-1}^{2}K_{L-2}=\sigma_{L}^{2}K_{L-1}K_{L}^{-1}K_{L-1}=K_{L-1}-\sigma_{L}^{-2}Y\big(U+\sigma_{L}^{-2}Y^{\top}K_{L-1}^{-1}Y\big)^{-1}Y^{\top}
\end{equation}

Inserting $t=\sigma_{L}^{-2}K_{L-1}^{-1}YU^{-1}$ into the definition
of $U$ leads to $U+\sigma_{L}^{-2}Y^{\top}K_{L-1}^{-1}Y=\sigma_{a}^{-2}U^{2}$
and thus 
\begin{equation}
K_{L-1}=\sigma_{L-1}^{2}K_{L-2}+\frac{\sigma_{a}^{2}}{\sigma_{L}^{2}}YU^{-2}Y^{\top}
\end{equation}
Evaluating $\sigma_{\ell}^{2}K_{\ell-1}=\sigma_{\ell+1}^{2}K_{\ell}K_{\ell+1}^{-1}K_{\ell}$
at $\ell=L-2$ leads to
\begin{equation}
\sigma_{L-2}^{2}K_{L-3}=\sigma_{L-1}^{2}K_{L-2}K_{L-1}^{-1}K_{L-2}=K_{L-2}-\sigma_{L-1}^{-2}Y\big(\frac{\sigma_{L}^{2}}{\sigma_{a}^{2}}U^{2}+\sigma_{L-1}^{-2}Y^{\top}K_{L-2}^{-1}Y\big)^{-1}Y^{\top}
\end{equation}
Inserting $K_{L-1}=\sigma_{L-1}^{2}K_{L-2}+\frac{\sigma_{a}^{2}}{\sigma_{L}^{2}}YU^{-2}Y^{\top}$
into $U+\sigma_{L}^{-2}Y^{\top}K_{L-1}^{-1}Y=\sigma_{a}^{-2}U^{2}$
leads to $\sigma_{a}^{-2}\sigma_{L}^{2}U^{2}+\sigma_{L-1}^{-2}Y^{\top}K_{L-2}^{-1}Y=\sigma_{a}^{-4}\sigma_{L}^{2}U^{3}$
and thus
\begin{equation}
K_{L-2}=\sigma_{L-2}^{2}K_{L-3}+\frac{\sigma_{a}^{4}}{\sigma_{L}^{2}\sigma_{L-1}^{2}}YU^{-3}Y^{\top}
\end{equation}
Iterating these steps yields
\begin{equation}
U=\sigma_{a}^{2}I+\frac{\sigma_{a}^{2L}}{\prod_{\ell=1}^{L}\sigma_{\ell}^{2}}U^{-L}Y^{\top}K_{0}^{-1}Y\label{eq:U_self_consistency}
\end{equation}
\textbf{Readout weight scaling:} For $Y^{\top}K_{0}^{-1}Y=O(P)$ we
need to scale $\sigma_{a}^{2}=O(1/P^{1/L})$ to achieve finite norm
of the readout weights, $U=O(1)$; not scaling $\sigma_{a}^{2}$ would
lead to $U=O(P^{1/(L+1)})$ and hence a growing readout weight norm.
In both cases the first term on the r.h.s.~can be neglected at large
$P$, leading to $U^{L+1}=\frac{\sigma_{a}^{2L}}{\prod_{\ell=1}^{L}\sigma_{\ell}^{2}}Y^{\top}K_{0}^{-1}Y$.

For the first layer kernel, iterating the above steps leads to
\begin{equation}
K_{1}=\sigma_{1}^{2}K_{0}+\frac{\sigma_{a}^{2(L-1)}}{\prod_{\ell=2}^{L}\sigma_{\ell}^{2}}YU^{-L}Y^{\top}
\end{equation}
For the kernels in the later layers $2\le\ell\le L-1$ we use $K_{\ell}=\sigma_{\ell}^{2}K_{\ell-1}+\frac{\sigma_{a}^{2(L-\ell)}}{\prod_{\ell'=\ell+1}^{L}\sigma_{\ell'}^{2}}YU^{-(L-\ell+1)}Y^{\top}$
and iterate, starting from $\ell=2$. For each $\ell$, $K_{\ell-1}$
contains a low rank contribution of strength $\sigma_{a}^{2(L-\ell+1)}$
which is suppressed by $O(1/P^{1/L})$ compared to the new low rank
contribution of strength $\sigma_{a}^{2(L-\ell)}$, hence
\begin{equation}
K_{\ell}=\prod_{\ell'=1}^{\ell}[\sigma_{\ell'}^{2}]K_{0}+\frac{\sigma_{a}^{2(L-\ell)}}{\prod_{\ell'=\ell+1}^{L}\sigma_{\ell'}^{2}}YU^{-(L-\ell+1)}Y^{\top},\quad2\le\ell\le L-1
\end{equation}
For the last layer we use $K_{L}=\sigma_{L}^{2}K_{L-1}+YU^{-1}Y^{\top}$
and obtain
\begin{equation}
K_{L}=\prod_{\ell=1}^{L}[\sigma_{\ell}^{2}]K_{0}+YU^{-1}Y^{\top}
\end{equation}
Setting $\sigma_{\ell}=1$ recovers the results stated in the main
text.

\fbox{\begin{minipage}[t]{1\columnwidth - 2\fboxsep - 2\fboxrule}%
\textbf{Summary:} Posterior distributions
\begin{equation}
P(a)=\mathcal{N}(a\,|\,0,U)
\end{equation}
\begin{equation}
P(z^{L}\,|\,a)=\mathcal{N}(z^{L}\,|\,YU^{-1}a,\sigma_{L}^{2}K_{L-1})
\end{equation}
\begin{equation}
P(z^{\ell})=\mathcal{N}(z^{\ell}\,|\,0,K_{\ell}),\quad\ell=1,\dots,L-1,
\end{equation}
with
\begin{equation}
U^{L+1}=\frac{\sigma_{a}^{2L}}{\prod_{\ell=1}^{L}\sigma_{\ell}^{2}}Y^{\top}K_{0}^{-1}Y,
\end{equation}
\begin{equation}
K_{L}=\prod_{\ell=1}^{L}[\sigma_{\ell}^{2}]K_{0}+YU^{-1}Y^{\top},
\end{equation}
\begin{equation}
K_{\ell}=\prod_{\ell'=1}^{\ell}[\sigma_{\ell'}^{2}]K_{0}+\frac{\sigma_{a}^{2(L-\ell)}}{\prod_{\ell'=\ell+1}^{L}\sigma_{\ell'}^{2}}YU^{-(L-\ell+1)}Y^{\top},\quad\ell=1,\dots,L-1
\end{equation}
and $K_{0}=\frac{1}{N_{0}}X^{\top}X$.%
\end{minipage}}

\subsection{Test Posterior}

For test inputs, we need to include the preactivations corresponding
to the set of $P_{*}$ test inputs $x_{*}$. Thus, we change variables
from the $N\times N$ (or $N\times N_{0}$) weights $W_{\ell}$ to
the $N\times(P+P_{*})$ preactivations $\hat{Z}_{\ell}=\frac{1}{\sqrt{N}}W_{\ell}\hat{Z}_{\ell-1}$
and $\hat{Z}_{1}=\frac{1}{\sqrt{N_{0}}}W_{1}\hat{X}$ where $\hat{X}$
denotes the $N_{0}\times(P+P_{*})$ matrix containing the training
and test inputs. Conditional on the previous layer preactivations,
the prior distribution of the preactivations is $P_{0}(\hat{Z}_{\ell})=\mathcal{N}(Z_{\ell}\,|\,0,I_{N}\otimes\sigma_{\ell}^{2}\hat{K}_{\ell-1})$
with $\hat{K}_{\ell}=\frac{1}{N}\hat{Z}_{\ell}^{\top}\hat{Z}_{\ell}$,
$\ell=1,\dots,L$, and $\hat{K}_{0}=\frac{1}{N_{0}}\hat{X}^{\top}\hat{X}$.
As for the training posterior, we can write the test posterior as
\begin{equation}
P(A,\hat{Z}_{L},\dots,\hat{Z}_{1})\propto\mathcal{N}\big(Y\,|\,\frac{1}{N}A^{\top}Z_{L},\frac{T}{N}I_{m}\otimes I_{P}\big)P_{0}(A)\prod_{\ell=1}^{L}P_{0}(\hat{Z}_{\ell}).
\end{equation}
In the first factor, only the training preactivations $Z_{\ell}$
appear since the loss does not depend on the test set.

The first factor decouples using (\ref{eq:icf_Gauss}); for the preactivations
we need to introduce the $(P+P_{*})\times(P+P_{*})$ kernels $\hat{K}_{\ell}$
for $\ell=1,\dots,L-1$. This leads to
\begin{align}
P(A,\hat{Z}_{L},\dots,\hat{Z}_{1})\propto\int & dt\prod_{\ell=1}^{L-1}[d\hat{K}_{\ell}d\tilde{\hat{K}}_{\ell}]\,e^{\frac{T}{2}N\tr t^{\top}t-N\tr t^{\top}Y-\frac{1}{2}N\sum_{\ell=1}^{L-1}\tr\tilde{\hat{K}}_{\ell}^{\top}\hat{K}_{\ell}}\nonumber \\
 & \times\prod_{i=1}^{N}\Big[\underbrace{P_{0}(a_{i})\mathcal{N}(\hat{z}_{i}^{L}\,|\,0,\sigma_{L}^{2}\hat{K}_{L-1})e^{a_{i}^{\top}t^{\top}z_{i}^{L}}}_{\propto P(a_{i},\hat{z}_{i}^{L})}\Big]\prod_{\ell=1}^{L-1}\Big[\prod_{j=1}^{N}\underbrace{\mathcal{N}(\hat{z}_{j}^{\ell}\,|\,0,\sigma_{\ell}^{2}\hat{K}_{\ell-1})e^{\frac{1}{2}\hat{z}_{j}^{\ell\top}\tilde{\hat{K}}_{\ell}\hat{z}_{j}^{\ell}}}_{\propto P(\hat{z}_{j}^{\ell})}\Big].
\end{align}
Note that in the last factor of $P(a_{i},\hat{z}_{i}^{L})$ only the
training preactivations appear. Normalizing the single neuron posteriors
yields
\begin{equation}
P(A,\hat{Z}_{L},\dots,\hat{Z}_{1})\propto\int dt\prod_{\ell=1}^{L-1}[d\hat{K}_{\ell}d\tilde{\hat{K}}_{\ell}]\,e^{-NE(t,\{\hat{K}_{\ell},\tilde{\hat{K}}_{\ell}\})}\prod_{i=1}^{N}[P(a_{i},\hat{z}_{i}^{L})]\prod_{\ell=1}^{L-1}[\prod_{j=1}^{N}P(\hat{z}_{j}^{\ell})],
\end{equation}
\begin{align}
E(t,\{\hat{K}_{\ell},\tilde{\hat{K}}_{\ell}\})= & -\frac{T}{2}\tr t^{\top}t+\tr t^{\top}Y+\frac{1}{2}\sum_{\ell=1}^{L-1}\tr\tilde{\hat{K}}_{\ell}^{\top}\hat{K}_{\ell}-\log\int dP_{0}(a)\int d\mathcal{N}(\hat{z}^{L}\,|\,0,\sigma_{L}^{2}\hat{K}_{L-1})\,e^{a^{\top}t^{\top}z^{L}}\nonumber \\
 & -\sum_{\ell=1}^{L-1}\log\int d\mathcal{N}(\hat{z}^{\ell}\,|\,0,\sigma_{\ell}^{2}\hat{K}_{\ell-1})\,e^{\frac{1}{2}\hat{z}^{\ell\top}\tilde{\hat{K}}_{\ell}\hat{z}^{\ell}}.
\end{align}
In the last layer the test preactivations can be marginalized, $\int d\mathcal{N}(\hat{z}^{L}\,|\,0,\sigma_{L}^{2}\hat{K}_{L-1})\,e^{a^{\top}t^{\top}z^{L}}=\int d\mathcal{N}(z^{L}\,|\,0,\sigma_{L}^{2}K_{L-1})\,e^{a^{\top}t^{\top}z^{L}}$.
Thus, the $P\times P_{*}$ training-test block $k_{L-1}$ and $P_{*}\times P_{*}$
test-test block $\kappa_{L-1}$ of $\hat{K}_{L-1}$ appear in the
energy $E(t,\{\hat{K}_{\ell},\tilde{\hat{K}}_{\ell}\})$ only in the
term $\tr\tilde{\hat{K}}_{L-1}^{\top}\hat{K}_{L-1}$. Accordingly,
the saddle-point equation for their conjugate kernels is $\tilde{k}_{L-1}=0$
and $\tilde{\kappa}_{L-1}=0$. This implies that the test preactivations
can be marginalized in the next layer, $\int d\mathcal{N}(\hat{z}^{L-1}\,|\,0,\sigma_{L-1}^{2}\hat{K}_{L-2})\,e^{\frac{1}{2}z^{L-1\top}\tilde{K}_{L-1}z^{L-1}}=\int d\mathcal{N}(z^{L-1}\,|\,0,\sigma_{L-1}^{2}K_{L-2})\,e^{\frac{1}{2}z^{L-1\top}\tilde{K}_{L-1}z^{L-1}}$
such that $k_{L-2}$ and $\kappa_{L-2}$ only appear in $\tr\tilde{\hat{K}}_{L-2}^{\top}\hat{K}_{L-2}$
and thus the saddle-point equation for their conjugate kernels is
$\tilde{k}_{L-2}=\tilde{\kappa}_{L-2}=0$. Iterating the argument
leads to the saddle-point equations $\tilde{k}_{\ell}=\tilde{\kappa}_{\ell}=0$
for $\ell=1,\dots,L-1$. For $\tilde{K}_{\ell}$, $\ell=1,\dots,L-1$,
the saddle-point equations remain identical to the training posterior.

The saddle-point equations for the kernels are $\hat{K}_{\ell}=\langle\hat{z}^{\ell}\hat{z}^{\ell\top}\rangle_{\hat{z}^{\ell}}$,
$\ell=1,\dots,L-1$. At $\tilde{k}_{\ell}=\tilde{\kappa}_{\ell}=0$,
the expectation is taken w.r.t.~the distribution $P(\hat{z}^{\ell})\propto\mathcal{N}(\hat{z}^{\ell}\,|\,0,\sigma_{\ell}^{2}\hat{K}_{\ell-1})\,e^{\frac{1}{2}z^{\ell\top}\tilde{K}_{\ell}z^{\ell}}$.
Since the second factor does not depend on the test preactivations
$z^{\ell,*}$, the conditional distribution of the test preactivations
given the training preactivations is identical to the prior distribution:
\begin{equation}
P(z^{\ell,*}\,|\,z^{\ell})=P_{0}(z^{\ell,*}\,|\,z^{\ell})=\mathcal{N}(z^{\ell,*}\,|\,k_{\ell-1}^{\top}K_{\ell-1}^{-1}z^{\ell},\sigma_{\ell}^{2}\kappa_{\ell-1}-\sigma_{\ell}^{2}k_{\ell-1}^{\top}K_{\ell-1}^{-1}k_{\ell-1}).\label{eq:zstar_cond_z_lin}
\end{equation}
The joint distribution of training and test preactivations is $P(\hat{z}^{\ell})=P_{0}(z^{\ell,*}\,|\,z^{\ell})P(z^{\ell})$
where $P(z^{\ell})$ is the single neuron posterior of the training
preactivations. Thus, the saddle-point equations for the training
kernels $K_{\ell}$ remain unchanged. For the train-test and the test-test
kernels, the saddle-point equations are
\begin{equation}
k_{\ell}=\langle z^{\ell}z^{\ell\top}\rangle_{z^{\ell}}K_{\ell-1}^{-1}k_{\ell-1}\label{eq:k_ell_linear}
\end{equation}
\begin{equation}
\kappa_{\ell}=k_{\ell-1}^{\top}K_{\ell-1}^{-1}\langle z^{\ell}z^{\ell\top}\rangle_{z^{\ell}}K_{\ell-1}^{-1}k_{\ell-1}+\sigma_{\ell}^{2}\kappa_{\ell-1}-\sigma_{\ell}^{2}k_{\ell-1}^{\top}K_{\ell-1}^{-1}k_{\ell-1}\label{eq:kappa_ell_linear}
\end{equation}
with $K_{\ell}=\langle z^{\ell}z^{\ell\top}\rangle_{z^{\ell}}$. In
the last layer, the conditional distribution of the test preactivations
given the training preactivations is also identical to the prior and
thus $P(a,\hat{z}^{L})=P_{0}(z^{L,*}\,|\,z^{L})P(a,z^{L})$. In total,
we arrive at
\begin{equation}
P(A,\hat{Z}_{L},\dots,\hat{Z}_{1})=\prod_{i=1}^{N}[P_{0}(z_{i}^{L,*}\,|\,z_{i}^{L})P(z_{i}^{L}\,|\,a_{i})P(a_{i})]\prod_{\ell=1}^{L-1}[\prod_{j=1}^{N}[P_{0}(z_{j}^{\ell,*}\,|\,z_{j}^{\ell})P(z_{j}^{\ell})]]
\end{equation}
for the test posterior.

For the mean predictor we consider a single test input $x$ and the
corresponding preactivations, leading with (\ref{eq:zstar_cond_z_lin}),
$P(z^{L}\,|\,a)=\mathcal{N}(z^{L}\,|\,YU^{-1}a,\sigma_{L}^{2}K_{L-1})$,
and (\ref{eq:a_posterior_lin}) to $f(x)=\langle a\langle\langle z^{L,*}\rangle_{z^{L,*}|z^{L}}\rangle_{z^{L}|a}\rangle_{a}=Y^{\top}K_{L-1}^{-1}k_{L-1}$.
The saddle-point equation (\ref{eq:k_ell_linear}) implies $K_{\ell}^{-1}k_{\ell}=K_{\ell-1}^{-1}k_{\ell-1}$
and thus by iteration
\begin{equation}
f_{r}(x)=k_{0}(x)^{\top}K_{0}^{-1}y_{r}.
\end{equation}

The test kernel is determined by (\ref{eq:kappa_ell_linear}) for
$\ell=1,\dots,L-1$ which evaluates with $K_{\ell}^{-1}k_{\ell}=K_{\ell-1}^{-1}k_{\ell-1}$
and $K_{\ell}=\sigma_{\ell}^{2}K_{\ell-1}+\frac{\sigma_{a}^{2(L-\ell)}}{\prod_{\ell'=\ell+1}^{L}\sigma_{\ell'}^{2}}YU^{-(L-\ell+1)}Y^{\top}$
to $\kappa_{\ell}=\sigma_{\ell}^{2}\kappa_{\ell-1}+\frac{\sigma_{a}^{2(L-\ell)}}{\prod_{\ell'=\ell+1}^{L}\sigma_{\ell'}^{2}}k_{0}^{\top}K_{0}^{-1}YU^{-(L-\ell+1)}Y^{\top}K_{0}^{-1}k_{0}$.
Thus $K_{\ell}$ and $\kappa_{\ell}$ obey the same recursion except
that $Y$ is replaced by $Y^{\top}K_{L-1}^{-1}k_{L-1}$, yielding
\begin{equation}
\kappa_{\ell}=\prod_{\ell'=1}^{\ell}[\sigma_{\ell'}^{2}]\kappa_{0}+\frac{\sigma_{a}^{2(L-\ell)}}{\prod_{\ell'=\ell+1}^{L}\sigma_{\ell'}^{2}}k_{0}^{\top}K_{0}^{-1}YU^{-(L-\ell+1)}Y^{\top}K_{0}^{-1}k_{0},\quad2\le\ell\le L-1.
\end{equation}
In the last layer, (\ref{eq:zstar_cond_z_lin}) leads to
\begin{equation}
\kappa_{L}=\prod_{\ell=1}^{L}[\sigma_{\ell}^{2}]\kappa_{0}+k_{0}^{\top}K_{0}^{-1}YU^{-1}Y^{\top}K_{0}^{-1}k_{0}.
\end{equation}
Evaluated for two test inputs $x_{1},x_{2}$ and for $\sigma_{\ell}=1$
leads to the kernel stated in the main text.

\fbox{\begin{minipage}[t]{1\columnwidth - 2\fboxsep - 2\fboxrule}%
\textbf{Summary:} Test preactivation posterior on test inputs $x_{*}$
\begin{equation}
P(z^{\ell,*}\,|\,z^{\ell})=\mathcal{N}(z^{\ell,*}\,|\,k_{0}^{\top}K_{0}^{-1}z^{\ell},\prod_{\ell'=1}^{\ell}[\sigma_{\ell'}^{2}][\kappa_{0}-k_{0}^{\top}K_{0}^{-1}k_{0}]),\quad\ell=1,\dots,L,
\end{equation}
with
\begin{equation}
\kappa_{\ell}=\prod_{\ell'=1}^{\ell}[\sigma_{\ell'}^{2}]\kappa_{0}+\frac{\sigma_{a}^{2(L-\ell)}}{\prod_{\ell'=\ell+1}^{L}\sigma_{\ell'}^{2}}k_{0}^{\top}K_{0}^{-1}YU^{-(L-\ell+1)}Y^{\top}K_{0}^{-1}k_{0},\quad\ell=1,\dots,L-1,
\end{equation}
\begin{equation}
\kappa_{L}=\prod_{\ell=1}^{L}[\sigma_{\ell}^{2}]\kappa_{0}+k_{0}^{\top}K_{0}^{-1}YU^{-1}Y^{\top}K_{0}^{-1}k_{0},
\end{equation}
and $K_{0}=\frac{1}{N_{0}}X^{\top}X$, $k_{0}=\frac{1}{N_{0}}X^{\top}x_{*}$,
$\kappa_{0}=\frac{1}{N_{0}}x_{*}^{\top}x_{*}$.

Predictor on test input $x$
\begin{equation}
f(x)=Y^{\top}K_{0}^{-1}k_{0}(x)
\end{equation}
with $K_{0}=\frac{1}{N_{0}}X^{\top}X$ and $k_{0}(x)=\frac{1}{N_{0}}X^{\top}x$.%
\end{minipage}}

\subsection{Consistency Check}

The above derivation relies on a high dimensional saddle point approximation
\citep{fedoryuk1977saddle,Fedoryuk1989}. This is an ad-hoc approximation
without further assumptions about the structure of the integral \citep{33dbad4f-5b73-3bc9-ac01-477e2b81fa2a,doi:10.1137/22M1495688}.
For the linear case the high dimensional saddle point approximation
can be circumvented using the methods from \citep{Li21_031059}. Here,
we show that this derivation leads to consistent results with the
high dimensional saddle point approximation for non-lazy networks.

\subsubsection{Partition Function}

We start with the self consistency equation for the readout covariance
$U$. To this end we consider the partition function
\begin{equation}
Z=\int dP_{0}(\Theta)\,\mathcal{N}\big(Y\,|\,f(X;\Theta),\frac{T}{N}I_{m}\otimes I_{P}\big)
\end{equation}
where we already rescaled $T\to T/N$, introduced the $m\times P$
matrix of training targets $Y$ and the $N_{0}\times P$ matrix of
training inputs $X$, and neglected irrelevant multiplicative factors
(which will be done throughout) to rewrite the contribution from the
loss as a matrix valued normal distribution (see, e.g., \citep{Gupta1999})
to ease the notation. Marginalizing the readout weights and changing
variables from the $N\times N$ (or $N\times N_{0}$) weights $W_{\ell}$
to the $N\times P$ preactivations $Z_{\ell}=\frac{1}{\sqrt{N}}W_{\ell}Z_{\ell-1}$
and $Z_{1}=\frac{1}{\sqrt{N_{0}}}W_{1}X$ leads to
\begin{equation}
Z=\int\prod_{\ell=1}^{L}dP_{0}(Z_{\ell})\,\mathcal{N}\big(Y\,|\,0,I_{m}\otimes\big[\frac{T}{N}I_{P}+\frac{\sigma_{a}^{2}}{N}K_{L}\big]\big)
\end{equation}
with $P_{0}(Z_{\ell})=\mathcal{N}(Z_{\ell}\,|\,0,I_{N}\otimes\sigma_{\ell}^{2}K_{\ell-1})$
and $K_{\ell}=\frac{1}{N}Z_{\ell}^{\top}Z_{\ell}$.

Next we marginalize the $Z_{\ell}$ layer by layer, starting from
the last layer. Using (\ref{eq:icf_Gauss}) makes take the $P_{0}(Z_{L})$
expectation tractable,
\begin{equation}
\langle e^{-\frac{\sigma_{a}^{2}}{2N^{2}}\tr t^{\top}Z_{L}^{\top}Z_{L}t}\rangle=e^{-\frac{N}{2}\log\det(I_{m}+\frac{\sigma_{a}^{2}\sigma_{L}^{2}}{N^{2}}t^{\top}K_{L-1}t)}.
\end{equation}
Introducing the $m\times m$ matrix $H_{L}=\frac{\sigma_{a}^{2}\sigma_{L}^{2}}{N^{2}}t^{\top}K_{L-1}t$
with conjugate variable $U_{L}$ leads to 
\begin{equation}
Z=\int dU_{L}\,e^{\frac{N}{2}\log\det U_{L}-\frac{N}{2}\tr U_{L}}\int\prod_{\ell=1}^{L-1}dP_{0}(Z_{\ell})\,\mathcal{N}\big(Y\,|\,0,\frac{T}{N}I_{m}\otimes I_{P}+\frac{\sigma_{a}^{2}}{N}\sigma_{L}^{2}U_{L}\otimes K_{L-1}\big)
\end{equation}
where $H_{L}=U_{L}^{-1}-I_{m}$ from a saddle-point approximation
at large $N$ was used.

Repeating these steps layer by layer leads to
\begin{equation}
Z=\int\prod_{\ell=1}^{L}dU_{\ell}\,e^{-\frac{N}{2}\sum_{\ell=1}^{L}[\tr U_{\ell}-\log\det U_{\ell}]}\,\mathcal{N}\big(Y\,|\,0,\frac{T}{N}I_{m}\otimes I_{P}+\frac{\sigma_{a}^{2}}{N}\prod_{\ell=1}^{L}\sigma_{\ell}^{2}U_{\ell}\otimes K_{0}\big)
\end{equation}
with $K_{0}=\frac{1}{N_{0}}X^{\top}X$. Note that without the rescaling
of temperature the contribution $\frac{\sigma_{a}^{2}}{N}\prod_{\ell=1}^{L}\sigma_{\ell}^{2}U_{\ell}\otimes K_{0}$
due to learning would be a $O(1/N)$ correction. The remaining $U_{\ell}$
integrals can be solved in a low dimensional saddle-point approximation;
at zero temperature the corresponding saddle-point equations are
\begin{equation}
I_{m}=U_{\ell}+\alpha I_{m}-\frac{1}{\sigma_{a}^{2}\prod_{\ell=1}^{L}\sigma_{\ell}^{2}}\big(\prod_{\ell'=1}^{\ell}U_{\ell'}\big)^{-1}YK_{0}^{-1}Y^{\top}\big(\prod_{\ell'=\ell+1}^{L}U_{\ell'}\big)^{-1}.
\end{equation}
Since the equations are identical for every $\ell$, we can set $\sigma_{a}^{2}U_{\ell}=U$,
leading to
\begin{equation}
U=\sigma_{a}^{2}I_{m}-\alpha\sigma_{a}^{2}I_{m}+\frac{\sigma_{a}^{2L}}{\prod_{\ell=1}^{L}\sigma_{\ell}^{2}}U^{-L}YK_{0}^{-1}Y^{\top}.
\end{equation}
In (\ref{eq:U_self_consistency}) the second term proportional to
$\alpha$ is missing, indicating that the high dimensional saddle
point approximation is in general not valid. However, the scaling
analysis of $U$ still applies, requiring $\sigma_{a}^{2}=O(P^{-1/L})$
to ensure $U=O(1)$. In this case the first and second term can be
neglected for large $P$ and we arrive at $U^{L+1}=\frac{\sigma_{a}^{2L}}{\prod_{\ell=1}^{L}\sigma_{\ell}^{2}}YK_{0}^{-1}Y^{\top}$,
consistent with the derivation based on the high dimensional saddle
point approximation.

\subsubsection{Readout Weight Posterior}

Next, we show that the readout weight posterior is consistent. We
consider the neurons averaged readout posterior 
\begin{equation}
P(a)=\frac{1}{N}\sum_{i=1}^{N}\langle\delta(a-a_{i})\rangle_{\Theta}
\end{equation}
where $a_{i}$ denotes the $m$ readout weights of an arbitrary neuron
$i$. Note that due to the permutation symmetry of the weight posterior,
$P(a)$ is identical to the single neuron posterior $\langle\delta(a-a_{i})\rangle_{\Theta}$.
We compute $P(a)$ from the generating functional
\begin{equation}
Z[j]=\int dP_{0}(\Theta)\,\exp\big(-N\beta\mathcal{L}(\Theta)+\sum_{i=1}^{N}j(a_{i})\big)
\end{equation}
via $P(a)=\frac{1}{N}\frac{\delta}{\delta j(a)}\log Z[j]\big|_{j=0}$.
Changing variables from $W_{\ell}$ to $Z_{\ell}$ and using the matrix
valued normal for convenience leads to
\begin{equation}
Z[j]=\int dP_{0}(A)\prod_{\ell=1}^{L}dP_{0}(Z_{\ell})\,\mathcal{N}\big(Y\,|\,\frac{1}{N}A^{\top}Z_{L},\frac{T}{N}I_{m}\otimes I_{P}\big)e^{\sum_{i=1}^{N}j(a_{i})}
\end{equation}
where $A$ denotes the $N\times m$ matrix of all readout weights.

Again, the expectations $Z_{\ell}$ are calculated layer by layer,
starting with $Z_{L}$. Using (\ref{eq:icf_Gauss}) the $Z_{L}$ expectation
is tractable,
\begin{equation}
\langle e^{\frac{i}{N}\tr tA^{\top}Z_{L}}\rangle=e^{-\frac{\sigma_{L}^{2}}{2N^{2}}\tr At^{\top}K_{L-1}tA^{\top}},
\end{equation}
which leads to
\begin{equation}
Z[j]=\int dP_{0}(A)\prod_{\ell=1}^{L-1}dP_{0}(Z_{\ell})\,\mathcal{N}\big(Y\,|\,0,\frac{T}{N}I_{m}\otimes I_{P}+\frac{\sigma_{L}^{2}}{N^{2}}A^{\top}A\otimes K_{L-1}\big)e^{\sum_{i=1}^{N}j(a_{i})}.
\end{equation}
Introducing the $m\times m$ order parameter $\sigma_{a}^{2}U_{L}=\frac{1}{N}A^{\top}A$
with conjugate variable $H_{L}$ and the single neuron partition function
\begin{equation}
\hat{Z}[j;H_{L}]=\int dP_{0}(a)\,e^{-\frac{1}{2}a^{\top}H_{L}a+j(a)}
\end{equation}
leads to
\begin{equation}
Z[j]=\int dU_{L}dH_{L}\,e^{\frac{N\sigma_{a}^{2}}{2}\tr U_{L}H_{L}+N\log\hat{Z}[j;H_{L}]}\int\prod_{\ell=1}^{L-1}dP_{0}(Z_{\ell})\,\mathcal{N}\big(Y\,|\,0,\frac{T}{N}I_{m}\otimes I_{P}+\frac{\sigma_{a}^{2}}{N}\sigma_{L}^{2}U_{L}\otimes K_{L-1}\big).
\end{equation}
The structure for the remaining $Z_{\ell}$ expectations is identical
to the partition function.

Integrating out the remaining $Z_{\ell}$ layer by layer yields
\begin{equation}
Z[j]=\int dH_{L}\prod_{\ell=1}^{L}dU_{\ell}\,e^{\frac{N\sigma_{a}^{2}}{2}\tr U_{L}H_{L}-\frac{N}{2}\sum_{\ell=1}^{L-1}[\tr U_{\ell}-\log\det U_{\ell}]+N\log\hat{Z}[j;H_{L}]}\,\mathcal{N}\big(Y\,|\,0,\frac{T}{N}I_{m}\otimes I_{P}+\frac{\sigma_{a}^{2}}{N}\prod_{\ell=1}^{L}\sigma_{\ell}^{2}U_{\ell}\otimes K_{0}\big).
\end{equation}
Taking the functional derivative and evaluating at $j=0$ leads to
\begin{equation}
P(a)=\mathcal{N}(a\,|\,0,\sigma_{a}^{2}U_{L})
\end{equation}
where $\log\hat{Z}[0;H_{L}]=-\frac{1}{2}\log\det(I_{m}+\sigma_{a}^{2}H_{L})$
and the corresponding saddle point equation $H_{L}=\sigma_{a}^{-2}(U_{L}^{-1}-I_{m})$
were used. The saddle point equations for $U_{\ell}$ are unchanged
and we arrive at $P(a)=\mathcal{N}(a\,|\,0,U)$ with $U=\sigma_{a}^{2}U_{L}$
determined by the self consistency equation for $U$.

\subsubsection{Predictor}

Finally, we compute the predictor statistics. To this end we consider
the generating function
\begin{equation}
Z(j)=\int dP_{0}(\Theta)\,\exp\big(-N\beta\mathcal{L}(\Theta)+ij^{\top}f(x;\Theta)\big)
\end{equation}
where $x$ denotes an arbitrary test input and $j$ is an $m$-dimensional
vector. Changing variables from $W_{\ell}$ to the $N\times(P+1)$
matrix $\hat{Z}_{\ell}$ of preactivations on the training inputs
and the test input $x$, using (\ref{eq:icf_Gauss}), and marginalizing
the readout weights leads to
\begin{align}
Z(j) & =\int dt\,\int\prod_{\ell=1}^{L}dP_{0}(\hat{Z}_{\ell})\,\exp\Big(-i\mathrm{tr}\,tY-\frac{T}{2N}\mathrm{tr}\,t^{\top}t-\frac{\sigma_{a}^{2}}{2N}\mathrm{tr}\,\hat{t}^{\top}\hat{Z}_{L}\hat{t}\Big)
\end{align}
where $\hat{t}$ is the $(P+1)\times m$ matrix obtained by by concatenating
$t$ and $j,$ $\hat{K}_{\ell}=\frac{1}{N}\hat{Z}_{\ell}^{\top}\hat{Z}_{\ell}$
is the $(P+1)\times(P+1)$ dimensional kernel on the training and
test input, and $P_{0}(\hat{Z}_{\ell})=\mathcal{N}(\hat{Z}_{\ell}\,|\,0,I_{N}\otimes\sigma_{\ell}^{2}\hat{K}_{\ell-1})$.
Marginalizing the $\hat{Z}_{\ell}$ can be done in analogy to the
partition function, leading to
\begin{equation}
Z(j)=\int\prod_{\ell=1}^{L}dU_{\ell}\,e^{-\frac{N}{2}\sum_{\ell=1}^{L}[\mathrm{tr}\,U_{\ell}-\log\det U_{\ell}]}\int dt\,\exp\Big(-i\mathrm{tr}\,tY-\frac{T}{2N}\mathrm{tr}\,t^{\top}t-\frac{\sigma_{a}^{2}}{2N}\mathrm{tr}\,\prod_{\ell=1}^{L}(\sigma_{\ell}^{2}U_{\ell})\hat{t}^{\top}\hat{K}_{0}\hat{t}\Big).
\end{equation}
We separate $j$ again,
\begin{equation}
\hat{t}^{\top}\hat{K}_{0}\hat{t}=t^{\top}K_{0}t+2t^{\top}k_{0}(x)j^{\top}+\kappa_{0}(x)jj^{\top}
\end{equation}
leading to (at $T=0$ for simplicity) 
\begin{equation}
Z(j)=e^{-\frac{NL}{2}[\mathrm{tr}\,U-\ln\det U]-\frac{\sigma_{a}^{2}\prod_{\ell=1}^{L}(\sigma_{\ell}^{2})}{2N}\kappa_{0}(x)j^{\top}U^{L}j}\mathcal{N}\big(Y\,|\,\frac{\sigma_{a}^{2}\prod_{\ell=1}^{L}(\sigma_{\ell}^{2})}{N}k_{0}(x)ij^{\top}U^{L},\frac{\sigma_{a}^{2}\prod_{\ell=1}^{L}(\sigma_{\ell}^{2})}{N}U^{L}\otimes K_{0}\big)
\end{equation}
after solving the $t$ integral and performing the saddle-point approximation
of the $U_{\ell}$ integrals.

The statistics of $f(x,\Theta)$ follow from the derivatives of $\log Z(j)$
evaluated at $j=0$. Thus, we only need the solution of the saddle-point
equations at $j=0$. Computing the derivatives at $j=0$ leads to
\begin{equation}
\langle f(x)\rangle=k_{0}(x)^{\top}K_{0}^{-1}Y,\label{eq:mean_predictor-1}
\end{equation}
\begin{equation}
\langle\delta f(x)\delta f(x)^{\top}\rangle=\frac{\sigma_{a}^{2}\prod_{\ell=1}^{L}(\sigma_{\ell}^{2})}{N}U^{L}\big(\kappa_{0}(x)-k_{0}(x)^{\top}K_{0}^{-1}k_{0}(x)\big).\label{eq:variance_predictor-1}
\end{equation}
All higher cumulants vanish because $\log Z(j)$ is quadratic in $j$.

\textbf{Scaling Analysis:} If $\sigma_{a}=O(1)$ the variance is $O(P^{L/(L+1)}/N)$;
if $\sigma_{a}$ is scaled to keep the readout norm $O(1)$ the variance
is $O(P^{(L-1)/L}/N)$. In all cases, the variance is suppressed compared
to the lazy case and the effect diminishes with increasing number
of layers. The strongest reduction occurs for single hidden layer
networks with scaled $\sigma_{a}$ where the predictor variance is
$O(1/N)$. Thus, a benefit of feature learning is that it suppresses
the variability of the predictor. Accordingly, the per sample generalization
error $\epsilon_{g}(x)=[y_{r}-f_{r}(x)]^{2}+\langle\delta f_{r}(x;\Theta)^{2}\rangle_{\Theta}$
is dominated by the $O(1)$ bias contribution $[y_{r}-f_{r}(x)]^{2}$,
particularly in shallow networks with scaled $\sigma_{a}$.

\subsubsection{Training Kernel}

We first consider the mean training kernel in the last layer $K_{L}=\frac{1}{N}\langle Z_{L}^{\top}Z_{L}\rangle_{\Theta}$.
We marginalize $A$ following the steps used for the partition function,
leading to
\begin{equation}
K_{L}=\frac{1}{N}\int\prod_{\ell=1}^{L}[dP_{0}(Z_{\ell})]\,\mathcal{N}\big(Y\,|\,0,I_{m}\otimes\big[\frac{T}{N}I_{P}+\frac{\sigma_{a}^{2}}{N^{2}}Z_{L}^{\top}Z_{L}\big]\big)Z_{L}^{\top}Z_{L}
\end{equation}

The presence of $\frac{1}{N}Z_{L}^{\top}Z_{L}$ modifies the relevant
part for the marginalization of $Z_{L}$ to
\begin{equation}
\int dP_{0}(Z_{L})\,\exp\Big(-\frac{\sigma_{a}^{2}}{2N^{2}}\mathrm{tr}\,t^{\top}Z_{L}^{\top}Z_{L}t\Big)\frac{1}{N}Z_{L}^{\top}Z_{L}=e^{-\frac{N}{2}\log\det(I_{m}+\frac{\sigma_{a}^{2}\sigma_{L}^{2}}{N^{2}}t^{\top}K_{L-1}t)}A(t),
\end{equation}
\begin{equation}
A(t)=\frac{\int dP_{0}(z_{L})\,z_{L}z_{L}^{\top}\exp\Big(-\frac{\sigma_{a}^{2}}{2N^{2}}\|t^{\top}z_{L}\|^{2}\Big)}{\int dP_{0}(z_{L})\,\exp\Big(-\frac{\sigma_{a}^{2}}{2N^{2}}\|t^{\top}z_{L}\|^{2}\Big)}.
\end{equation}
$A(t)$ is the second moment of a zero-mean Gaussian with covariance
$\big((\sigma_{L}^{2}K_{L-1})^{-1}+\frac{\sigma_{a}^{2}}{N^{2}}tt^{\top}\big)^{-1}=\sigma_{L}^{2}K_{L-1}-\sigma_{L}^{2}K_{L-1}\frac{\sigma_{a}^{2}}{N^{2}}t\big(I+H_{L}\big)^{-1}t^{\top}\sigma_{L}^{2}K_{L-1}$,
thus 
\begin{equation}
A(t)=\sigma_{L}^{2}\Big(K_{L-1}-\frac{\sigma_{a}^{2}\sigma_{L}^{2}}{N^{2}}K_{L-1}tU_{L}t^{\top}K_{L-1}\Big)
\end{equation}
where we used $H_{L}=U_{L}^{-1}-I_{m}$. Since
\begin{align*}
\frac{\sigma_{a}^{2}\sigma_{L}^{2}}{N^{2}}\int dt\,tU_{L}t^{\top} & e^{-i\mathrm{tr}\,t^{\top}Y-\frac{\sigma_{a}^{2}\sigma_{L}^{2}}{2N}\mathrm{tr}\,U_{L}t^{\top}K_{L-1}t}\\
 & =-\frac{2}{N}\frac{\partial}{\partial K_{L-1}}\mathcal{N}\big(Y\,|\,0,\sigma_{a}^{2}\sigma_{L}^{2}N^{-1}U_{L}\otimes K_{L-1}\big)\\
 & =-\frac{1}{\sigma_{a}^{2}\sigma_{L}^{2}}(K_{L-1})^{-1}YU_{L}^{-1}Y^{\top}(K_{L-1})^{-1}\mathcal{N}\big(Y\,|\,0,\sigma_{a}^{2}\sigma_{L}^{2}N^{-1}U_{L}\otimes K_{L-1}\big)
\end{align*}
where we neglect a $O(m/N)$ term in the last step, we obtain
\begin{equation}
\langle K_{L}\rangle=\sigma_{L}^{2}\langle K_{L-1}\rangle+\frac{1}{\sigma_{a}^{2}}YU_{L}^{-1}Y^{\top}.
\end{equation}
Using the same steps for the next layer leads to
\begin{equation}
\langle K_{L}\rangle=\sigma_{L}^{2}\sigma_{L-1}^{2}\langle K_{L-2}\rangle+\frac{1}{\sigma_{a}^{2}}Y(U_{L-1}U_{L})^{-1}Y^{\top}+\frac{1}{\sigma_{a}^{2}}YU_{L}^{-1}Y^{\top}.
\end{equation}
Iterating this to the first layer and using $\sigma_{a}^{2}U_{\ell}=U$
at the saddle-point we obtain
\begin{equation}
\langle K_{L}\rangle=\prod_{\ell=1}^{L}(\sigma_{\ell}^{2})K_{0}+Y(\sum_{\ell=1}^{L}\sigma_{a}^{2(\ell-1)}U^{-\ell})Y^{\top}.
\end{equation}
Due to the scaling $\sigma_{a}^{2}=O(1/P^{1/L})$ only the first term
in the sum is relevant, $\sum_{\ell=1}^{L}\sigma_{a}^{2(\ell-1)}U^{-\ell}=U^{-1}+O(1/P^{1/L})$.
Alternatively, the sum can be performed using the geometric series.

Using the same steps we obtain for $\ell<L$
\begin{equation}
\langle K_{\ell}\rangle=\prod_{\ell'=1}^{\ell}(\sigma_{\ell'}^{2})K_{0}+\frac{1}{\prod_{\ell'=\ell+1}^{L}(\sigma_{\ell'}^{2})}Y(\sum_{\ell'=L-\ell+1}^{L}\sigma_{a}^{2(\ell'-1)}U^{-\ell'})Y^{\top}
\end{equation}
where $(M_{\ell})_{ss'}=\delta_{ss'}\sum_{\ell'=L-\ell+1}^{L}u_{s}^{-\ell'}=\delta_{ss'}\frac{1}{u_{s}^{L}}\frac{1-u_{s}^{\ell}}{1-u_{s}}$
from the geometric series. Again only the first term in the sum is
relevant, $\sum_{\ell'=L-\ell+1}^{L}\sigma_{a}^{2(\ell'-1)}U^{-\ell'}=\sigma_{a}^{2(L-\ell)}U^{-L+\ell-1}$,
yielding once again consistent results with the high dimensional saddle
point approximation.

\section{Sigmoidal Networks}

\subsection{Training Posterior}

As in the linear case, we change variables from the weights $W_{\ell}$
to the preactivations $Z_{\ell}$. Repeating the derivation for the
linear network leads to

\begin{equation}
P(A,Z_{L},\dots,Z_{1})\propto\int dt\prod_{\ell=1}^{L-1}[dK_{\ell}d\tilde{K}_{\ell}]\,e^{-NE(t,\{K_{\ell},\tilde{K}_{\ell}\})}\prod_{i=1}^{N}[P(a_{i},z_{i}^{L})]\prod_{\ell=1}^{L-1}[\prod_{j=1}^{N}P(z_{j}^{\ell})],
\end{equation}
\begin{align}
E(t,\{K_{\ell},\tilde{K}_{\ell}\})= & -\frac{T}{2}\tr t^{\top}t+\tr t^{\top}Y+\frac{1}{2}\sum_{\ell=1}^{L-1}\tr\tilde{K}_{\ell}^{\top}K_{\ell}-\log\int dP_{0}(a)d\mathcal{N}(z^{L}\,|\,0,\sigma_{L}^{2}K_{L-1})e^{a^{\top}t^{\top}\phi(z^{L})}\nonumber \\
 & -\sum_{\ell=1}^{L-1}\log\int d\mathcal{N}(z^{\ell}\,|\,0,\sigma_{\ell}^{2}K_{\ell-1})e^{\frac{1}{2}\phi(z^{\ell})^{\top}\tilde{K}_{\ell}\phi(z^{\ell})},
\end{align}
where the single neuron posteriors are not Gaussian,
\begin{equation}
P(a,z^{L})\propto P_{0}(a)\mathcal{N}(z^{L}\,|\,0,\sigma_{L}^{2}K_{L-1})e^{a^{\top}t^{\top}\phi(z^{L})},
\end{equation}
\begin{equation}
P(z^{\ell})\propto\mathcal{N}(z^{\ell}\,|\,0,\sigma_{\ell}^{2}K_{\ell-1})e^{\frac{1}{2}\phi(z^{\ell})^{\top}\tilde{K}_{\ell}\phi(z^{\ell})}.
\end{equation}
The saddle point approximation of the $t$ and $K_{\ell}$, $\tilde{K}_{\ell}$
integrals leads to
\begin{equation}
P(A,Z_{L},\dots,Z_{1})=\prod_{i=1}^{N}[P(a_{i})P(z_{i}^{L}\,|\,a_{i})]\prod_{\ell=1}^{L-1}[\prod_{j=1}^{N}P(z_{j}^{\ell})],
\end{equation}
the corresponding saddle point equations are
\begin{equation}
y_{\mu}^{r}=\langle a_{r}\phi(z_{\mu}^{L})\rangle_{a,z^{L}}+Tt_{\mu}^{r},
\end{equation}
\begin{equation}
K_{\ell}=\langle\phi(z^{\ell})\phi(z^{\ell})^{\top}\rangle_{z^{\ell}},\quad\ell=1,\dots,L-1,
\end{equation}
\begin{equation}
\tilde{K}_{\ell-1}=\frac{1}{\sigma_{\ell}^{2}}K_{\ell-1}^{-1}\langle z^{\ell}z^{\ell\top}\rangle_{z^{\ell}}K_{\ell-1}^{-1}-K_{\ell-1}^{-1},\quad\ell=2,\dots,L.
\end{equation}
Note that $K_{\ell}$ depends on the activation kernel and $\tilde{K}_{\ell-1}$
on the preactivation kernel; in the linear case this distinction was
not relevant.

We now set $\sigma_{a}^{2}=1/P$. The readout posterior is
\begin{equation}
\log P(a)=-P\Big(\frac{1}{2}a^{\top}a-\frac{1}{P}\log\int d\mathcal{N}(z^{L}\,|\,0,\sigma_{L}^{2}K_{L-1})e^{a^{\top}t^{\top}\phi(z^{L})}\Big)+\text{const.}\equiv-PE(a\,|\,K_{L-1},t)+\text{const.}
\end{equation}
At large $P$ the posterior concentrates at the minima of $E(a\,|\,K_{L-1},t)$,
\begin{equation}
P(a)=\sum_{\gamma=1}^{n}P_{\gamma}\delta(a-a_{\gamma}),\qquad P_{\gamma}=\frac{e^{-PE(a_{\gamma}\,|\,K_{L-1},t)}}{\sum_{\gamma'=1}^{n}e^{-PE(a_{\gamma'}\,|\,K_{L-1},t)}},\label{eq:P_a_sigmoidal}
\end{equation}
where the $a_{\gamma}$ are determined by
\begin{equation}
a_{\gamma}=\frac{1}{P}t^{\top}\langle\phi(z^{L})\rangle_{z^{L}|a_{\gamma}},\qquad P(z\,|\,a)=\frac{P_{0}(z)e^{(at)^{\top}\phi(z)}}{\int dP_{0}(z)\,e^{(at)^{\top}\phi(z)}}.\label{eq:P_z_a_sigmoidal}
\end{equation}

An interesting aspect is that the weights are given by $P_{\gamma}\propto\exp[-PE(a_{\gamma}\,|\,K_{L-1},t)]$
where the energy $E(a\,|\,K_{L-1},t)$ is $O(1)$. Thus, the energy
needs to be quasi-degenerate between the branches, $E(a_{\gamma}\,|\,K_{L-1},t)-E(a_{\gamma'}\,|\,K_{L-1},t)=O(1/P)$,
to achieve finite weights $P_{\gamma}$. Since this quasi-degeneracy
can occur even though the solutions are not connected by permutation
symmetry, the order parameter $t$ as well as the readouts weights
$a_{\gamma}$ must be precisely fine tuned.

In contrast to the readout posterior, the conditional distribution
\begin{equation}
P(z^{L}\,|\,a)\propto\mathcal{N}(z^{L}\,|\,0,\sigma_{L}^{2}K_{L-1})e^{a^{\top}t^{\top}\phi(z^{L})}
\end{equation}
does not concentrate; all quantities appearing in $P(z^{L}\,|\,a)$
are $O(1)$. Nonetheless, as evidenced by the empirical sampling,
$P(z^{L}\,|\,a)$ can posses a pronounced mean and comparably modest
fluctuations. The simplification of $P(a)$ due to the concentration
makes it advantageous to factorize the joint posterior into the conditional
distributions $P(a,z^{L})=P(z^{L}\,|\,a)P(a)$, which simplifies the
preactivation posterior to $P(z^{L})=\sum_{\gamma=1}^{n}P_{\gamma}P(z^{L}\,|\,a_{\gamma})$,
the saddle point equation for $t$ to $y_{\mu}^{r}=\sum_{\gamma=1}^{n}P_{\gamma}a_{\gamma}^{r}\langle\phi(z_{\mu}^{L})\rangle_{z^{L}|a_{\gamma}}+Tt_{\mu}^{r}$,
and the last layer kernel to $K_{L}=\sum_{\gamma=1}^{n}P_{\gamma}\langle\phi(z^{L})\phi(z^{L})^{\top}\rangle_{z^{L}|a_{\gamma}}$.

\fbox{\begin{minipage}[t]{1\columnwidth - 2\fboxsep - 2\fboxrule}%
\textbf{Summary:} Posterior distributions
\begin{equation}
P(a)=\sum_{\gamma=1}^{n}P_{\gamma}\delta(a-a_{\gamma})
\end{equation}
\begin{equation}
P(z^{L}\,|\,a)\propto\mathcal{N}(z^{L}\,|\,0,\sigma_{L}^{2}K_{L-1})e^{a^{\top}t^{\top}\phi(z^{L})}
\end{equation}
\begin{equation}
P(z^{\ell})\propto\mathcal{N}(z^{\ell}\,|\,0,\sigma_{\ell}^{2}K_{\ell-1})e^{\frac{1}{2}\phi(z^{\ell})^{\top}\tilde{K}_{\ell}\phi(z^{\ell})},\quad\ell=1,\dots,L-1,
\end{equation}
with
\begin{equation}
a_{\gamma}=\frac{1}{P}t^{\top}\langle\phi(z^{L})\rangle_{z^{L}|a_{\gamma}},\quad\gamma=1,\dots,n,
\end{equation}
\begin{equation}
y_{\mu}^{r}=\sum_{\gamma=1}^{n}P_{\gamma}a_{\gamma}^{r}\langle\phi(z_{\mu}^{L})\rangle_{z^{L}|a_{\gamma}}+Tt_{\mu}^{r},\quad r=1,\dots,m,\quad\mu=1,\dots,P
\end{equation}
\begin{equation}
K_{\ell}=\langle\phi(z^{\ell})\phi(z^{\ell})^{\top}\rangle_{z^{\ell}},\quad\ell=1,\dots,L-1,
\end{equation}
\begin{equation}
\tilde{K}_{\ell-1}=\frac{1}{\sigma_{\ell}^{2}}K_{\ell-1}^{-1}\langle z^{\ell}z^{\ell\top}\rangle_{z^{\ell}}K_{\ell-1}^{-1}-K_{\ell-1}^{-1},\quad\ell=2,\dots,L,
\end{equation}
$K_{0}=\frac{1}{N_{0}}X^{\top}X$, and $\sigma_{a}^{2}=1/P$.%
\end{minipage}}

\textbf{Finite $P$ correction:} For finite $P$, we include quadratic
fluctuations around the minima of $E(a\,|\,K_{L-1},t)\equiv E(a)$,
leading to 
\begin{equation}
P(a)=\sum_{\gamma=1}^{n}P_{\gamma}\mathcal{N}(a\,|\,a_{\gamma},\frac{1}{P}E^{\prime\prime}(a_{\gamma})^{-1})
\end{equation}
where $E^{\prime\prime}(a)=I_{m}-\frac{1}{P}t^{\top}\langle(\phi(z^{L})-\langle\phi(z^{L})\rangle_{z|a})(\phi(z^{L})-\langle\phi(z^{L})\rangle_{z|a})^{\top}\rangle_{z|a}t$.
Thus, the sum of Dirac deltas is replaced by a Gaussian mixture.

\textbf{Orthogonal inputs, $L=1$:} For single hidden layer $L=1$
and orthogonal data $K_{0}=I$, the energy simplifies to $E(a\,|\,t)=\frac{1}{2}a^{\top}a-\frac{1}{P}\sum_{\mu=1}^{P}\log\langle e^{\phi(\sigma_{1}\xi)\sum_{r=1}^{m}a_{r}t_{\mu}^{r}}\rangle$
with $\xi\sim\mathcal{N}(0,1)$. Similarly, the conditional distribution
of the preactivations factorizes into $P(z\,|\,a)\propto\prod_{\mu=1}^{P}[\mathcal{N}(z_{\mu}\,|\,0,\sigma_{1}^{2})e^{\phi(z_{\mu})\sum_{r=1}^{m}a_{r}t_{\mu}^{r}}]$.

\textbf{General inputs, $L=1$:} For general $K_{0}$, we use an ad-hoc
saddle point approximation $P(z\,|\,a_{\gamma})\approx\delta(z-z_{\gamma})$
justified by the strong mean apparent in the preactivations. This
leads to $E(a\,|\,t)=\frac{1}{2}a^{\top}a+\frac{1}{P}E(z\,|\,a,t)$
with $E(z\,|\,a,t)=\frac{1}{2\sigma_{1}^{2}}z^{\top}K_{0}^{+}z-(at)^{\top}\phi(z)$
and $z$ determined by the minimum of $E(z\,|\,a,t)$, $K_{0}^{+}z_{\gamma}=\sigma_{1}^{2}\phi^{\prime}(z_{\gamma})ta_{\gamma}$.
It also simplifies the saddle point equations to $y_{r}=\sum_{\gamma=1}^{n}P_{\gamma}a_{\gamma}^{r}\phi(z_{\gamma})+Tt$
and $a_{\gamma}=\frac{1}{P}t^{\top}\phi(z_{\gamma})$.

\textbf{Multimodality of lower layer preactivation posterior:} Since
$\tilde{K}_{\ell}$ is symmetric we can use the eigenvalue decomposition
$\tilde{K}_{\ell}=\frac{1}{P}V_{\ell}\Lambda_{\ell}V_{\ell}^{\top}$
(here we normalize the eigenvectors such that $\frac{1}{P}V_{\ell}^{\top}V_{\ell}=I$)
to decouple $e^{\frac{1}{2}\phi(z^{\ell})^{\top}\tilde{K}_{\ell}\phi(z^{\ell})}=\int d\mathcal{N}(a^{\ell}\,|\,0,\frac{1}{P}I)\,e^{a^{\ell\top}\sqrt{\Lambda_{\ell}}V_{\ell}^{\top}\phi(z^{\ell})}$.
Thus, we have $P(z^{\ell})=\int dP(a^{\ell})\,P(z^{\ell}\,|\,a^{\ell})$
with $P(z^{\ell}\,|\,a^{\ell})\propto\mathcal{N}(z^{\ell}\,|\,0,\sigma_{\ell}^{2}K_{\ell-1})e^{a^{\ell\top}\sqrt{\Lambda_{\ell}}V_{\ell}^{\top}\phi(z^{\ell})}$
and $P(a^{\ell})\propto\mathcal{N}(a^{\ell}\,|\,0,\frac{1}{P}I)\int d\mathcal{N}(z^{\ell}\,|\,0,\sigma_{\ell}^{2}K_{\ell-1})e^{a^{\ell\top}\sqrt{\Lambda_{\ell}}V_{\ell}^{\top}\phi(z^{\ell})}$
which recovers the structure of the last layer posterior. The difference
is that $a^{\ell}$ is $\mathrm{rank}(\tilde{K}_{\ell})$-dimensional
whereas in the last layer $a$ is $m$-dimensional. Empirically $\tilde{K}_{\ell}$
is low rank and the dominating eigenvalues are positive and $O(1)$
such that the discussion of $P(z^{L}\,|\,a)$ and $P(a)$ applies
identically here: $P(a^{\ell})$ concentrates, $P(a^{\ell})=\sum_{\gamma=1}^{n_{\ell}}P_{\gamma}\delta(a^{\ell}-a_{\gamma}^{\ell})$,
leading to a multimodal posterior $P(z^{\ell})=\sum_{\gamma=1}^{n_{\ell}}P_{\gamma}P(z^{\ell}\,|\,a_{\gamma}^{\ell})$.
The difference leading to the frozen code in the last layer but not
in the previous layers is that in the last layer the joint distribution
$P(z^{L}\,|\,a)P(a)$ is sampled, which freezes due to the disconnected
structure of $P(a)$, whereas in the previous layers the marginal
$P(z^{\ell})$ is sampled which is multimodal but not disconnected.

\textbf{Multiple self-consistent solutions:} The energy corresponding
to the order parameters $E(t,\{K_{\ell},\tilde{K}_{\ell}\})$ can
also posses multiple minima, corresponding to different coding schemes.
Taking for simplicity $L=1$ such that $E(t,\{K_{\ell},\tilde{K}_{\ell}\})=E(t)$,
the relative probability of two solutions $t_{1},t_{2}$ is $e^{-N[E(t_{1})-E(t_{2})]}$.
We did not observe degeneracy of $E(t)$ such that the sampled coding
scheme is unique. We show multiple solutions in Fig.~\ref{fig:sigmoidal_code_multisolution}
on the toy task.

\textbf{Weight space dynamics:} Starting from an initial condition
sampled from the prior, both Langevin dynamics and gradient descent
first minimize the error and bring the network close to the solution
space, as in lazy networks \citep{avidan2023connecting}. Already
during this early stage a coding scheme emerges. Subsequently, the
Langevin dynamics stay close to the solution space and are dominated
by the regularization imposed by the prior, which further shapes the
coding scheme and eventually leads to the coding scheme shown in (Fig.~\ref{fig:sigmoidal_code}(B)).
Interestingly, also gradient descent without noise can arrive at the
same solution (see supplement).

\subsection{Test Posterior}

As in the linear case, we change variables from the weights $W_{\ell}$
to the matrix of training and test preactivations $\hat{Z}_{\ell}$.
Repeating the steps for the linear network leads to

\begin{equation}
P(A,\hat{Z}_{L},\dots,\hat{Z}_{1})\propto\int dt\prod_{\ell=1}^{L-1}d\hat{K}_{\ell}d\tilde{\hat{K}}_{\ell}\,e^{-NE(t,\{\hat{K}_{\ell},\tilde{\hat{K}}_{\ell}\})}\prod_{i=1}^{N}[P(a_{i},\hat{z}_{i}^{L})]\prod_{\ell=1}^{L-1}[\prod_{j=1}^{N}P(\hat{z}_{j}^{\ell})],
\end{equation}
\begin{align}
E(t,\{\hat{K}_{\ell},\tilde{\hat{K}}_{\ell}\})= & -\frac{T}{2}\tr t^{\top}t+\tr t^{\top}Y+\frac{1}{2}\sum_{\ell=1}^{L-1}\tr\tilde{\hat{K}}_{\ell}^{\top}\hat{K}_{\ell}-\log\int dP_{0}(a)\int d\mathcal{N}(\hat{z}^{L}\,|\,0,\sigma_{L}^{2}\hat{K}_{L-1})\,e^{a^{\top}t^{\top}\phi(z^{L})}\nonumber \\
 & -\sum_{\ell=1}^{L-1}\log\int d\mathcal{N}(\hat{z}^{\ell}\,|\,0,\sigma_{\ell}^{2}\hat{K}_{\ell-1})\,e^{\frac{1}{2}\tr\phi(\hat{z}^{\ell})^{\top}\tilde{\hat{K}}_{\ell}\phi(\hat{z}^{\ell})},
\end{align}
with the single neuron posteriors
\begin{equation}
P(a,\hat{z}^{L})\propto P_{0}(a)\mathcal{N}(\hat{z}^{L}\,|\,0,\sigma_{L}^{2}\hat{K}_{L-1})e^{a^{\top}t^{\top}\phi(z^{L})}
\end{equation}
\begin{equation}
P(\hat{z}^{\ell})\propto\mathcal{N}(\hat{z}^{\ell}\,|\,0,\sigma_{\ell}^{2}\hat{K}_{\ell-1})e^{\frac{1}{2}\tr\phi(\hat{z}^{\ell})^{\top}\tilde{\hat{K}}_{\ell}\phi(\hat{z}^{\ell})}
\end{equation}
By the same arguments as in the linear case, the saddle-point equations
for the train-test and test-test conjugate kernels are $\tilde{k}_{\ell}=\tilde{\kappa}_{\ell}=0$
for $\ell=1,\dots,L-1$. For $\tilde{K}_{\ell}$, $\ell=1,\dots,L-1$,
the saddle-point equations remain identical to the training case.

The saddle-point equations for the kernels are $\hat{K}_{\ell}=\langle\phi(\hat{z}^{\ell})\phi(\hat{z}^{\ell})^{\top}\rangle_{\hat{z}^{\ell}}$,
$\ell=1,\dots,L-1$. At $\tilde{k}_{\ell}=\tilde{\kappa}_{\ell}=0$,
the conditional distribution of the test preactivations given the
training preactivations is identical to the prior distribution in
all layers:
\begin{equation}
P(z^{\ell,*}\,|\,z^{\ell})=P_{0}(z^{\ell,*}\,|\,z^{\ell})=\mathcal{N}(z^{\ell,*}\,|\,k_{\ell-1}^{\top}K_{\ell-1}^{+}z^{\ell},\sigma_{\ell}^{2}\kappa_{\ell-1}-\sigma_{\ell}^{2}k_{\ell-1}^{\top}K_{\ell-1}^{+}k_{\ell-1})
\end{equation}
where $K^{+}$ denotes the pseudo inverse. The saddle-point equations
for the training kernels $K_{\ell}$ remain unchanged. For the train-test
and the test-test kernels, the saddle-point equations at $\tilde{k}_{\ell}=\tilde{\kappa}_{\ell}=0$
are
\begin{equation}
k_{\ell}=\langle\phi(z^{\ell})\langle\phi(z^{\ell,*})^{\top}\rangle_{z^{\ell,*}|z^{\ell}}\rangle_{z^{\ell}},\label{eq:k_ell_sigmoid}
\end{equation}
\begin{equation}
\kappa_{\ell}=\langle\langle\phi(z^{\ell,*})\phi(z^{\ell,*})^{\top}\rangle_{z^{\ell,*}|z^{\ell}}\rangle_{z^{\ell}},\label{eq:kappa_ell_sigmoidal}
\end{equation}
and $k_{0}=\frac{1}{N_{0}}X^{\top}x_{*}$, $\kappa_{0}=\frac{1}{N_{0}}x_{*}^{\top}x_{*}$.
In the last layer, the conditional distribution of the test preactivations
given the training preactivations is also identical to the prior and
thus $P(a,\hat{z}^{L})=P_{0}(z^{L,*}\,|\,z^{L})P(a,z^{L})$. In total,
we arrive at
\begin{equation}
P(A,\hat{Z}_{L},\dots,\hat{Z}_{1})=\prod_{i=1}^{N}[P_{0}(z_{i}^{L,*}\,|\,z_{i}^{L})P(z_{i}^{L}\,|\,a_{i})P(a_{i})]\prod_{\ell=1}^{L-1}[\prod_{j=1}^{N}[P_{0}(z_{j}^{\ell,*}\,|\,z_{j}^{\ell})P(z_{j}^{\ell})]]
\end{equation}
for the test posterior.

The mean predictor follows immediately from the definition of the
joint posterior,
\begin{equation}
f_{r}(x)=\sum_{\gamma=1}^{n}P_{\gamma}a_{\gamma}^{r}\langle\langle\phi(z^{L,*})\rangle_{z^{L,*}|z^{L}}\rangle_{z^{L}|a_{\gamma}}
\end{equation}
where $z^{L,*}$ denotes the preactivation corresponding to the input
$x$. For $\phi(z)=\frac{1}{2}[1+\mathrm{erf}(\sqrt{\pi}z/2)]$ the
Gaussian expectation $z^{L,*}|z^{L}$ is solvable \citep{doi:10.1080/03610918008812164},
leading to
\begin{equation}
\langle\phi(z^{L,*})\rangle_{z^{L,*}|z^{L}}=\phi\big(c_{L}(x)k_{L-1}(x)^{\top}K_{L-1}^{+}z^{L}\big),\qquad c_{L}(x)=\frac{1}{\sqrt{1+\frac{\pi\sigma_{L}^{2}}{2}[\kappa_{L-1}(x)-k_{L-1}(x)^{\top}K_{L-1}^{+}k_{L-1}(x)]}}
\end{equation}
where we made the dependence of $\kappa_{L-1}$ and $k_{L-1}$ on
the input $x$ explicit. Thus, the mean predictor is
\begin{equation}
f_{r}(x)=\sum_{\gamma=1}^{n}P_{\gamma}a_{\gamma}^{r}\langle\phi[c(x)k_{L-1}(x)^{\top}K_{L-1}^{+}z^{L}]\rangle_{z^{L}|a_{\gamma}}.
\end{equation}
The mean kernels for a pair of inputs $x_{1},x_{2}$ also follow
immediately from the posterior,
\begin{equation}
K_{L}(x_{1},x_{2})=\sum_{\gamma=1}^{n}P_{\gamma}\langle\langle\phi(z_{1}^{L})\phi(z_{2}^{L})\rangle_{z_{1}^{L},z_{2}^{L}|z^{L}}\rangle_{z^{L}|a_{\gamma}},
\end{equation}
\begin{equation}
K_{\ell}(x_{1},x_{2})=\langle\langle\phi(z_{1}^{\ell})\phi(z_{2}^{\ell})\rangle_{z_{1}^{\ell},z_{2}^{\ell}|z^{\ell}}\rangle_{z^{\ell}},\quad\ell=1,\dots,L-1,
\end{equation}
where $z_{1}^{\ell},z_{2}^{\ell}$ denote the preactivations corresponding
to the inputs $x_{1},x_{2}$. The preactivation kernels are
\begin{equation}
K_{\ell}^{z}(x_{1},x_{2})=\sigma_{\ell}^{2}\kappa_{\ell-1}(x_{1},x_{2})+\sigma_{\ell}^{2}k_{\ell-1}(x_{1})^{\top}\tilde{K}_{\ell-1}k_{\ell-1}(x_{2}),\quad\ell=1,\dots,L,
\end{equation}
with $\kappa_{0}(x_{1},x_{2})=\frac{1}{N_{0}}x_{1}^{\top}x_{2}$.

For $\phi(z)=\frac{1}{2}[1+\mathrm{erf}(\sqrt{\pi}z/2)]$ the Gaussian
expectation $\langle\phi(z_{1}^{\ell})\phi(z_{2}^{\ell})\rangle_{z_{1}^{\ell},z_{2}^{\ell}|z^{\ell}}$
can be solved in terms of Owen's T function \citep{doi:10.1080/03610918008812164,PhysRevResearch.3.043077};
for simplicity we assume $\kappa_{\ell-1}(x_{1},x_{2})-k_{\ell-1}(x_{1})^{\top}K_{\ell-1}^{+}k_{\ell-1}(x_{2})\approx0$
for $x_{1}\neq x_{2}$. This decouples the fluctuations of $z_{1}^{\ell},z_{2}^{\ell}|z^{\ell}$,
leading to $\langle\phi(z_{1}^{\ell})\phi(z_{2}^{\ell})\rangle_{z_{1}^{\ell},z_{2}^{\ell}|z^{\ell}}\approx\phi\big(c_{\ell}(x_{1})k_{\ell-1}(x_{1})^{\top}K_{\ell-1}^{+}z^{\ell}\big)\phi\big(c_{\ell}(x_{2})k_{\ell-1}(x_{2})^{\top}K_{\ell-1}^{+}z^{\ell}\big)$
for $x_{1}\neq x_{2}$.

\fbox{\begin{minipage}[t]{1\columnwidth - 2\fboxsep - 2\fboxrule}%
\textbf{Summary:} Test preactivation posterior on test inputs $x_{*}$
\begin{equation}
P(z^{\ell,*}\,|\,z^{\ell})=\mathcal{N}(z^{\ell,*}\,|\,k_{\ell-1}^{\top}K_{\ell-1}^{+}z^{\ell},\sigma_{\ell}^{2}\kappa_{\ell-1}-\sigma_{\ell}^{2}k_{\ell-1}^{\top}K_{\ell-1}^{+}k_{\ell-1}),\quad\ell=1,\dots,L,
\end{equation}
with
\begin{equation}
k_{\ell}=\langle\phi(z^{\ell})\langle\phi(z^{\ell,*})^{\top}\rangle_{z^{\ell,*}|z^{\ell}}\rangle_{z^{\ell}},\quad\ell=1,\dots,L,
\end{equation}
\begin{equation}
\kappa_{\ell}=\langle\langle\phi(z^{\ell,*})\phi(z^{\ell,*})^{\top}\rangle_{z^{\ell,*}|z^{\ell}}\rangle_{z^{\ell}},\quad\ell=1,\dots,L,
\end{equation}
and $K_{0}=\frac{1}{N_{0}}X^{\top}X$, $k_{0}=\frac{1}{N_{0}}X^{\top}x_{*}$,
$\kappa_{0}=\frac{1}{N_{0}}x_{*}^{\top}x_{*}$.

Predictor on single test input $x$:
\begin{equation}
f_{r}(x)=\sum_{\gamma=1}^{n}P_{\gamma}a_{\gamma}^{r}\langle\langle\phi(z^{L,*})\rangle_{z^{L,*}|z^{L}}\rangle_{z^{L}|a_{\gamma}}
\end{equation}
where $z^{\ell,*}$ are the preactivations on $x$. For $\phi(z)=\frac{1}{2}[1+\mathrm{erf}(\sqrt{\pi}z/2)]$:
$\langle\phi(z^{\ell,*})\rangle_{z^{\ell,*}|z^{\ell}}=\phi\big(c_{\ell}k_{\ell-1}^{\top}K_{\ell-1}^{+}z^{\ell}\big)$
with $c_{\ell}=1/\sqrt{1+\frac{\pi\sigma_{\ell}^{2}}{2}[\kappa_{\ell-1}-k_{\ell-1}^{\top}K_{\ell-1}^{+}k_{\ell-1}]}$.%
\end{minipage}}

\textbf{Regime $\alpha_{0}>1$:} For $\alpha_{0}>1$, the first layer
preactivations are restricted to a $N_{0}$ dimensional subspace and
the kernel $K_{0}$ is singular. The pseudo-inverse appearing in the
theory restricts the matrix inversion to the corresponding subspace.
To demonstrate that the theory is correct beyond $\alpha_{0}=1$ we
randomly project the images to a $N_{0}=50$ dimensional subspace
such that $\alpha_{0}=2$ for $P=100$ (see Fig.~\ref{fig:sigmoidal_mnist_projected}).
The nature of the solution indeed stays the same, a redundant code
with neurons coding for either one or two classes, and the theory
is in accurate quantitative agreement.

\section{ReLU Networks ($L=1$)}

\subsection{Training \& Test Posterior}

We directly start with the joint posterior on training and test data
but restrict the theory to $L=1$ and set $\sigma_{a}^{2}=1/P$. Following
the same initial steps as for the linear and sigmoidal networks leads
for the joint posterior of the readout weight matrix $A$ and the
$N\times(P+P_{*})$ preactivation matrix $\hat{Z}$ on $P_{*}$ test
inputs $x_{*}$ to
\begin{equation}
P(A,\hat{Z})\propto\int dt\,e^{-NE(t)}\prod_{i=1}^{N}P(a_{i},\hat{z}_{i}),
\end{equation}
\begin{align}
E(t)= & -\frac{T}{2}\tr t^{\top}t+\tr t^{\top}Y-\log\int dP_{0}(a)\int d\mathcal{N}(z\,|\,0,\sigma_{1}^{2}K_{0})\,e^{a^{\top}t^{\top}\phi(z)}
\end{align}
\begin{equation}
P(a,\hat{z})\propto P_{0}(a)\mathcal{N}(\hat{z}\,|\,0,\sigma_{1}^{2}\hat{K}_{0})e^{a^{\top}t^{\top}\phi(z)}
\end{equation}
where we dropped the layer index. We explicitly break permutation
symmetry for $n$ outlier neurons, leading to
\begin{equation}
P(A,\hat{Z})\propto\int dt\,e^{-NE(t)}\prod_{i=1}^{n}[P(a_{i},\hat{z}_{i})]\prod_{i=n+1}^{N}[P(a_{i},\hat{z}_{i})],
\end{equation}
\begin{align}
E(t)= & -\frac{T}{2}\tr t^{\top}t+\tr t^{\top}Y-\big(1-\frac{n}{N}\big)\log\int dP_{0}(a)\int d\mathcal{N}(z\,|\,0,\sigma_{1}^{2}K_{0})\,e^{a^{\top}t^{\top}\phi(z)}\nonumber \\
 & +\frac{1}{N}\sum_{i=1}^{n}\log\int dP_{0}(a_{i})\int d\mathcal{N}(z_{i}\,|\,0,\sigma_{1}^{2}K_{0})\,e^{a_{i}^{\top}t^{\top}\phi(z_{i})}.
\end{align}
Rescaling $a_{i}\to\sqrt{N}\bar{a}_{i}$ and $z_{i}\to\sqrt{N}\bar{z}_{i}$
for $i=1,\dots,n$ ensures that the contribution from the outliers
to the energy $E(t)$ is not suppressed by $O(1/N)$. The corresponding
integrals can be solved in saddle-point approximation,
\begin{equation}
\frac{1}{N}\log\int dP_{0}(\sqrt{N}\bar{a}_{i})\int d\mathcal{N}(\sqrt{N}\bar{z}_{i}\,|\,0,\sigma_{1}^{2}K_{0})\,e^{N\bar{a}_{i}^{\top}t^{\top}\phi(\bar{z}_{i})}=-\frac{1}{2\sigma_{a}^{2}}\bar{a}_{i}^{\top}\bar{a}_{i}-\frac{1}{2\sigma_{1}^{2}}\bar{z}_{i}^{\top}K_{0}^{+}\bar{z}_{i}+\bar{a}_{i}^{\top}t^{\top}\phi(\bar{z}_{i}),
\end{equation}
where we used the homogeneity of ReLU, $\phi(\sqrt{N}\bar{z}_{i})=\sqrt{N}\phi(\bar{z}_{i})$,
with saddle-point equations
\begin{equation}
\bar{a}_{i}=\sigma_{a}^{2}t^{\top}\phi(\bar{z}_{i}),\qquad K_{0}^{+}\bar{z}_{i}=\phi^{\prime}(\bar{z}_{i})t\bar{a}_{i},\qquad i=1,\dots,n,
\end{equation}
where $\phi^{\prime}(z)$ denotes a diagonal matrix with elements
$\phi^{\prime}(z_{\mu})$. The saddle point approximation of the $t$
integral leads to
\begin{equation}
P(A,\hat{Z})=\prod_{i=1}^{n}[P(a_{i},\hat{z}_{i})]\prod_{i=n+1}^{N}[P(a_{i},\hat{z}_{i})].
\end{equation}

The readout posterior of the $N-n$ bulk neurons concentrates, similar
to the sigmoidal case, leading to
\begin{equation}
P(a_{i},\hat{z}_{i})=\delta(a_{i}-a_{0})P(z\,|\,a_{0})P_{0}(z^{*}\,|\,z),\quad i=n+1,\dots,N.
\end{equation}
Due to the rescaling with $\sqrt{N}$ all posteriors of the outlier
neurons concentrate, leading to

\begin{equation}
P(a_{i},\hat{z}_{i})=\delta(a_{i}-\sqrt{N}\bar{a}_{i})\delta(z_{i}-\sqrt{N}\bar{z}_{i})\delta(z_{i}^{*}-\sqrt{N}\bar{z}_{i}^{*}),\quad i=1,\dots,n,
\end{equation}
with $\bar{z}_{i}^{*}=k_{0}^{\top}K_{0}^{+}\bar{z}_{i}$. The saddle
point equation for $t$ simplifies to
\begin{equation}
Y=\big(1-\frac{n}{N}\big)\langle\phi(z)\rangle_{z|a_{0}}a_{0}^{\top}+\sum_{i=1}^{n}\phi(\bar{z}_{i})\bar{a}_{i}^{\top}+Tt
\end{equation}
where the contribution due to the outliers is $O(1)$ because both
readout weight and preactivation are $O(\sqrt{N})$.

The predictor on a test input $x$ is
\begin{equation}
f_{r}(x)=\big(1-\frac{n}{N}\big)a_{0}^{r}\langle\langle\phi[z(x)]\rangle_{z(x)|z}\rangle_{z|a_{0}}+\sum_{i=1}^{n}\bar{a}_{i}^{r}\phi[k_{0}(x)^{\top}K_{0}^{+}\bar{z}_{i}]
\end{equation}
with $k_{0}(x)=\frac{1}{N_{0}}X^{\top}x$. The Gaussian expectation
$\langle\phi[z(x)]\rangle_{z(x)|z}$ can be solved,
\begin{equation}
\langle\phi[z(x)]\rangle_{z(x)|z}=\frac{\sigma_{z(x)|z}}{\sqrt{2\pi}}e^{-\frac{\mu_{z(x)|z}^{2}}{2\sigma_{z(x)|z}^{2}}}+\mu_{z(x)|z}\,H[-\mu_{z(x)|z}/\sigma_{z(x)|z}]
\end{equation}
with $\mu_{z(x)|z}=k_{0}(x)^{\top}K_{0}^{+}z$ and $\sigma_{z(x)|z}^{2}=\sigma_{1}^{2}[\kappa_{0}(x)-k_{0}(x)^{\top}K_{0}^{+}k_{0}(x)]$.

The kernel on a pair of test inputs $x_{1},x_{2}$ is
\begin{equation}
K(x_{1},x_{2})=\langle\langle\phi(z_{1})\phi(z_{2})\rangle_{z_{1},z_{2}|z}\rangle_{z|a_{0}}+\sum_{i=1}^{n}\phi[k_{0}(x_{1})^{\top}K_{0}^{+}\bar{z}_{i}]\phi[k_{0}(x_{2})^{\top}K_{0}^{+}\bar{z}_{i}]
\end{equation}
where $z_{1},z_{2}$ denotes the preactivation on the test inputs.

\fbox{\begin{minipage}[t]{1\columnwidth - 2\fboxsep - 2\fboxrule}%
\textbf{Summary ($L=1$):} Posterior distribution on a set of test
inputs $x_{*}$ separates into $N-n$ bulk neurons where
\begin{equation}
P(a)=\delta(a-a_{0}),
\end{equation}
\begin{equation}
P(z\,|\,a)\propto\mathcal{N}(z\,|\,0,\sigma_{1}^{2}K_{0})e^{a^{\top}t^{\top}\phi(z)},
\end{equation}

\begin{equation}
P(z^{*}\,|\,z)=\mathcal{N}(z^{*}\,|\,k_{0}^{\top}K_{0}^{+}z,\sigma_{1}^{2}\kappa_{0}-\sigma_{1}^{2}k_{0}^{\top}K_{0}^{+}k_{0}),
\end{equation}
with $a_{0}=\frac{1}{P}t^{\top}\langle\phi(z)\rangle_{z|a_{0}}$ and
$n$ outlier neurons where 
\begin{equation}
P(a_{i})=\delta(a_{i}-\sqrt{N}\bar{a}_{i}),\quad i=1,\dots,n,
\end{equation}
\begin{equation}
P(z_{i}\,|\,a_{i})=\delta(z_{i}-\sqrt{N}\bar{z}_{i}),\quad i=1,\dots,n,
\end{equation}

\begin{equation}
P(z_{i}^{*}\,|\,z_{i})=\delta(z_{i}^{*}-\sqrt{N}k_{0}^{\top}K_{0}^{+}\bar{z}_{i}),\quad i=1,\dots,n,
\end{equation}
with $\bar{a}_{i}=\frac{1}{P}t^{\top}\phi(\bar{z}_{i})$ and $K_{0}^{+}\bar{z}_{i}=\phi^{\prime}(\bar{z}_{i})t\bar{a}_{i}$;
$t$ is determined by $Y=\big(1-\frac{n}{N}\big)\langle\phi(z)\rangle_{z|a_{0}}a_{0}^{\top}+\sum_{i=1}^{n}\phi(\bar{z}_{i})\bar{a}_{i}^{\top}+Tt$.
Finally, $K_{0}=\frac{1}{N_{0}}X^{\top}X$, $k_{0}=\frac{1}{N_{0}}X^{\top}x_{*}$,
$\kappa_{0}=\frac{1}{N_{0}}x_{*}^{\top}x_{*}$, and $\sigma_{a}^{2}=1/P$.

Predictor on test input $x$:
\begin{equation}
f_{r}(x)=a_{0}^{r}\langle\langle\phi[z(x)]\rangle_{z(x)|z}\rangle_{z|a_{0}}+\sum_{i=1}^{n}\bar{a}_{i}^{r}\phi[k_{0}(x)^{\top}K_{0}^{+}\bar{z}_{i}]
\end{equation}
\end{minipage}}

\textbf{Dynamics: }The Langevin dynamics are similar to the dynamics
of the sigmoidal networks: first, the loss decreases rapidly which
brings the network close to the solution space. Already at this early
stage a coding scheme is visible which is, however, not yet sparse.
Subsequently, the dynamics is dominated by the regularizer. In this
phase sparsity increases over time until the solution with three outliers
is reached. In contrast to the sigmoidal case, gradient descent without
noise does not reach this maximally sparse solution.

\section{Numerics}

The code will be made available upon publication on GitHub.

\subsection{Empirical Sampling}

For all experiments we relied heavily on NumPy \citep{Harris20_357}
and SciPy \citep{Virtanen20_261}. To obtain samples from the weight
posterior we use Hamiltonian Monte Carlo with the no-u-turn sampler
(see, e.g., \citep{Betancourt17_arXiv,pmlr-v139-izmailov21a}) implemented
in numpyro \citep{phan2019composable,bingham2019pyro} with jax \citep{jax2018github}
as a back-end; the model itself is implemented using the front-end
provided by pymc \citep{Salvatier2016}. Unless specified otherwise,
in all sampling experiments we use a small but finite temperature
$T=10^{-4}$ and keep $\sigma_{\ell}=1$ for $1\le\ell\le L$. For
all experiments, we preprocess each input by subtracting the mean
and normalizing such that $\|x_{\mu}\|^{2}=N$ for all $\mu$.

\subsection{Theory}

For linear networks, the solution is straightforward to implement.
For nonlinear networks, we solve the saddle point equations by constructing
an auxiliary squared loss function with its global minima at the self
consistent solutions, e.g., $\sum_{r=1}^{m}\sum_{\mu=1}^{P}[y_{\mu}^{r}-\langle a_{r}\phi(z_{\mu})\rangle_{a,z}-Tt_{\mu}^{r}]^{2}$
for $t$. This auxiliary loss is minimized using the gradient based
minimizer available in jaxopt.ScipyMinimize \citep{jaxopt_implicit_diff};
after convergence it is checked that the solution achieves zero auxiliary
loss, i.e., that a self-consistent solution is obtained.

On the toy example, the minimization is performed jointly for all
$a_{\gamma}$ and $t$; the one dimensional $z$ integrals are solved
numerically using Gauss-Hermite quadrature with fixed order for sigmoidal
networks or analytical solutions for ReLU networks.

Beyond the toy example, the minimization is performed jointly for
all $z_{\gamma}$, $a_{\gamma}$, and $t$. For ReLU networks, we
approximate $\max(0,x)\approx\frac{1}{c}\log(1+e^{cx})$ with $c=10$
to make the auxiliary loss differentiable.

\subsection{GP Theory}

In the GP theory \citep{Neal96,Williams96_ae5e3ce4,Lee18,Matthews18}
the predictor mean and variance are
\begin{equation}
f_{r}(x)=k_{GP}(x)^{\top}K_{GP}^{-1}y_{r}
\end{equation}
\begin{equation}
\langle\delta f_{r}(x;\Theta)^{2}\rangle_{\Theta}=\kappa_{GP}(x)+k_{GP}(x)^{\top}K_{GP}^{-1}k_{GP}(x)
\end{equation}
such that the generalization error is $\epsilon_{g}(x)=[y_{r}-f_{r}(x)]^{2}+\langle\delta f_{r}(x;\Theta)^{2}\rangle_{\Theta}$
where, unlike in non-lazy networks, the second contribution due to
the variance cannot be neglected. The associated GP kernel is determined
recursively for arbitrary inputs $x_{1},x_{2}$ by $K_{\ell}(x_{1},x_{2})=\langle\phi(z_{1}^{\ell})\phi(z_{2}^{\ell})\rangle$
where $z_{1}^{\ell},z_{2}^{\ell}$ are zero mean Gaussian with covariance
$\langle z_{1}^{\ell}z_{2}^{\ell}\rangle=\sigma_{\ell}^{2}K_{\ell-1}(x_{1},x_{2})$
and $K_{0}(x_{1},x_{2})=\frac{1}{N_{0}}x_{1}^{\top}x_{2}$. In the
predictor the last layer kernel appears, $K_{GP}(x_{1},x_{2})=K_{L}(x_{1},x_{2})$,
and $K_{GP}$ denotes the associated $P\times P$ matrix on the training
inputs $K_{GP}(x_{\mu},x_{\nu})$, $k_{GP}(x)$ the $P$-dimensional
vector containing $K_{GP}(x,x_{\mu})$, and $\kappa_{GP}(x)=K_{GP}(x,x)$.

For the nonlinearities cosidered here the Gaussian expectation $\langle\phi(z_{1}^{\ell})\phi(z_{2}^{\ell})\rangle$
is tractable for ReLU \citep{NIPS2009_5751ec3e} and erf \citep{doi:10.1080/03610918008812164,Williams96_ae5e3ce4}.
We employed the analytical formulas for the implementation and performed
the inversion using Cholesky decomposition.

\section{Supplemental Figures}

\begin{figure}
\includegraphics[width=0.5\columnwidth]{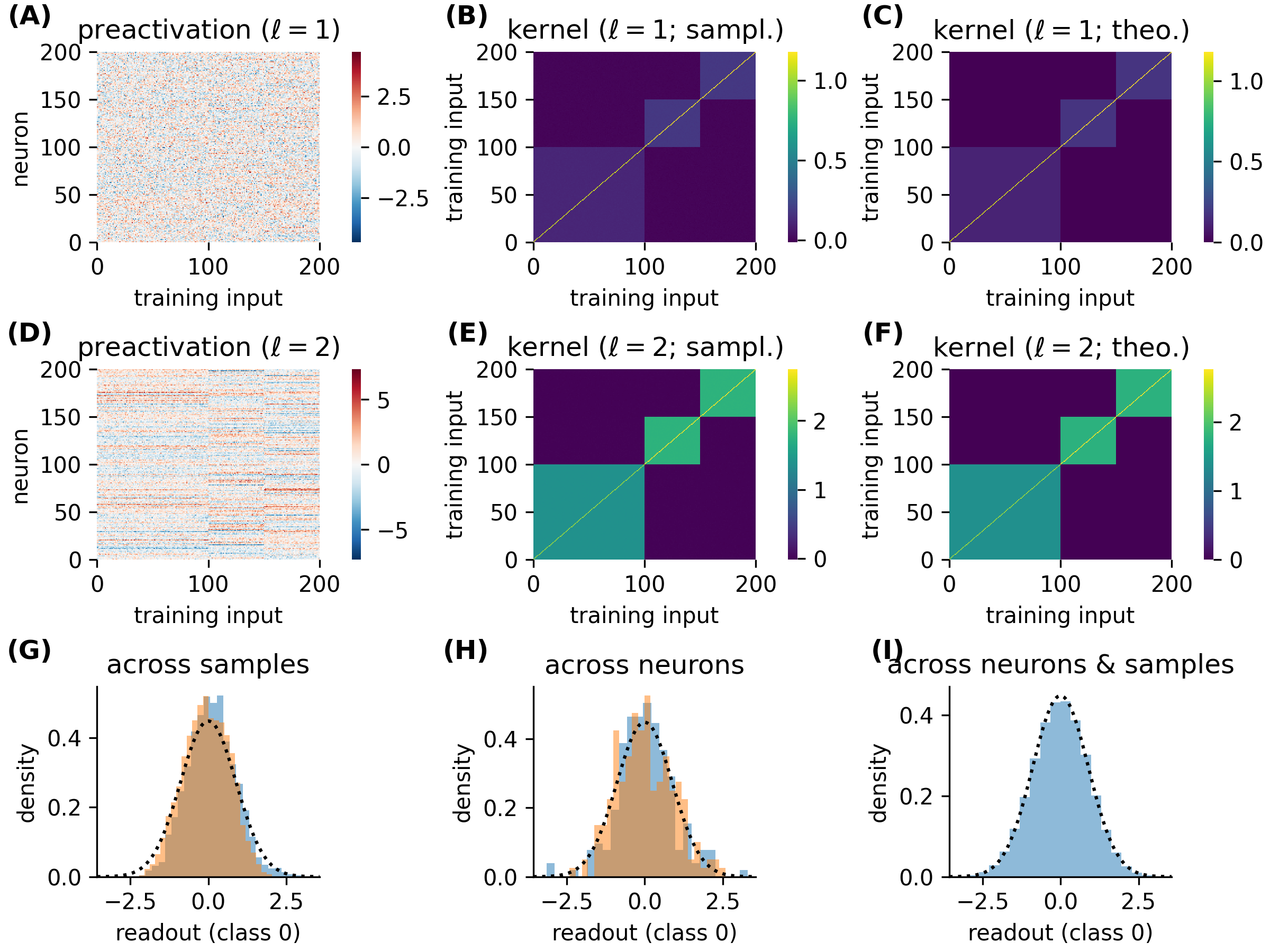}

\caption{Coding scheme in two hidden layer linear networks on random classification
task. (\textbf{A},\textbf{D}) Preactivations of all neurons on all
training inputs for a given weight sample in the first (A) and last
(D) hidden layer. (\textbf{B},\textbf{C},\textbf{E},\textbf{F}) Kernels
on training data from sampling (B,E) and theory (C,F) in the first
(B,C) and last (E,F) hidden layer. (\textbf{G}-\textbf{I}) Distribution
of readout weight on first class across samples for two given neurons
(G), across neurons for two given samples (H), and across neurons
and samples (I). Theoretical distribution (\ref{eq:a_posterior_linear})
as black dashed line. Parameters: $N=P=200$, $N_{0}=220$, classes
assigned with fixed ratios $[1/2,1/4,1/4]$, targets $y_{+}=1$ and
$y_{-}=0$. \label{fig:linear_code_suppmat}}
\end{figure}

\begin{figure}
\includegraphics[width=0.5\columnwidth]{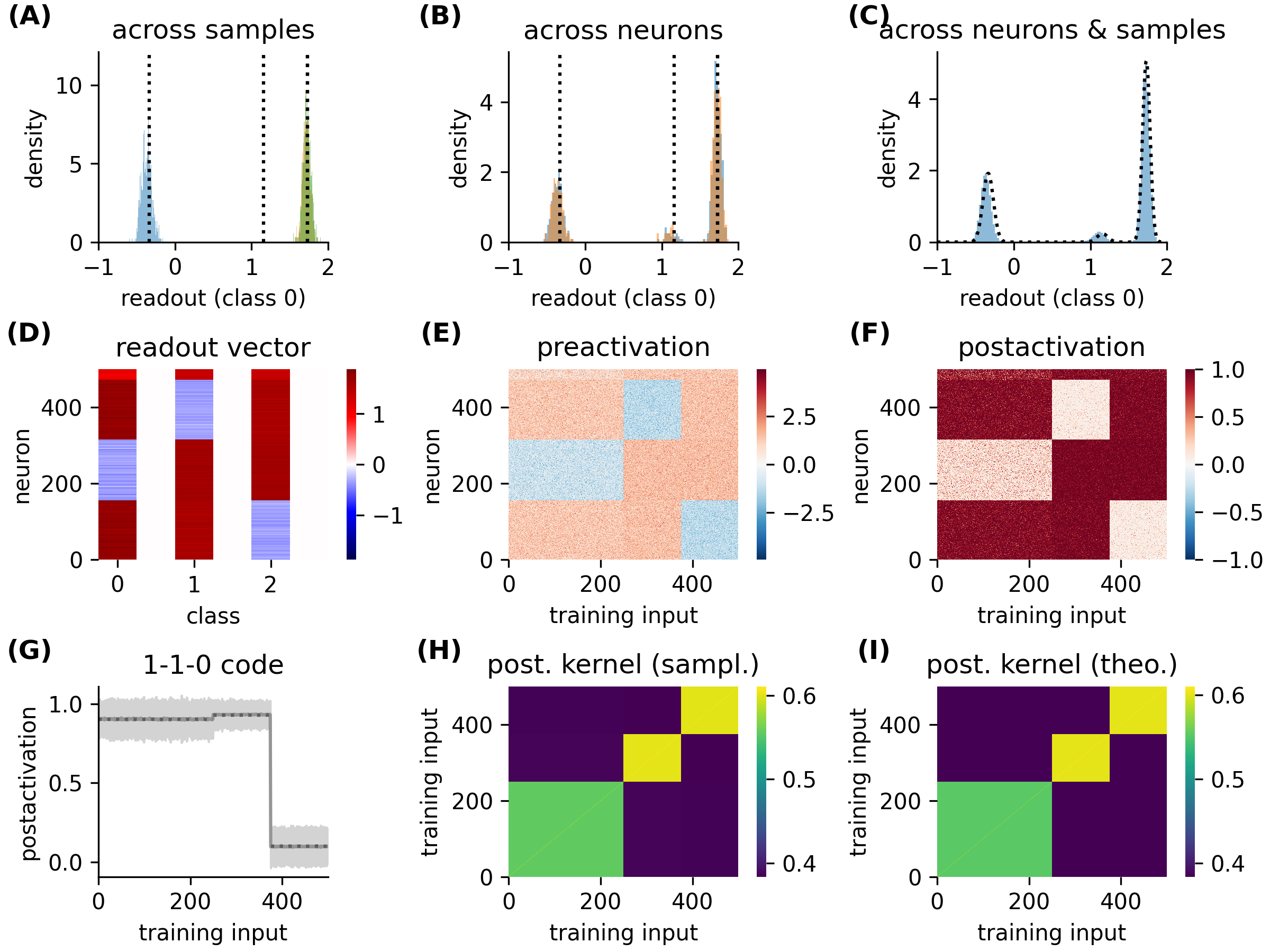}

\caption{Coding scheme in one hidden layer sigmoidal networks on random classification
task. (\textbf{A}-\textbf{C}) Distribution of readout weight on first
class across samples for three given neurons (A), across neurons for
two given samples (B), and across neurons and samples (C). Theoretical
distribution (\ref{eq:a_posterior_sigmoidal}) as black dashed line,
in (C) including finite $P$ correction. (\textbf{D}) Sample of the
readout weights. (\textbf{E},\textbf{F}) Pre- (E) and postactivations
(F) of all neurons on all training inputs for a given weight sample.
(\textbf{G}) Postactivation of neurons with 1-0-1 code from sampling
(gray) and theory (black dashed). Shaded area shows standard deviation
across neurons for a given weight sample. (\textbf{H},\textbf{I})
Postactivation kernel on training data from sampling (H) and theory
(I). Parameters: $N=P=500$, $N_{0}=520$, classes assigned with fixed
ratios $[1/2,1/4,1/4]$, targets $y_{+}=1$ and $y_{-}=1/2$. \label{fig:sigmoidal_code_suppmat}}
\end{figure}

\begin{figure}
\includegraphics[width=0.5\columnwidth]{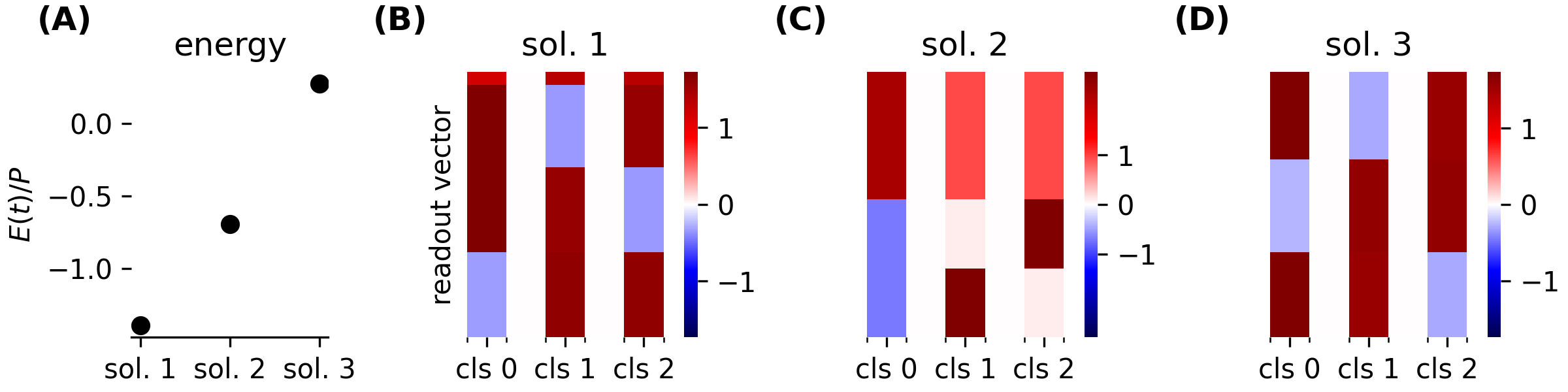}

\caption{Multiple self-consistent solutions. (\textbf{A}) Energy $E(t)$ for
multiple self-consistent solutions for $t$. Solutions with higher
energy are suppressed with probability $\exp[-N\Delta E]$. (\textbf{B}-\textbf{D})
Readout vector corresponding to the self consistent solutions for
$t$. Parameters: $N=P=500$, $N_{0}=520$, classes assigned with
fixed ratios $[1/2,1/4,1/4]$, targets $y_{+}=1$ and $y_{-}=1/2$.
\label{fig:sigmoidal_code_multisolution}}
\end{figure}

\begin{figure}
\includegraphics[width=0.5\columnwidth]{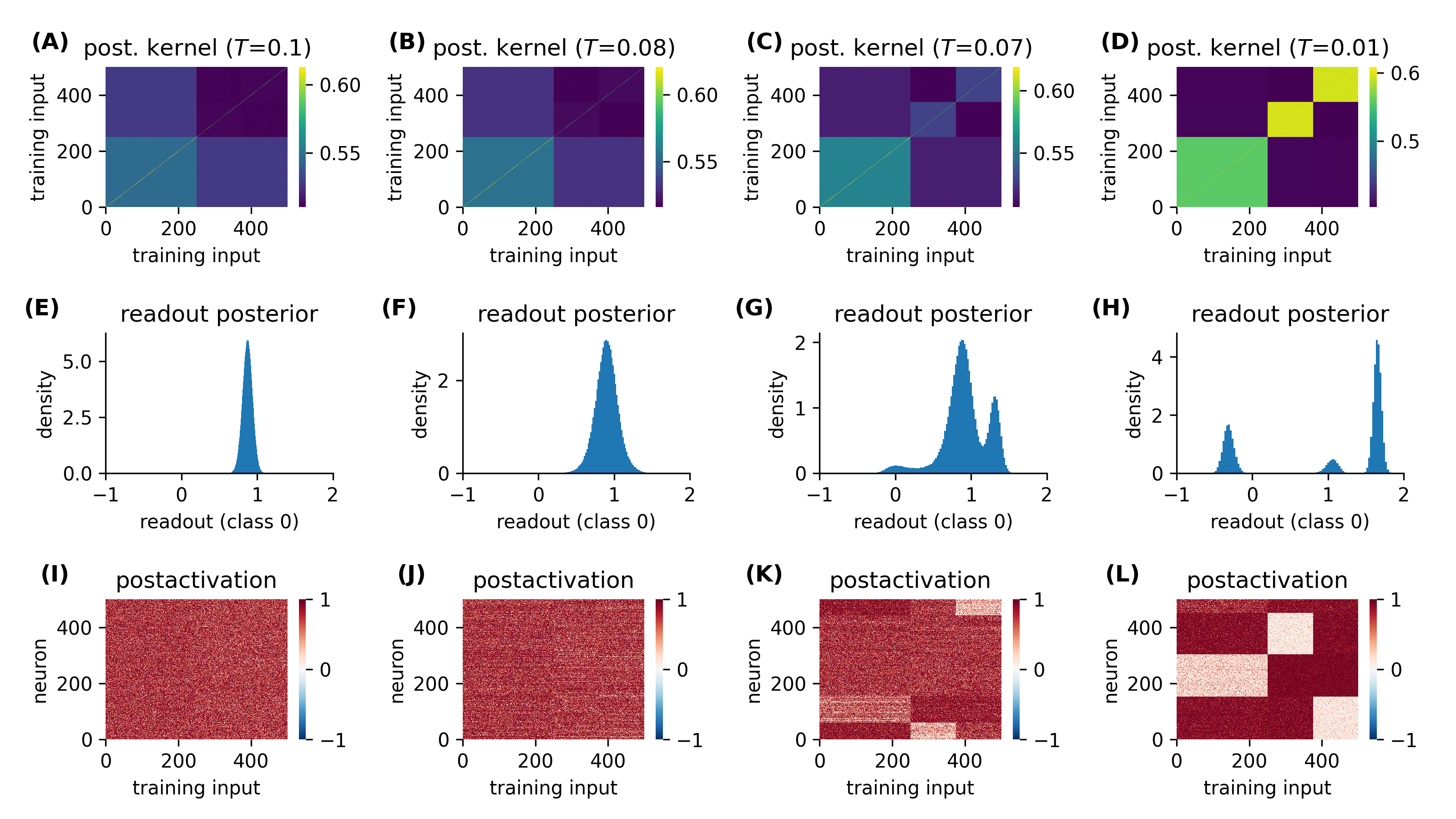}

\caption{Emergence of code with decreasing temperature in one hidden layer
sigmoidal networks on random classification task. (\textbf{A}-\textbf{D})
Postactivation kernel on training data from sampling for varying temperature.
(\textbf{E}-\textbf{H}) Readout posterior of first class for varying
temperature. (\textbf{I}-\textbf{L}) Postactivations of all neurons
on all training inputs for a given weight sample for varying temperature.
Parameters: $N=P=500$, $N_{0}=520$, classes assigned with fixed
ratios $[1/2,1/4,1/4]$, targets $y_{+}=1$ and $y_{-}=1/2$. \label{fig:sigmoidal_code_temp}}
\end{figure}

\begin{figure}
\includegraphics[width=0.5\columnwidth]{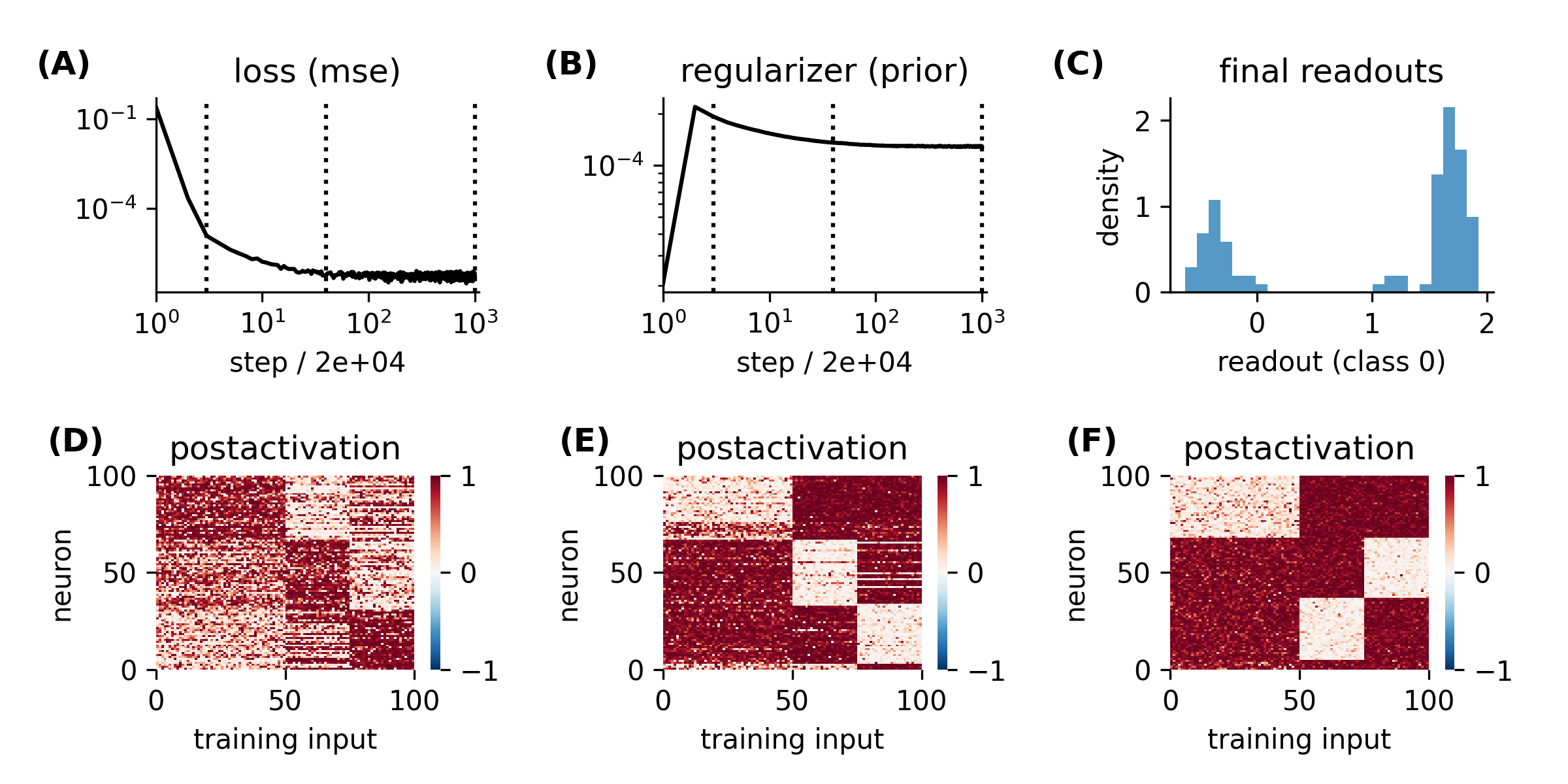}

\caption{Langevin dynamics of one hidden layer sigmoidal network on random
classification task. (\textbf{A},\textbf{B}) Mean squared error (A)
and regularizer (B) over time. (\textbf{C}) Readout posterior of first
class at final step. (\textbf{D}-\textbf{F}) Postactivations of all
neurons on all training inputs for a given weight sample at the steps
indicated in (A,B). Parameters: $N=P=100$, $N_{0}=120$, classes
assigned with fixed ratios $[1/2,1/4,1/4]$, targets $y_{+}=1$ and
$y_{-}=1/2$. \label{fig:sigmoidal_langevin}}
\end{figure}

\begin{figure}
\includegraphics[width=0.5\columnwidth]{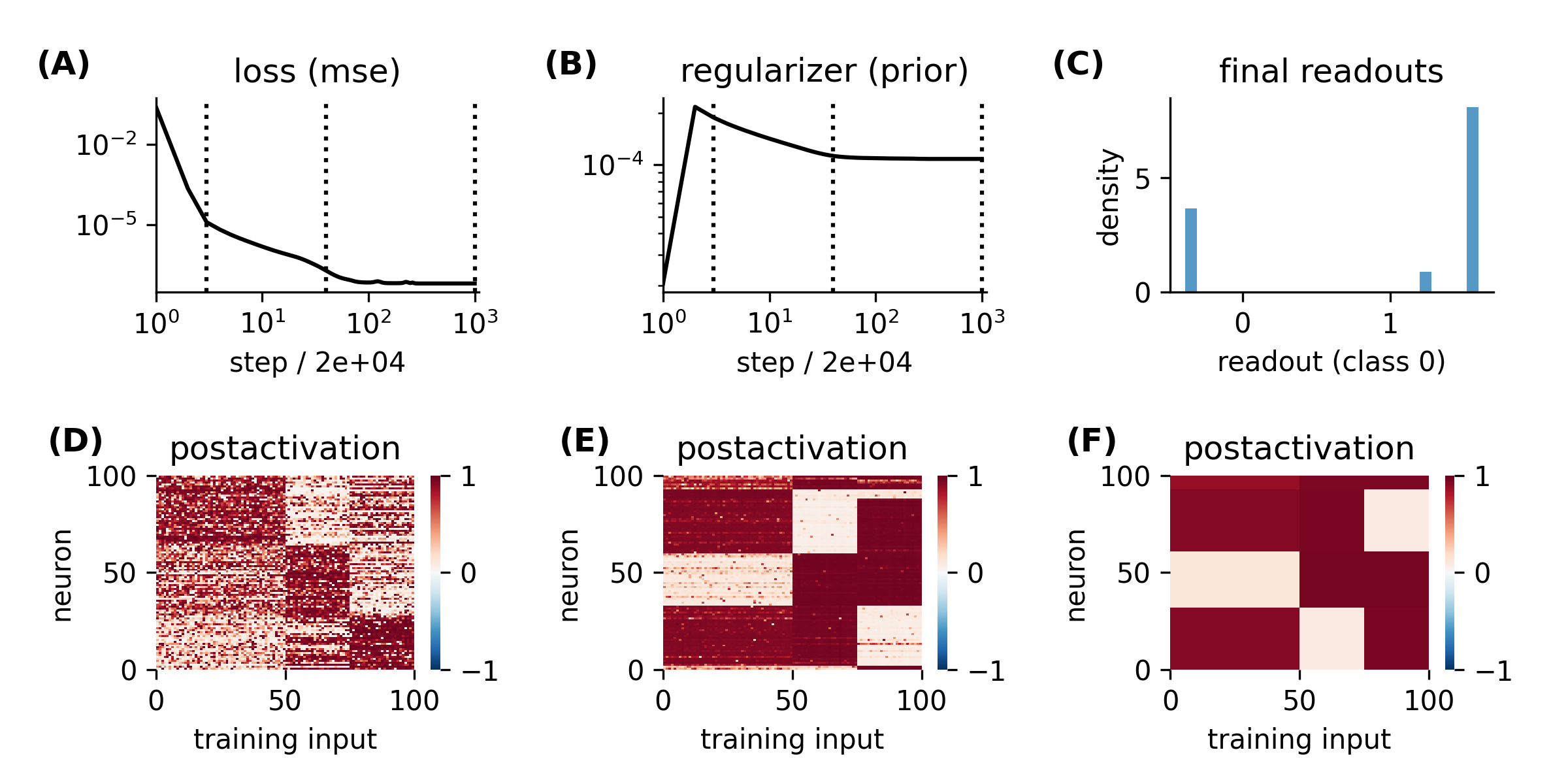}

\caption{Gradient descent dynamics of one hidden layer sigmoidal network on
random classification task. (\textbf{A},\textbf{B}) Mean squared error
(A) and regularizer (B) over time. (\textbf{C}) Readout posterior
of first class at final step. (\textbf{D}-\textbf{F}) Postactivations
of all neurons on all training inputs for a given weight sample at
the steps indicated in (A,B). Parameters: $N=P=100$, $N_{0}=120$,
classes assigned with fixed ratios $[1/2,1/4,1/4]$, targets $y_{+}=1$
and $y_{-}=1/2$. \label{fig:sigmoidal_dynamics}}
\end{figure}

\begin{figure}
\includegraphics[width=0.5\columnwidth]{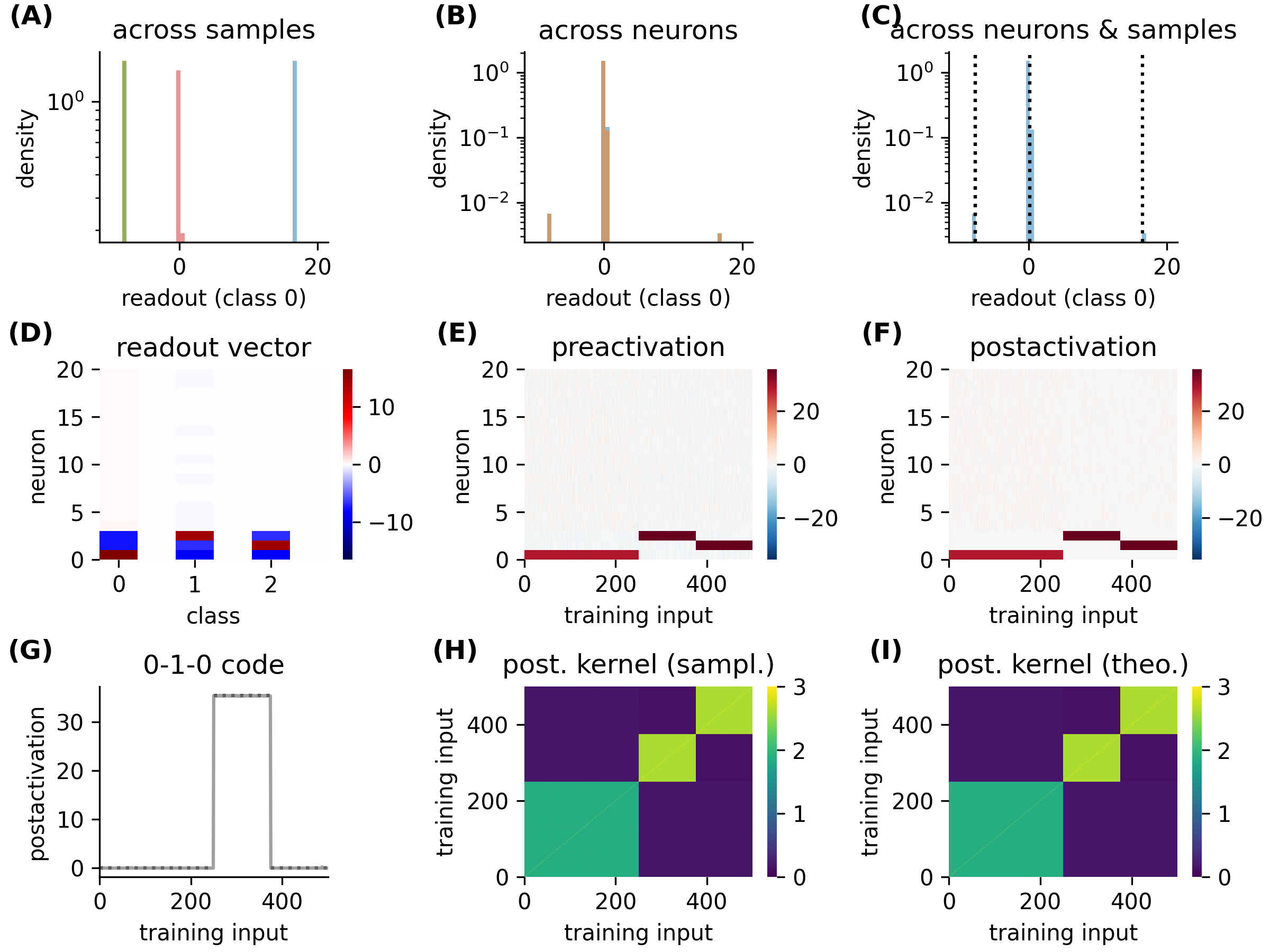}

\caption{Coding scheme in one hidden layer ReLU networks on random classification
task. (\textbf{A}-\textbf{C}) Distribution of readout weight on first
class across samples for four given neurons (A), across neurons for
two given samples (B), and across neurons and samples (C). Theoretical
distribution (\ref{eq:a_posterior_sigmoidal}) as black dashed line.
(\textbf{D}) Sample of the readout weights for a sorted subset of
$20$ neurons. (\textbf{E},\textbf{F}) Pre- (E) and postactivations
(F) for a sorted subset of $20$ neurons on all training inputs for
a given weight sample. (\textbf{G}) Postactivation of the neuron with
0-1-0 code from sampling (gray) and theory (black dashed). (\textbf{H},\textbf{I})
Postactivation kernel on training data from sampling (H) and theory
(I). Parameters: $N=P=500$, $N_{0}=520$, classes assigned with fixed
ratios $[1/2,1/4,1/4]$, targets $y_{+}=1$ and $y_{-}=-1/2$. \label{fig:relu_code_suppmat}}
\end{figure}

\begin{figure}
\includegraphics[width=0.5\columnwidth]{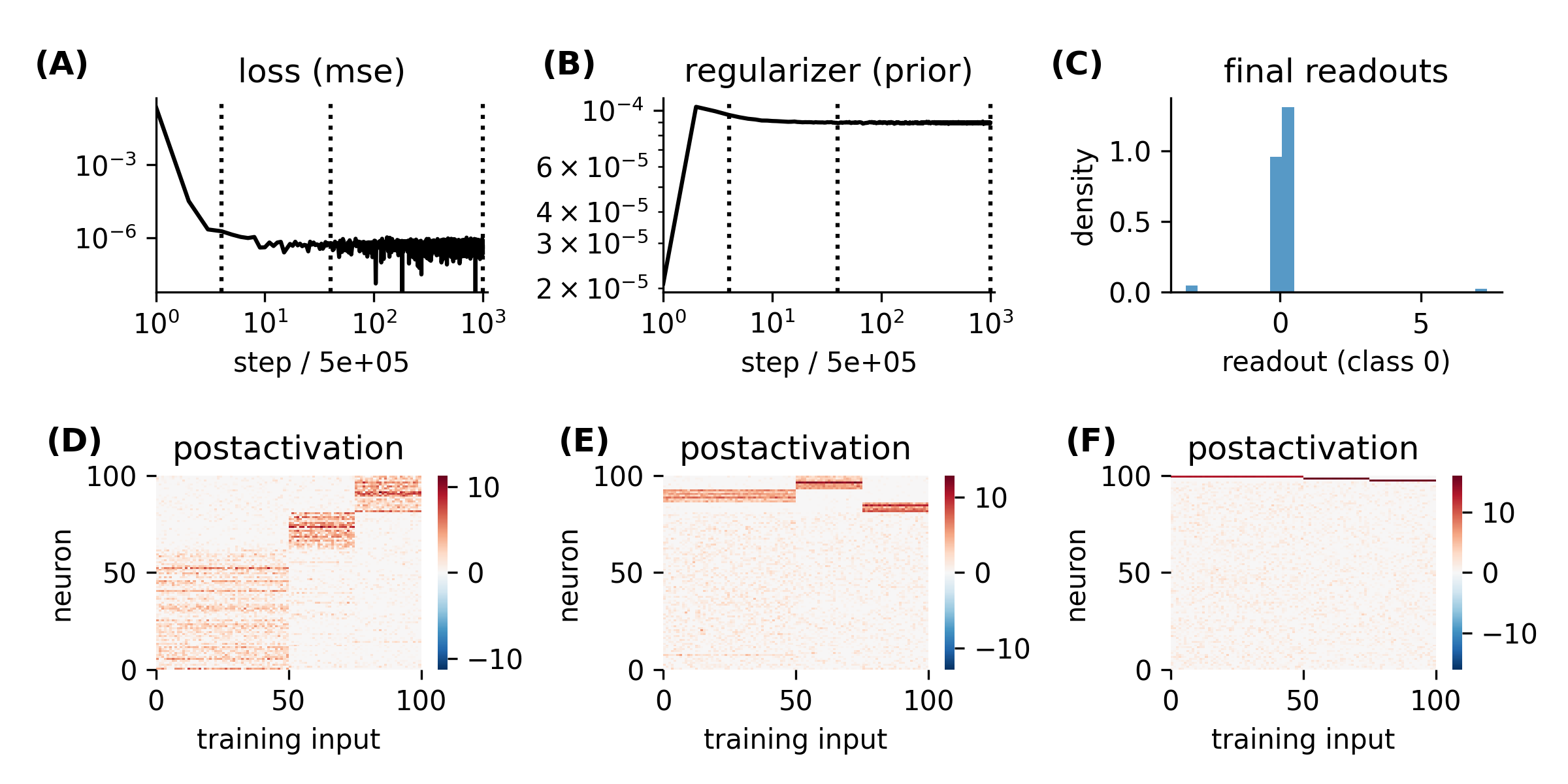}

\caption{Langevin dynamics of one hidden layer ReLU network on random classification
task. (\textbf{A},\textbf{B}) Mean squared error (A) and regularizer
(B) over time. (\textbf{C}) Readout posterior of first class at final
step. (\textbf{D}-\textbf{F}) Postactivations of all neurons on all
training inputs for a given weight sample at the steps indicated in
(A,B). Parameters: $N=P=100$, $N_{0}=120$, classes assigned with
fixed ratios $[1/2,1/4,1/4]$, targets $y_{+}=1$ and $y_{-}=-1/2$.
\label{fig:relu_langevin}}
\end{figure}

\begin{figure}
\includegraphics[width=0.5\columnwidth]{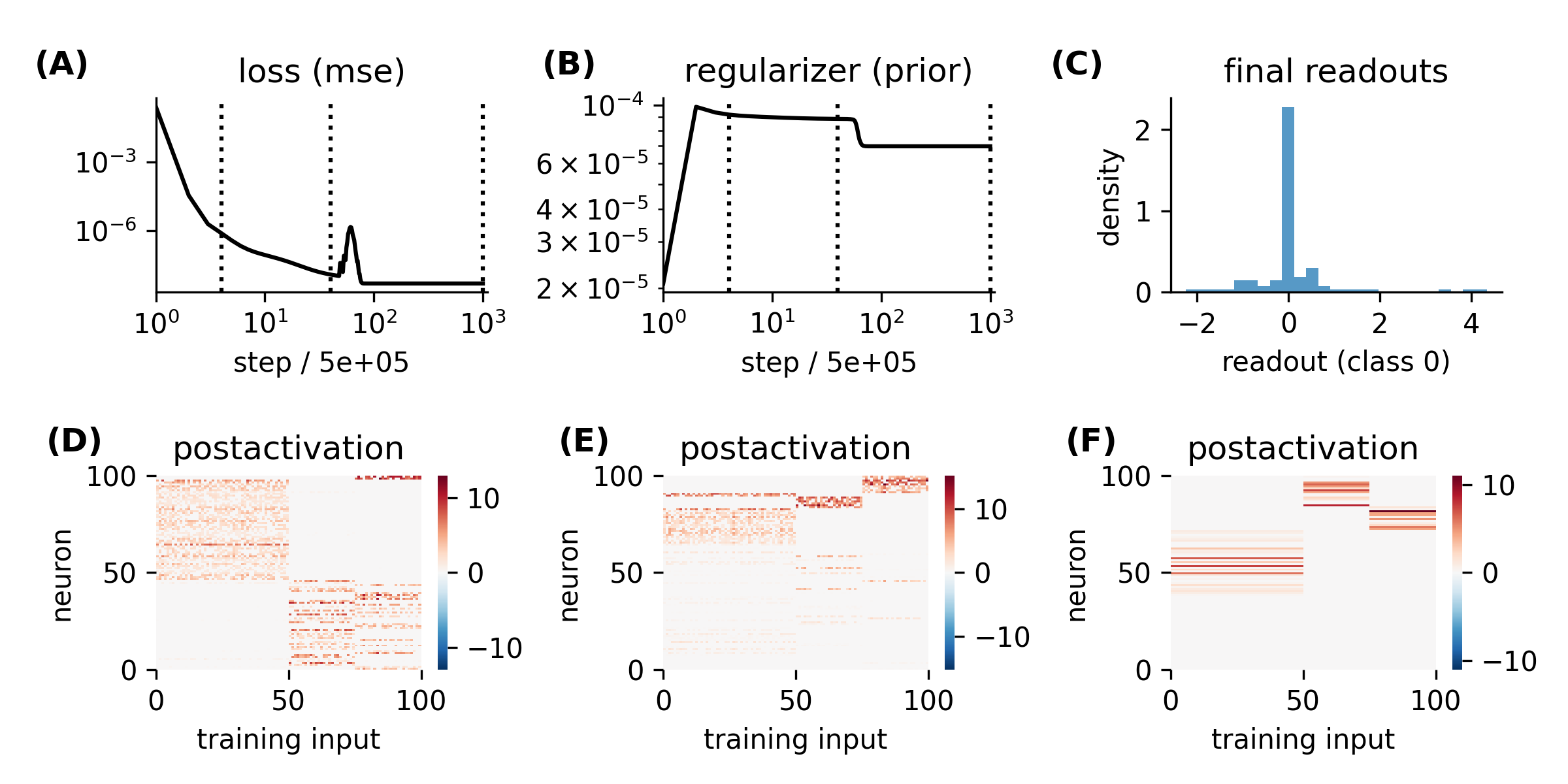}

\caption{Gradient decsent dynamics of one hidden layer ReLU network on random
classification task. (\textbf{A},\textbf{B}) Mean squared error (A)
and regularizer (B) over time. (\textbf{C}) Readout posterior of first
class at final step. (\textbf{D}-\textbf{F}) Postactivations of all
neurons on all training inputs for a given weight sample at the steps
indicated in (A,B). Parameters: $N=P=100$, $N_{0}=120$, classes
assigned with fixed ratios $[1/2,1/4,1/4]$, targets $y_{+}=1$ and
$y_{-}=-1/2$. \label{fig:relu_dynamics}}
\end{figure}

\begin{figure}
\includegraphics[width=0.5\columnwidth]{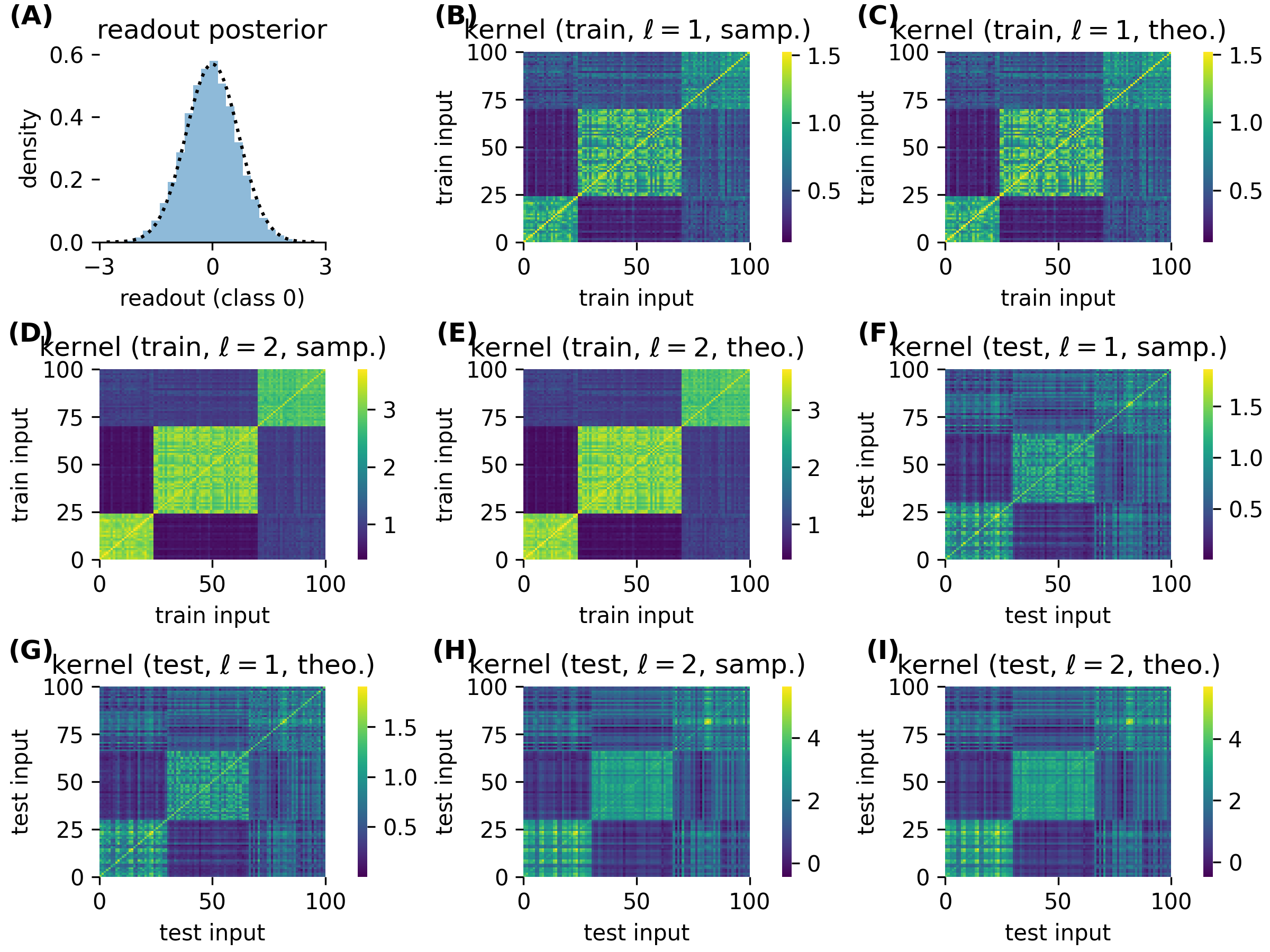}

\caption{Generalization of two hidden layer linear networks on MNIST. (\textbf{A})
Readout posterior of first class. (\textbf{B}-\textbf{I}) Kernel on
training (A-E) and test (F-I) data in first (B,C,F,G) and second layer
(D,E,H,I) from sampling (B,D,F,H) and theory (C,E,G,I). Parameters:
$N=P=100$, $N_{0}=784$, classes 0, 1, 2 assigned randomly with probability
$1/3$, targets $y_{+}=1$ and $y_{-}=0$. \label{fig:linear_mnist_suppmat}}
\end{figure}

\begin{figure}
\includegraphics[width=0.5\columnwidth]{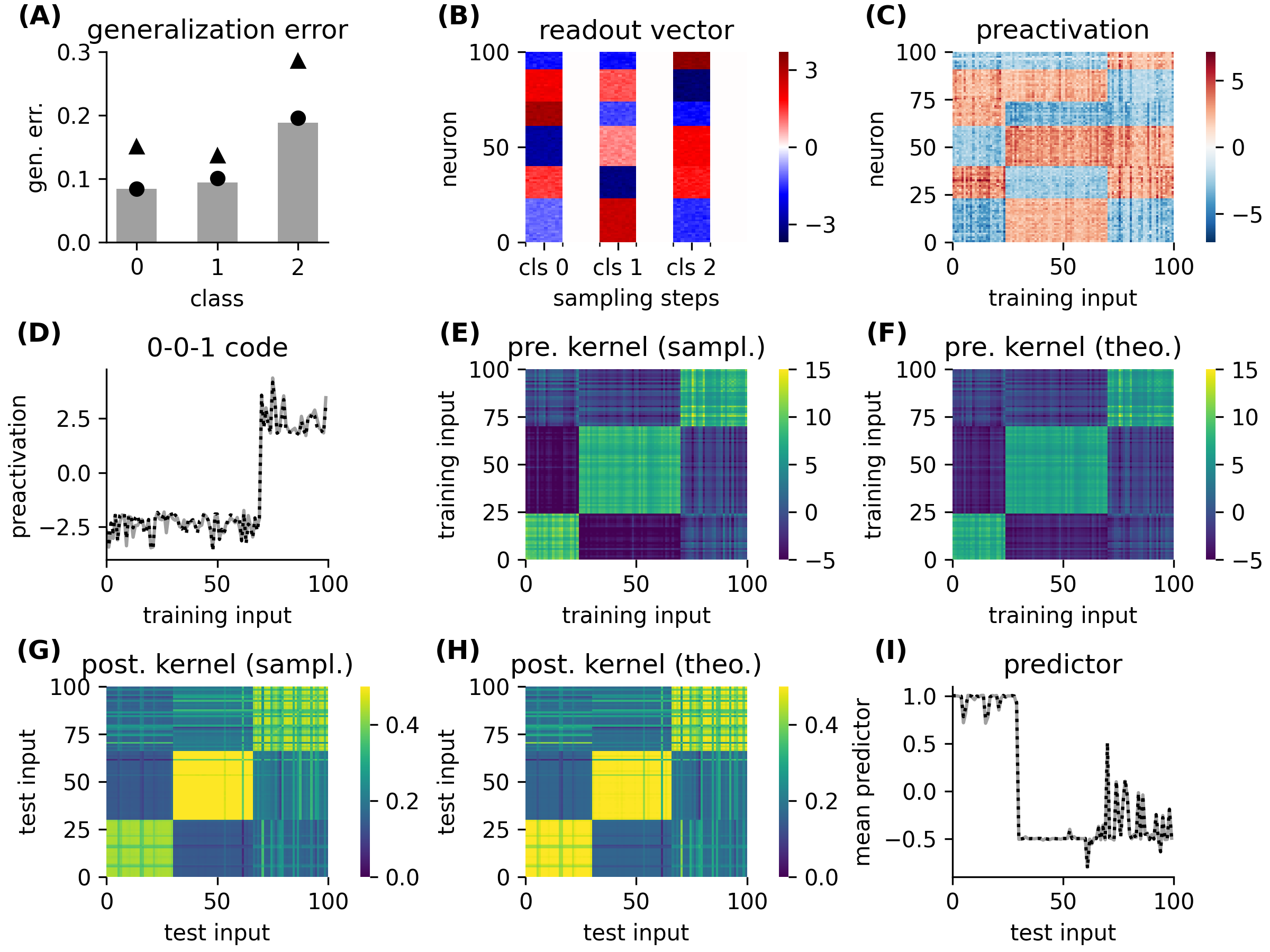}

\caption{Generalization of one hidden layer sigmoidal networks on MNIST. (\textbf{A})
Generalization error for each class averaged over $1,000$ test inputs
from sampling (gray bars), theory (Eq.~(\ref{eq:predictor_sigmoidal}),
black circles), and GP theory (back triangles). (\textbf{B}) Sample
of the readout weights for all neurons. (\textbf{C}) Preactivations
of all neurons on all training inputs for a given weight sample. (\textbf{D})
Averaged preactivation of neurons with 0-0-1 code from sampling (gray)
and theory (black dashed). (\textbf{E},\textbf{F}) Preactivation kernel
on training data from sampling (E) and theory (F). (\textbf{G},\textbf{H})
Postactivation kernel on test data from sampling (G) and theory (H).
(\textbf{I}) Mean predictor for class 0 from sampling (gray) and theory
(black dashed). Parameters: $N=P=100$, $N_{0}=784$, classes 0, 1,
2 assigned randomly with probability $1/3$, targets $y_{+}=1$ and
$y_{-}=-1/2$. \label{fig:sigmoidal_mnist_suppmat}}
\end{figure}

\begin{figure}
\includegraphics[width=0.5\columnwidth]{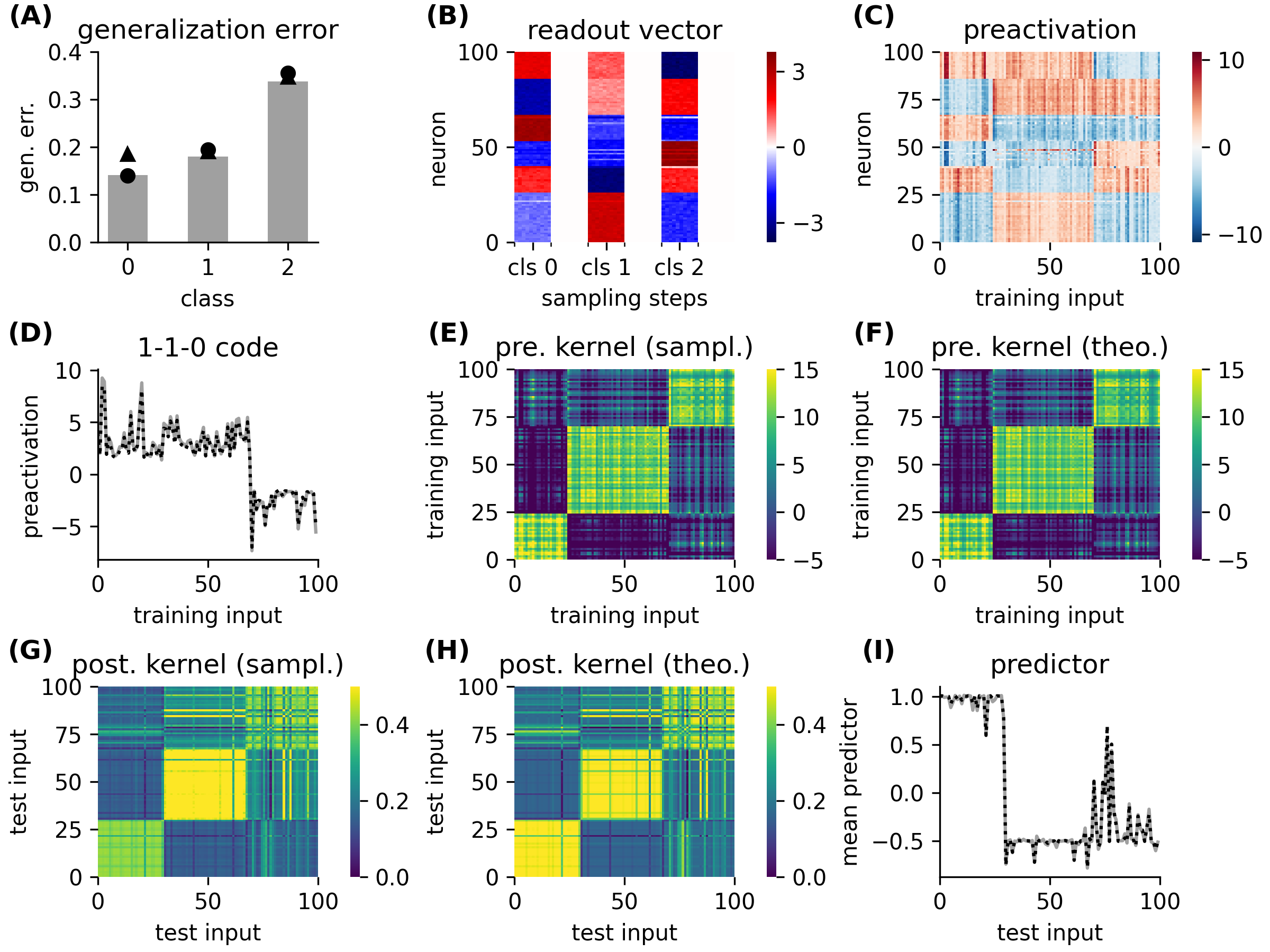}

\caption{Generalization of one hidden layer sigmoidal networks on projected
MNIST. Identical to Fig.~\ref{fig:sigmoidal_mnist_suppmat} except
that the data was randomly projected to a $N_{0}=50$ dimensional
subspace. \label{fig:sigmoidal_mnist_projected}}
\end{figure}

Brief summary and discussion of the supplemental figures (since we
include both pre- and postactivations, we here refer to the latter
as postactivations instead of activations):
\begin{itemize}
\item Fig.~\ref{fig:linear_code_suppmat}: Extended view of the results
presented in Fig.~\ref{fig:linear_code} including the theoretical
kernel (C,F) and a comparison of the readout weight posterior across
samples for a single neuron (G) and across neurons for a given sample
(H).
\item Fig.~\ref{fig:sigmoidal_code_suppmat}: Extended view of the results
presented in Fig.~\ref{fig:sigmoidal_code} including a comparison
of the readout weight posterior across samples for a single neuron
(A) and across neurons for a given sample (B), the training preactivations
(E), and the theoretical kernel (I).
\item Fig.~\ref{fig:sigmoidal_code_multisolution}: Multiple self-consistent
theoretical solutions for the setup from Fig.~\ref{fig:sigmoidal_code}.
Each solution corresponds to a saddle point of the $t$ integral.
Comparing $E(t)$ shows that the sampled solution in Fig.~\ref{fig:sigmoidal_code}
has the lowest energy (A). The lowest energy solution does not correspond
to the minimal number of codes needed to solve the task: the lowest
energy solution comprises 4 codes (B); the other solutions comprise
only 3 codes (C,D).
\item Fig.~\ref{fig:sigmoidal_code_temp}: Temperature dependent phase
transition of the coding scheme from Fig.~\ref{fig:sigmoidal_code}.
At a high (rescaled) temperature $T=0.1$ the activations do not exhibit
a code and the readout weight posterior is unimodal (A,E,I). At a
lower temperature, a coding scheme emerges: $T=0.08$ is still above
the critical temperature (B,F,J) but at $T=0.07$ a coding scheme
emerges involving a subset of the neurons and the readout weight posterior
becomes multimodal (C,G,K). Further lowering the temperature to $T=0.01$
leads to the solution identical to the one shown in Fig.~\ref{fig:sigmoidal_code}
at $T=10^{-4}$.
\item Fig.~\ref{fig:sigmoidal_langevin}: Approach to equilibrium using
Langevin dynamics for the setup from Fig.~\ref{fig:sigmoidal_code}.
Starting from initial conditions sampled from the prior, the Langevin
dynamics first rapidly decrease the loss (A; note the logarithmic
x-axis) until the solution space is approximately reached. Afterwards,
the dynamics decrease the regularizer imposed by the prior while staying
at the solution space (B) until the final redundant coding scheme
is reached (C,F). After the first phase of approaching the solution
space, a coding scheme starts to emerge (D). During the minimization
of the regularizer the coding scheme is sharpened (E) until it reaches
the final solution (F).
\item Fig.~\ref{fig:sigmoidal_dynamics}: Gradient ascent on the log posterior
for the setup from Fig.~\ref{fig:sigmoidal_code}, i.e., identical
to Fig.~\ref{fig:sigmoidal_langevin} except that the dynamics are
noiseless. The dynamics go through the same phases of loss minimization
(A) and minimization of the regularizer at the solution space (B).
The coding scheme is identical to the Langevin case except for missing
variability due to the missing noise (C,F). This, however, is dependent
on the random initialization; for other random initializations there
can be a few neurons with a deviating code (not shown). Similar to
Langevin dynamics a code appears after the first phase of approaching
the solution space and is further sharpened during the minimization
of the regularizer (D,E,F).
\item Fig.~\ref{fig:relu_code_suppmat}: Extended view of the results presented
in Fig.~\ref{fig:relu_code} including a comparison of the readout
weight posterior across samples for a single neuron (A) and across
neurons for a given sample (B), the training preactivations of the
outlier neurons (E), and the theoretical kernel (I).
\item Fig.~\ref{fig:relu_langevin}: Approach to equilibrium using Langevin
dynamics for the setup from Fig.~\ref{fig:relu_code}. Similar to
the sigmoidal case, the dynamics go through the phases of loss minimization
(A) and minimization of the regularizer at the solution space (B)
until the final sparse code is reached (C,F). After the first phase
of approaching the solution space, a coding scheme starts to emerge
which is not yet sparse (D). During the minimization of the regularizer
the coding scheme becomes sparser (E) until it reaches the final solution
(F).
\item Fig.~\ref{fig:relu_dynamics}: Gradient ascent on the log posterior
for the setup from Fig.~\ref{fig:relu_code}, i.e., identical to
Fig.~\ref{fig:relu_langevin} except that the dynamics are noiseless.
The dynamics go through the same phases of loss minimization (A) and
minimization of the regularizer at the solution space (B). However,
unlike in the sigmoidal case, the final solution is different from
the Langevin case (C,F). This holds for all tested random initializations.
Similar to Langevin dynamics a coding scheme involving all neurons
appears after the first phase of approaching the solution space and
is further sharpened during the minimization of the regularizer (D,E,F).
However, the final solution is less sparse than the solution based
on Langevin dynamics. This is likely due to local minima (A) which
cannot be overcome without noise.
\item Fig.~\ref{fig:linear_mnist_suppmat}: Extended view of the results
presented in Fig.~\ref{fig:linear_mnist} including training and
test kernels in all layers from theory and sampling (D-I).
\item Fig.~\ref{fig:sigmoidal_mnist_suppmat}: Extended view of the results
presented in Fig.~\ref{fig:sigmoidal_mnist} including the training
preactivations (C), training preactivations of neurons with 0-0-1
code (D), the training preactivation kernel from sampling (E) and
theory (F), and the test postactivation kernel from sampling (G) and
theory (H). Note that the variability of the preactivations across
inputs (C,D) is not apparent in the postactivations (Fig.~\ref{fig:sigmoidal_mnist}(B))
since the preactivations saturate the sigmoidal nonlinearity; correspondingly
the preactivation kernel exhibits more structure (E,F).
\item Fig.~\ref{fig:sigmoidal_mnist_projected}: Application of the theory
to the regime $P>N_{0}$. Identical to Fig.~\ref{fig:sigmoidal_mnist_suppmat}
except that the inputs are randomly projected to a $N_{0}=50$ dimensional
subspace such that $P=2N_{0}$.
\item Fig.~\ref{fig:sigmoidal_cifar}: Application of the theory to three
categories of gray scaled CIFAR10 instead of MNIST. Identical to Fig.~\ref{fig:sigmoidal_mnist_suppmat}
except for the replacement of MNIST by gray scaled CIFAR10 inputs.
While the coding scheme on training data is identical for CIFAR10
(B,C) and MNIST (Fig.~\ref{fig:sigmoidal_mnist_suppmat}(B,C)), the
test kernels are drastically different and exhibit no structure corresponding
to the task (G,H) and the predictor is almost random (I), leading
to a high generalization error (A). Put differently, the solution
fits the training data but does not generalize.
\item Fig.~\ref{fig:sigmoidal_code_multilayer_mnist}: Drifting representation
in the first layer of a two hidden layer sigmoidal network on MNIST.
Similar to Fig.~\ref{fig:sigmoidal_code_multilayer} except that
MNIST is used instead of the toy task.
\item Fig.~\ref{fig:sigmoidal_cifar_full}: $L=2$ layer sigmoidal network
with $N=1000$ neurons on all classes of CIFAR10 using the full size
$P=50000$ of the training data set. The last layer postactivations
display a clear coding of the task on the training inputs. These representations
generalize to varying extend to test inputs, leading to an overall
accuracy of $0.45$ on the full validation set.
\item Fig.~\ref{fig:relu_mnist_suppmat}: Extended view of the results
presented in Fig.~\ref{fig:relu_mnist} including the training preactivations
(C), the training preactivation kernel from sampling (E) and theory
(F), and the test postactivation kernel from sampling (G) and theory
(H). Note that the preactivations are negative on inputs belonging
to classes that the neuron does not code (C), leading to zero postactivations
on those inputs.
\end{itemize}
\clearpage{}

\begin{figure}
\includegraphics[width=0.5\columnwidth]{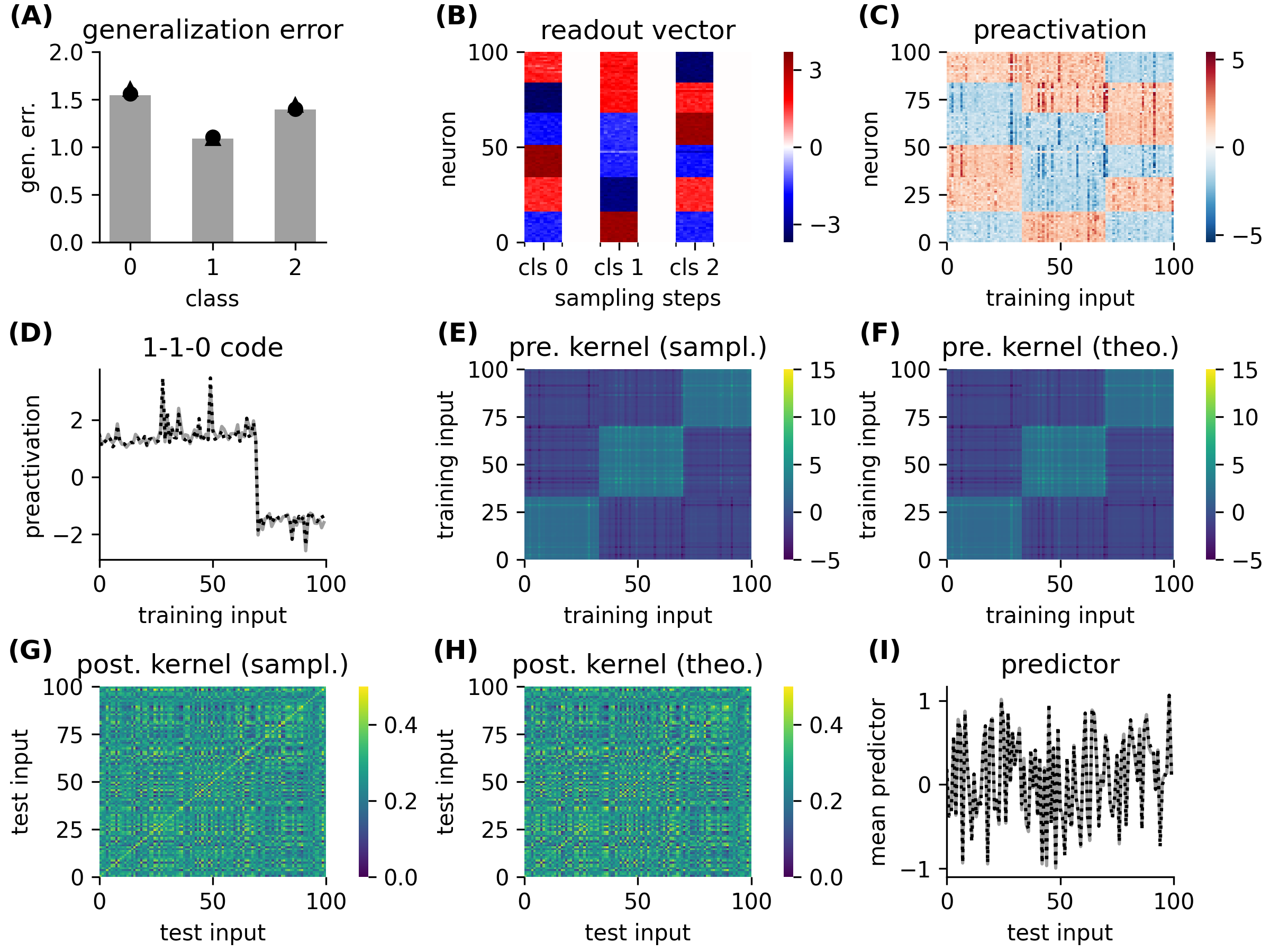}

\caption{Generalization of one hidden layer sigmoidal networks on grayscale
CIFAR10. Identical to Fig.~\ref{fig:sigmoidal_mnist_suppmat} except
that CIFAR10 instead of MNIST was used. The images were converted
to grayscale such that $N_{0}=1024$. \label{fig:sigmoidal_cifar}}
\end{figure}
\begin{figure}
\includegraphics[width=0.5\columnwidth]{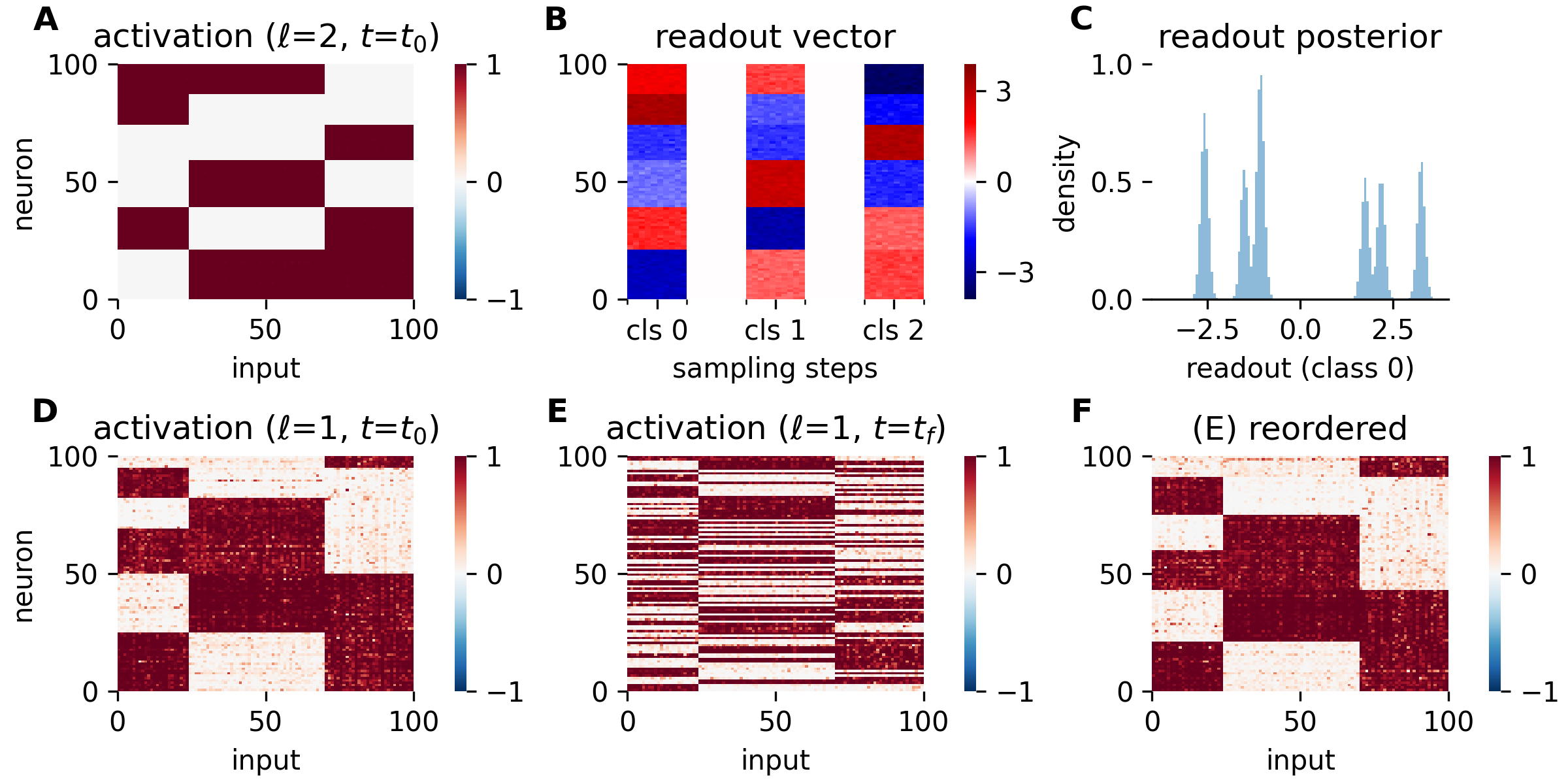}

\caption{Coding scheme in two layer sigmoidal networks on MNIST. (\textbf{A})
Postactivations of all second layer neurons on all training inputs
for a given weight sample. (\textbf{B}) Samples of the readout weights
of all three classes. (\textbf{C}) Readout posterior of first class.
(\textbf{D},\textbf{E}) Postactivations of all first layer neurons
on all training inputs using first (D) and last (E) weight sample.
(\textbf{F}) Postactivations from (E) but neurons are reordered. Parameters:
$N=P=100$, $N_{0}=784$, classes 0, 1, 2 assigned randomly with probability
$1/3$, targets $y_{+}=1$ and $y_{-}=-1/2$. \label{fig:sigmoidal_code_multilayer_mnist}}
\end{figure}

\begin{figure}
\includegraphics[width=0.5\columnwidth]{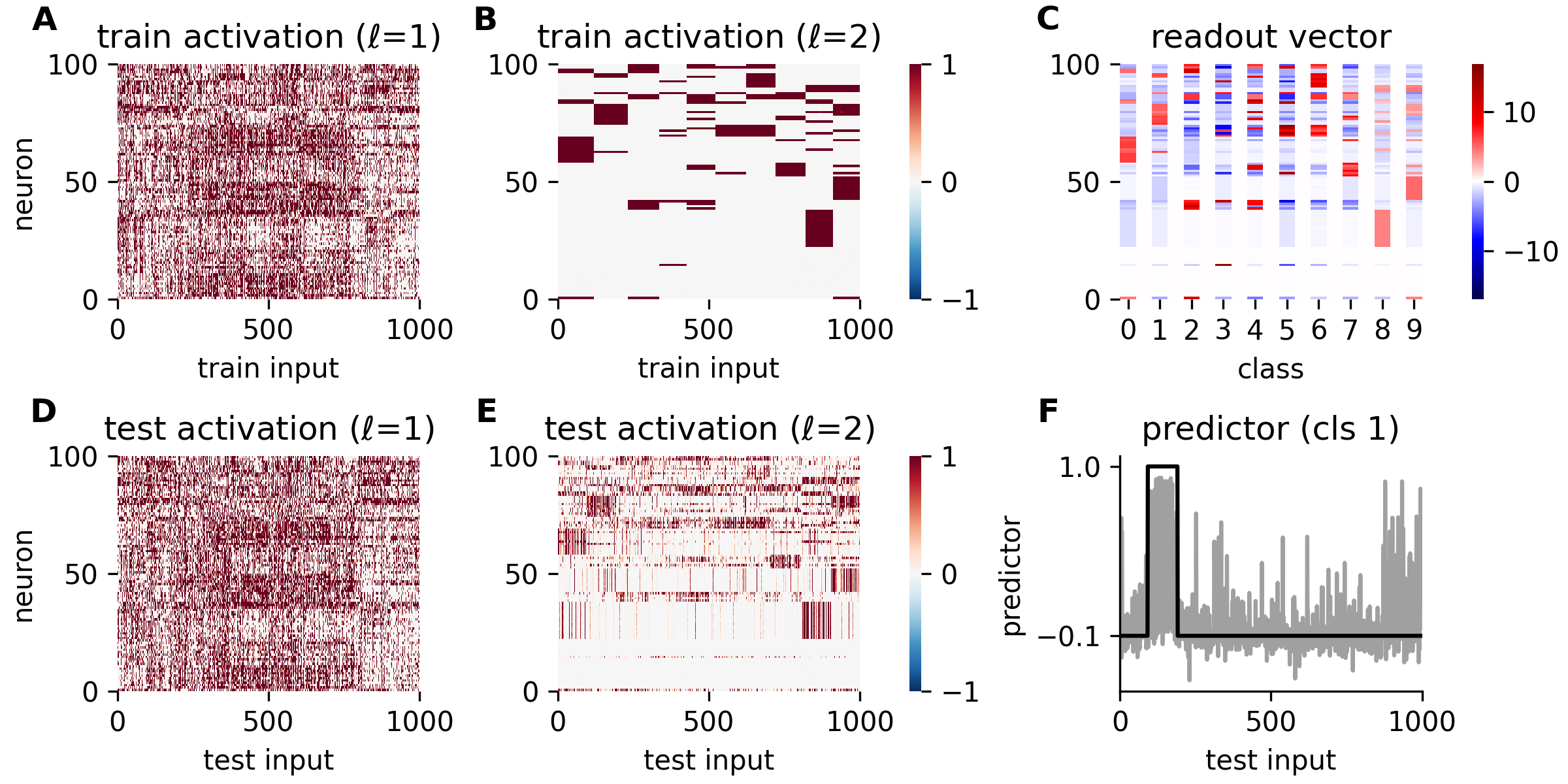}

\caption{Generalization of two hidden layer sigmoidal networks on full CIFAR10
data set. (\textbf{A},\textbf{B},\textbf{D},\textbf{E}) Postactivations
of $100$ randomly selected first (A,D) and second (B,E) layer neurons
on $1000$ randomly selected training (A,B) and test (D,E) inputs
for a given weight sample. (\textbf{C}) Readout weights of all three
classes for the $100$ selected neurons. (\textbf{F}) Predictor for
class 1 using the given weight sample (gray) and training target (black).
Average accuracy for all classes $0.45$. Parameters: $N=1000$, $P=50000$
$N_{0}=3072$, all classes, targets $y_{+}=1$ and $y_{-}=-1/10$.
\label{fig:sigmoidal_cifar_full}}
\end{figure}

\begin{figure}
\includegraphics[width=0.5\columnwidth]{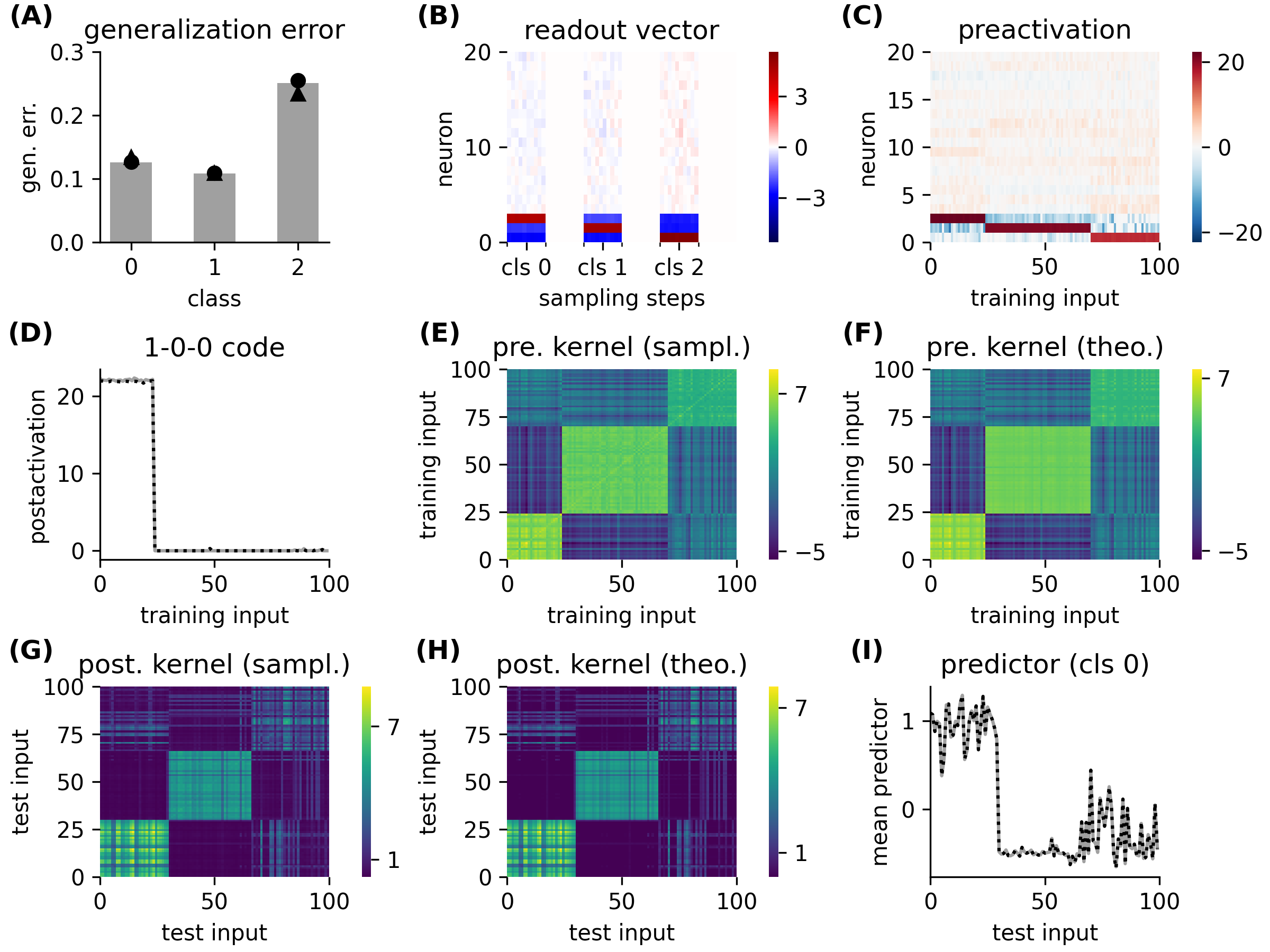}

\caption{Generalization of one hidden layer ReLU networks on MNIST. (\textbf{A})
Generalization error for each class averaged over $1,000$ test inputs
from sampling (gray bars), theory (Eq.~(\ref{eq:predictor_sigmoidal}),
black circles), and GP theory (back triangles). (\textbf{B}) Sample
of the readout weights for outlier neurons. (\textbf{C}) Preactivations
of outlier neurons on all training inputs for a given weight sample.
(\textbf{D}) Averaged postactivation of neurons with 1-0-0 code from
sampling (gray) and theory (black dashed). (\textbf{E},\textbf{F})
Preactivation kernel on training data from sampling (E) and theory
(F). (\textbf{G},\textbf{H}) Postactivation kernel on test data from
sampling (H) and theory (I). (\textbf{I}) Mean predictor for class
0 from sampling (gray) and theory (black dashed). Parameters: $N=P=100$,
$N_{0}=784$, classes 0, 1, 2 assigned randomly with probability $1/3$,
targets $y_{+}=1$ and $y_{-}=-1/2$. \label{fig:relu_mnist_suppmat}}
\end{figure}


\begin{thebibliography}{80}%
\makeatletter
\providecommand \@ifxundefined [1]{%
 \@ifx{#1\undefined}
}%
\providecommand \@ifnum [1]{%
 \ifnum #1\expandafter \@firstoftwo
 \else \expandafter \@secondoftwo
 \fi
}%
\providecommand \@ifx [1]{%
 \ifx #1\expandafter \@firstoftwo
 \else \expandafter \@secondoftwo
 \fi
}%
\providecommand \natexlab [1]{#1}%
\providecommand \enquote  [1]{``#1''}%
\providecommand \bibnamefont  [1]{#1}%
\providecommand \bibfnamefont [1]{#1}%
\providecommand \citenamefont [1]{#1}%
\providecommand \href@noop [0]{\@secondoftwo}%
\providecommand \href [0]{\begingroup \@sanitize@url \@href}%
\providecommand \@href[1]{\@@startlink{#1}\@@href}%
\providecommand \@@href[1]{\endgroup#1\@@endlink}%
\providecommand \@sanitize@url [0]{\catcode `\\12\catcode `\$12\catcode
  `\&12\catcode `\#12\catcode `\^12\catcode `\_12\catcode `\%12\relax}%
\providecommand \@@startlink[1]{}%
\providecommand \@@endlink[0]{}%
\providecommand \url  [0]{\begingroup\@sanitize@url \@url }%
\providecommand \@url [1]{\endgroup\@href {#1}{\urlprefix }}%
\providecommand \urlprefix  [0]{URL }%
\providecommand \Eprint [0]{\href }%
\providecommand \doibase [0]{https://doi.org/}%
\providecommand \selectlanguage [0]{\@gobble}%
\providecommand \bibinfo  [0]{\@secondoftwo}%
\providecommand \bibfield  [0]{\@secondoftwo}%
\providecommand \translation [1]{[#1]}%
\providecommand \BibitemOpen [0]{}%
\providecommand \bibitemStop [0]{}%
\providecommand \bibitemNoStop [0]{.\EOS\space}%
\providecommand \EOS [0]{\spacefactor3000\relax}%
\providecommand \BibitemShut  [1]{\csname bibitem#1\endcsname}%
\let\auto@bib@innerbib\@empty
\bibitem [{\citenamefont {Bengio}\ \emph {et~al.}(2013)\citenamefont {Bengio},
  \citenamefont {Courville},\ and\ \citenamefont
  {Vincent}}]{10.1109/TPAMI.2013.50}%
  \BibitemOpen
  \bibfield  {author} {\bibinfo {author} {\bibfnamefont {Y.}~\bibnamefont
  {Bengio}}, \bibinfo {author} {\bibfnamefont {A.}~\bibnamefont {Courville}},\
  and\ \bibinfo {author} {\bibfnamefont {P.}~\bibnamefont {Vincent}},\ }\href
  {https://doi.org/10.1109/TPAMI.2013.50} {\bibfield  {journal} {\bibinfo
  {journal} {IEEE Trans. Pattern Anal. Mach. Intell.}\ }\textbf {\bibinfo
  {volume} {35}},\ \bibinfo {pages} {1798–1828} (\bibinfo {year}
  {2013})}\BibitemShut {NoStop}%
\bibitem [{\citenamefont {LeCun}\ \emph {et~al.}(2015)\citenamefont {LeCun},
  \citenamefont {Bengio},\ and\ \citenamefont {Hinton}}]{Lecun2015deep}%
  \BibitemOpen
  \bibfield  {author} {\bibinfo {author} {\bibfnamefont {Y.}~\bibnamefont
  {LeCun}}, \bibinfo {author} {\bibfnamefont {Y.}~\bibnamefont {Bengio}},\ and\
  \bibinfo {author} {\bibfnamefont {G.}~\bibnamefont {Hinton}},\ }\href@noop {}
  {\bibfield  {journal} {\bibinfo  {journal} {Nature}\ }\textbf {\bibinfo
  {volume} {521}},\ \bibinfo {pages} {436} (\bibinfo {year}
  {2015})}\BibitemShut {NoStop}%
\bibitem [{\citenamefont {Goodfellow}\ \emph {et~al.}(2016)\citenamefont
  {Goodfellow}, \citenamefont {Bengio},\ and\ \citenamefont
  {Courville}}]{Goodfellow-et-al-2016}%
  \BibitemOpen
  \bibfield  {author} {\bibinfo {author} {\bibfnamefont {I.}~\bibnamefont
  {Goodfellow}}, \bibinfo {author} {\bibfnamefont {Y.}~\bibnamefont {Bengio}},\
  and\ \bibinfo {author} {\bibfnamefont {A.}~\bibnamefont {Courville}},\
  }\href@noop {} {\emph {\bibinfo {title} {Deep Learning}}}\ (\bibinfo
  {publisher} {MIT Press},\ \bibinfo {year} {2016})\ \bibinfo {note}
  {\url{http://www.deeplearningbook.org}}\BibitemShut {NoStop}%
\bibitem [{\citenamefont {Zhang}\ \emph {et~al.}(2017)\citenamefont {Zhang},
  \citenamefont {Bengio}, \citenamefont {Hardt}, \citenamefont {Recht},\ and\
  \citenamefont {Vinyals}}]{zhang2017understanding}%
  \BibitemOpen
  \bibfield  {author} {\bibinfo {author} {\bibfnamefont {C.}~\bibnamefont
  {Zhang}}, \bibinfo {author} {\bibfnamefont {S.}~\bibnamefont {Bengio}},
  \bibinfo {author} {\bibfnamefont {M.}~\bibnamefont {Hardt}}, \bibinfo
  {author} {\bibfnamefont {B.}~\bibnamefont {Recht}},\ and\ \bibinfo {author}
  {\bibfnamefont {O.}~\bibnamefont {Vinyals}},\ }in\ \href
  {https://openreview.net/forum?id=Sy8gdB9xx} {\emph {\bibinfo {booktitle}
  {International Conference on Learning Representations}}}\ (\bibinfo {year}
  {2017})\BibitemShut {NoStop}%
\bibitem [{\citenamefont {Zhang}\ \emph {et~al.}(2021)\citenamefont {Zhang},
  \citenamefont {Bengio}, \citenamefont {Hardt}, \citenamefont {Recht},\ and\
  \citenamefont {Vinyals}}]{10.1145/3446776}%
  \BibitemOpen
  \bibfield  {author} {\bibinfo {author} {\bibfnamefont {C.}~\bibnamefont
  {Zhang}}, \bibinfo {author} {\bibfnamefont {S.}~\bibnamefont {Bengio}},
  \bibinfo {author} {\bibfnamefont {M.}~\bibnamefont {Hardt}}, \bibinfo
  {author} {\bibfnamefont {B.}~\bibnamefont {Recht}},\ and\ \bibinfo {author}
  {\bibfnamefont {O.}~\bibnamefont {Vinyals}},\ }\href
  {https://doi.org/10.1145/3446776} {\bibfield  {journal} {\bibinfo  {journal}
  {Commun. ACM}\ }\textbf {\bibinfo {volume} {64}},\ \bibinfo {pages}
  {107–115} (\bibinfo {year} {2021})}\BibitemShut {NoStop}%
\bibitem [{\citenamefont {Belkin}\ \emph {et~al.}(2019)\citenamefont {Belkin},
  \citenamefont {Hsu}, \citenamefont {Ma},\ and\ \citenamefont
  {Mandal}}]{doi:10.1073/pnas.1903070116}%
  \BibitemOpen
  \bibfield  {author} {\bibinfo {author} {\bibfnamefont {M.}~\bibnamefont
  {Belkin}}, \bibinfo {author} {\bibfnamefont {D.}~\bibnamefont {Hsu}},
  \bibinfo {author} {\bibfnamefont {S.}~\bibnamefont {Ma}},\ and\ \bibinfo
  {author} {\bibfnamefont {S.}~\bibnamefont {Mandal}},\ }\href
  {https://doi.org/10.1073/pnas.1903070116} {\bibfield  {journal} {\bibinfo
  {journal} {Proceedings of the National Academy of Sciences}\ }\textbf
  {\bibinfo {volume} {116}},\ \bibinfo {pages} {15849} (\bibinfo {year}
  {2019})}\BibitemShut {NoStop}%
\bibitem [{\citenamefont {Nakkiran}\ \emph {et~al.}(2020)\citenamefont
  {Nakkiran}, \citenamefont {Kaplun}, \citenamefont {Bansal}, \citenamefont
  {Yang}, \citenamefont {Barak},\ and\ \citenamefont
  {Sutskever}}]{Nakkiran2020Deep}%
  \BibitemOpen
  \bibfield  {author} {\bibinfo {author} {\bibfnamefont {P.}~\bibnamefont
  {Nakkiran}}, \bibinfo {author} {\bibfnamefont {G.}~\bibnamefont {Kaplun}},
  \bibinfo {author} {\bibfnamefont {Y.}~\bibnamefont {Bansal}}, \bibinfo
  {author} {\bibfnamefont {T.}~\bibnamefont {Yang}}, \bibinfo {author}
  {\bibfnamefont {B.}~\bibnamefont {Barak}},\ and\ \bibinfo {author}
  {\bibfnamefont {I.}~\bibnamefont {Sutskever}},\ }in\ \href
  {https://openreview.net/forum?id=B1g5sA4twr} {\emph {\bibinfo {booktitle}
  {International Conference on Learning Representations}}}\ (\bibinfo {year}
  {2020})\BibitemShut {NoStop}%
\bibitem [{\citenamefont {Belkin}(2021)}]{Belkin_2021}%
  \BibitemOpen
  \bibfield  {author} {\bibinfo {author} {\bibfnamefont {M.}~\bibnamefont
  {Belkin}},\ }\href {https://doi.org/10.1017/S0962492921000039} {\bibfield
  {journal} {\bibinfo  {journal} {Acta Numerica}\ }\textbf {\bibinfo {volume}
  {30}},\ \bibinfo {pages} {203–248} (\bibinfo {year} {2021})}\BibitemShut
  {NoStop}%
\bibitem [{\citenamefont {Shwartz-Ziv}\ \emph {et~al.}(2024)\citenamefont
  {Shwartz-Ziv}, \citenamefont {Goldblum}, \citenamefont {Bansal},
  \citenamefont {Bruss}, \citenamefont {LeCun},\ and\ \citenamefont
  {Wilson}}]{shwartzziv2024just}%
  \BibitemOpen
  \bibfield  {author} {\bibinfo {author} {\bibfnamefont {R.}~\bibnamefont
  {Shwartz-Ziv}}, \bibinfo {author} {\bibfnamefont {M.}~\bibnamefont
  {Goldblum}}, \bibinfo {author} {\bibfnamefont {A.}~\bibnamefont {Bansal}},
  \bibinfo {author} {\bibfnamefont {C.~B.}\ \bibnamefont {Bruss}}, \bibinfo
  {author} {\bibfnamefont {Y.}~\bibnamefont {LeCun}},\ and\ \bibinfo {author}
  {\bibfnamefont {A.~G.}\ \bibnamefont {Wilson}},\ }\href@noop {} {\bibinfo
  {title} {Just how flexible are neural networks in practice?}} (\bibinfo
  {year} {2024}),\ \Eprint {https://arxiv.org/abs/2406.11463}
  {arXiv:2406.11463} \BibitemShut {NoStop}%
\bibitem [{\citenamefont {MacKay}(2003)}]{Mackay2003}%
  \BibitemOpen
  \bibfield  {author} {\bibinfo {author} {\bibfnamefont {D.~J.}\ \bibnamefont
  {MacKay}},\ }\href@noop {} {\emph {\bibinfo {title} {Information theory,
  inference and learning algorithms}}}\ (\bibinfo  {publisher} {Cambridge
  university press},\ \bibinfo {year} {2003})\BibitemShut {NoStop}%
\bibitem [{\citenamefont {Bahri}\ \emph {et~al.}(2020)\citenamefont {Bahri},
  \citenamefont {Kadmon}, \citenamefont {Pennington}, \citenamefont
  {Schoenholz}, \citenamefont {Sohl-Dickstein},\ and\ \citenamefont
  {Ganguli}}]{Bahri20_501}%
  \BibitemOpen
  \bibfield  {author} {\bibinfo {author} {\bibfnamefont {Y.}~\bibnamefont
  {Bahri}}, \bibinfo {author} {\bibfnamefont {J.}~\bibnamefont {Kadmon}},
  \bibinfo {author} {\bibfnamefont {J.}~\bibnamefont {Pennington}}, \bibinfo
  {author} {\bibfnamefont {S.~S.}\ \bibnamefont {Schoenholz}}, \bibinfo
  {author} {\bibfnamefont {J.}~\bibnamefont {Sohl-Dickstein}},\ and\ \bibinfo
  {author} {\bibfnamefont {S.}~\bibnamefont {Ganguli}},\ }\href
  {https://doi.org/10.1146/annurev-conmatphys-031119-050745} {\bibfield
  {journal} {\bibinfo  {journal} {Annu. Rev. Condens. Matter Phys.}\ }\textbf
  {\bibinfo {volume} {11}},\ \bibinfo {pages} {501} (\bibinfo {year}
  {2020})}\BibitemShut {NoStop}%
\bibitem [{\citenamefont {Neal}(1996)}]{Neal96}%
  \BibitemOpen
  \bibfield  {author} {\bibinfo {author} {\bibfnamefont {R.~M.}\ \bibnamefont
  {Neal}},\ }\href {https://doi.org/10.1007/978-1-4612-0745-0} {\emph {\bibinfo
  {title} {Bayesian Learning for Neural Networks}}}\ (\bibinfo  {publisher}
  {Springer New York},\ \bibinfo {year} {1996})\BibitemShut {NoStop}%
\bibitem [{\citenamefont {Williams}(1996)}]{Williams96_ae5e3ce4}%
  \BibitemOpen
  \bibfield  {author} {\bibinfo {author} {\bibfnamefont {C.}~\bibnamefont
  {Williams}},\ }in\ \href
  {https://proceedings.neurips.cc/paper/1996/file/ae5e3ce40e0404a45ecacaaf05e5f735-Paper.pdf}
  {\emph {\bibinfo {booktitle} {Advances in Neural Information Processing
  Systems}}},\ Vol.~\bibinfo {volume} {9},\ \bibinfo {editor} {edited by\
  \bibinfo {editor} {\bibfnamefont {M.}~\bibnamefont {Mozer}}, \bibinfo
  {editor} {\bibfnamefont {M.}~\bibnamefont {Jordan}},\ and\ \bibinfo {editor}
  {\bibfnamefont {T.}~\bibnamefont {Petsche}}}\ (\bibinfo  {publisher} {MIT
  Press},\ \bibinfo {year} {1996})\BibitemShut {NoStop}%
\bibitem [{\citenamefont {Lee}\ \emph {et~al.}(2018)\citenamefont {Lee},
  \citenamefont {Sohl-Dickstein}, \citenamefont {Pennington}, \citenamefont
  {Novak}, \citenamefont {Schoenholz},\ and\ \citenamefont {Bahri}}]{Lee18}%
  \BibitemOpen
  \bibfield  {author} {\bibinfo {author} {\bibfnamefont {J.}~\bibnamefont
  {Lee}}, \bibinfo {author} {\bibfnamefont {J.}~\bibnamefont {Sohl-Dickstein}},
  \bibinfo {author} {\bibfnamefont {J.}~\bibnamefont {Pennington}}, \bibinfo
  {author} {\bibfnamefont {R.}~\bibnamefont {Novak}}, \bibinfo {author}
  {\bibfnamefont {S.}~\bibnamefont {Schoenholz}},\ and\ \bibinfo {author}
  {\bibfnamefont {Y.}~\bibnamefont {Bahri}},\ }in\ \href
  {https://openreview.net/forum?id=B1EA-M-0Z} {\emph {\bibinfo {booktitle}
  {International Conference on Learning Representations}}}\ (\bibinfo {year}
  {2018})\BibitemShut {NoStop}%
\bibitem [{\citenamefont {Matthews}\ \emph {et~al.}(2018)\citenamefont
  {Matthews}, \citenamefont {Hron}, \citenamefont {Rowland}, \citenamefont
  {Turner},\ and\ \citenamefont {Ghahramani}}]{Matthews18}%
  \BibitemOpen
  \bibfield  {author} {\bibinfo {author} {\bibfnamefont {A.~G. d.~G.}\
  \bibnamefont {Matthews}}, \bibinfo {author} {\bibfnamefont {J.}~\bibnamefont
  {Hron}}, \bibinfo {author} {\bibfnamefont {M.}~\bibnamefont {Rowland}},
  \bibinfo {author} {\bibfnamefont {R.~E.}\ \bibnamefont {Turner}},\ and\
  \bibinfo {author} {\bibfnamefont {Z.}~\bibnamefont {Ghahramani}},\ }in\ \href
  {https://openreview.net/forum?id=H1-nGgWC-} {\emph {\bibinfo {booktitle}
  {International Conference on Learning Representations}}}\ (\bibinfo {year}
  {2018})\BibitemShut {NoStop}%
\bibitem [{\citenamefont {Yang}(2019)}]{NEURIPS2019_5e69fda3}%
  \BibitemOpen
  \bibfield  {author} {\bibinfo {author} {\bibfnamefont {G.}~\bibnamefont
  {Yang}},\ }in\ \href@noop {} {\emph {\bibinfo {booktitle} {Advances in Neural
  Information Processing Systems}}},\ Vol.~\bibinfo {volume} {32},\ \bibinfo
  {editor} {edited by\ \bibinfo {editor} {\bibfnamefont {H.}~\bibnamefont
  {Wallach}}, \bibinfo {editor} {\bibfnamefont {H.}~\bibnamefont {Larochelle}},
  \bibinfo {editor} {\bibfnamefont {A.}~\bibnamefont {Beygelzimer}}, \bibinfo
  {editor} {\bibfnamefont {F.}~\bibnamefont {d\textquotesingle Alch\'{e}-Buc}},
  \bibinfo {editor} {\bibfnamefont {E.}~\bibnamefont {Fox}},\ and\ \bibinfo
  {editor} {\bibfnamefont {R.}~\bibnamefont {Garnett}}}\ (\bibinfo  {publisher}
  {Curran Associates, Inc.},\ \bibinfo {year} {2019})\BibitemShut {NoStop}%
\bibitem [{\citenamefont {Naveh}\ \emph {et~al.}(2021)\citenamefont {Naveh},
  \citenamefont {Ben~David}, \citenamefont {Sompolinsky},\ and\ \citenamefont
  {Ringel}}]{PhysRevE.104.064301}%
  \BibitemOpen
  \bibfield  {author} {\bibinfo {author} {\bibfnamefont {G.}~\bibnamefont
  {Naveh}}, \bibinfo {author} {\bibfnamefont {O.}~\bibnamefont {Ben~David}},
  \bibinfo {author} {\bibfnamefont {H.}~\bibnamefont {Sompolinsky}},\ and\
  \bibinfo {author} {\bibfnamefont {Z.}~\bibnamefont {Ringel}},\ }\href
  {https://doi.org/10.1103/PhysRevE.104.064301} {\bibfield  {journal} {\bibinfo
   {journal} {Phys. Rev. E}\ }\textbf {\bibinfo {volume} {104}},\ \bibinfo
  {pages} {064301} (\bibinfo {year} {2021})}\BibitemShut {NoStop}%
\bibitem [{\citenamefont {Segadlo}\ \emph {et~al.}(2022)\citenamefont
  {Segadlo}, \citenamefont {Epping}, \citenamefont {van Meegen}, \citenamefont
  {Dahmen}, \citenamefont {Krämer},\ and\ \citenamefont
  {Helias}}]{Segadlo_2022}%
  \BibitemOpen
  \bibfield  {author} {\bibinfo {author} {\bibfnamefont {K.}~\bibnamefont
  {Segadlo}}, \bibinfo {author} {\bibfnamefont {B.}~\bibnamefont {Epping}},
  \bibinfo {author} {\bibfnamefont {A.}~\bibnamefont {van Meegen}}, \bibinfo
  {author} {\bibfnamefont {D.}~\bibnamefont {Dahmen}}, \bibinfo {author}
  {\bibfnamefont {M.}~\bibnamefont {Krämer}},\ and\ \bibinfo {author}
  {\bibfnamefont {M.}~\bibnamefont {Helias}},\ }\href
  {https://doi.org/10.1088/1742-5468/ac8e57} {\bibfield  {journal} {\bibinfo
  {journal} {Journal of Statistical Mechanics: Theory and Experiment}\ }\textbf
  {\bibinfo {volume} {2022}},\ \bibinfo {pages} {103401} (\bibinfo {year}
  {2022})}\BibitemShut {NoStop}%
\bibitem [{\citenamefont {Hron}\ \emph {et~al.}(2022)\citenamefont {Hron},
  \citenamefont {Novak}, \citenamefont {Pennington},\ and\ \citenamefont
  {Sohl-Dickstein}}]{pmlr-v162-hron22a}%
  \BibitemOpen
  \bibfield  {author} {\bibinfo {author} {\bibfnamefont {J.}~\bibnamefont
  {Hron}}, \bibinfo {author} {\bibfnamefont {R.}~\bibnamefont {Novak}},
  \bibinfo {author} {\bibfnamefont {J.}~\bibnamefont {Pennington}},\ and\
  \bibinfo {author} {\bibfnamefont {J.}~\bibnamefont {Sohl-Dickstein}},\ }in\
  \href@noop {} {\emph {\bibinfo {booktitle} {Proceedings of the 39th
  International Conference on Machine Learning}}},\ \bibinfo {series}
  {Proceedings of Machine Learning Research}, Vol.\ \bibinfo {volume} {162},\
  \bibinfo {editor} {edited by\ \bibinfo {editor} {\bibfnamefont
  {K.}~\bibnamefont {Chaudhuri}}, \bibinfo {editor} {\bibfnamefont
  {S.}~\bibnamefont {Jegelka}}, \bibinfo {editor} {\bibfnamefont
  {L.}~\bibnamefont {Song}}, \bibinfo {editor} {\bibfnamefont {C.}~\bibnamefont
  {Szepesvari}}, \bibinfo {editor} {\bibfnamefont {G.}~\bibnamefont {Niu}},\
  and\ \bibinfo {editor} {\bibfnamefont {S.}~\bibnamefont {Sabato}}}\ (\bibinfo
   {publisher} {PMLR},\ \bibinfo {year} {2022})\ pp.\ \bibinfo {pages}
  {8926--8945}\BibitemShut {NoStop}%
\bibitem [{\citenamefont {Li}\ and\ \citenamefont
  {Sompolinsky}(2021)}]{Li21_031059}%
  \BibitemOpen
  \bibfield  {author} {\bibinfo {author} {\bibfnamefont {Q.}~\bibnamefont
  {Li}}\ and\ \bibinfo {author} {\bibfnamefont {H.}~\bibnamefont
  {Sompolinsky}},\ }\href {https://doi.org/10.1103/PhysRevX.11.031059}
  {\bibfield  {journal} {\bibinfo  {journal} {Physical Review X}\ }\textbf
  {\bibinfo {volume} {11}},\ \bibinfo {pages} {031059} (\bibinfo {year}
  {2021})}\BibitemShut {NoStop}%
\bibitem [{\citenamefont {Naveh}\ and\ \citenamefont
  {Ringel}(2021)}]{Naveh21_NeurIPS}%
  \BibitemOpen
  \bibfield  {author} {\bibinfo {author} {\bibfnamefont {G.}~\bibnamefont
  {Naveh}}\ and\ \bibinfo {author} {\bibfnamefont {Z.}~\bibnamefont {Ringel}},\
  }in\ \href {https://openreview.net/forum?id=vBYwwBxVcsE} {\emph {\bibinfo
  {booktitle} {Advances in Neural Information Processing Systems}}},\ \bibinfo
  {editor} {edited by\ \bibinfo {editor} {\bibfnamefont {A.}~\bibnamefont
  {Beygelzimer}}, \bibinfo {editor} {\bibfnamefont {Y.}~\bibnamefont
  {Dauphin}}, \bibinfo {editor} {\bibfnamefont {P.}~\bibnamefont {Liang}},\
  and\ \bibinfo {editor} {\bibfnamefont {J.~W.}\ \bibnamefont {Vaughan}}}\
  (\bibinfo {year} {2021})\BibitemShut {NoStop}%
\bibitem [{\citenamefont {Zavatone-Veth}\ \emph {et~al.}(2022)\citenamefont
  {Zavatone-Veth}, \citenamefont {Tong},\ and\ \citenamefont
  {Pehlevan}}]{ZavatoneVeth22_064118}%
  \BibitemOpen
  \bibfield  {author} {\bibinfo {author} {\bibfnamefont {J.~A.}\ \bibnamefont
  {Zavatone-Veth}}, \bibinfo {author} {\bibfnamefont {W.~L.}\ \bibnamefont
  {Tong}},\ and\ \bibinfo {author} {\bibfnamefont {C.}~\bibnamefont
  {Pehlevan}},\ }\href {https://doi.org/10.1103/PhysRevE.105.064118} {\bibfield
   {journal} {\bibinfo  {journal} {Phys. Rev. E}\ }\textbf {\bibinfo {volume}
  {105}},\ \bibinfo {pages} {064118} (\bibinfo {year} {2022})}\BibitemShut
  {NoStop}%
\bibitem [{\citenamefont {Cui}\ \emph {et~al.}(2023)\citenamefont {Cui},
  \citenamefont {Krzakala},\ and\ \citenamefont
  {Zdeborova}}]{pmlr-v202-cui23b}%
  \BibitemOpen
  \bibfield  {author} {\bibinfo {author} {\bibfnamefont {H.}~\bibnamefont
  {Cui}}, \bibinfo {author} {\bibfnamefont {F.}~\bibnamefont {Krzakala}},\ and\
  \bibinfo {author} {\bibfnamefont {L.}~\bibnamefont {Zdeborova}},\ }in\ \href
  {https://proceedings.mlr.press/v202/cui23b.html} {\emph {\bibinfo {booktitle}
  {Proceedings of the 40th International Conference on Machine Learning}}},\
  \bibinfo {series} {Proceedings of Machine Learning Research}, Vol.\ \bibinfo
  {volume} {202},\ \bibinfo {editor} {edited by\ \bibinfo {editor}
  {\bibfnamefont {A.}~\bibnamefont {Krause}}, \bibinfo {editor} {\bibfnamefont
  {E.}~\bibnamefont {Brunskill}}, \bibinfo {editor} {\bibfnamefont
  {K.}~\bibnamefont {Cho}}, \bibinfo {editor} {\bibfnamefont {B.}~\bibnamefont
  {Engelhardt}}, \bibinfo {editor} {\bibfnamefont {S.}~\bibnamefont {Sabato}},\
  and\ \bibinfo {editor} {\bibfnamefont {J.}~\bibnamefont {Scarlett}}}\
  (\bibinfo  {publisher} {PMLR},\ \bibinfo {year} {2023})\ pp.\ \bibinfo
  {pages} {6468--6521}\BibitemShut {NoStop}%
\bibitem [{\citenamefont {Hanin}\ and\ \citenamefont
  {Zlokapa}(2023)}]{doi:10.1073/pnas.2301345120}%
  \BibitemOpen
  \bibfield  {author} {\bibinfo {author} {\bibfnamefont {B.}~\bibnamefont
  {Hanin}}\ and\ \bibinfo {author} {\bibfnamefont {A.}~\bibnamefont
  {Zlokapa}},\ }\href {https://doi.org/10.1073/pnas.2301345120} {\bibfield
  {journal} {\bibinfo  {journal} {Proceedings of the National Academy of
  Sciences}\ }\textbf {\bibinfo {volume} {120}},\ \bibinfo {pages}
  {e2301345120} (\bibinfo {year} {2023})}\BibitemShut {NoStop}%
\bibitem [{\citenamefont {Pacelli}\ \emph {et~al.}(2023)\citenamefont
  {Pacelli}, \citenamefont {Ariosto}, \citenamefont {Pastore}, \citenamefont
  {Ginelli}, \citenamefont {Gherardi},\ and\ \citenamefont
  {Rotondo}}]{Pacelli2023}%
  \BibitemOpen
  \bibfield  {author} {\bibinfo {author} {\bibfnamefont {R.}~\bibnamefont
  {Pacelli}}, \bibinfo {author} {\bibfnamefont {S.}~\bibnamefont {Ariosto}},
  \bibinfo {author} {\bibfnamefont {M.}~\bibnamefont {Pastore}}, \bibinfo
  {author} {\bibfnamefont {F.}~\bibnamefont {Ginelli}}, \bibinfo {author}
  {\bibfnamefont {M.}~\bibnamefont {Gherardi}},\ and\ \bibinfo {author}
  {\bibfnamefont {P.}~\bibnamefont {Rotondo}},\ }\href
  {https://doi.org/10.1038/s42256-023-00767-6} {\bibfield  {journal} {\bibinfo
  {journal} {Nature Machine Intelligence}\ }\textbf {\bibinfo {volume} {5}},\
  \bibinfo {pages} {1497–1507} (\bibinfo {year} {2023})}\BibitemShut
  {NoStop}%
\bibitem [{\citenamefont {Fischer}\ \emph {et~al.}(2024)\citenamefont
  {Fischer}, \citenamefont {Lindner}, \citenamefont {Dahmen}, \citenamefont
  {Ringel}, \citenamefont {Krämer},\ and\ \citenamefont
  {Helias}}]{fischer2024critical}%
  \BibitemOpen
  \bibfield  {author} {\bibinfo {author} {\bibfnamefont {K.}~\bibnamefont
  {Fischer}}, \bibinfo {author} {\bibfnamefont {J.}~\bibnamefont {Lindner}},
  \bibinfo {author} {\bibfnamefont {D.}~\bibnamefont {Dahmen}}, \bibinfo
  {author} {\bibfnamefont {Z.}~\bibnamefont {Ringel}}, \bibinfo {author}
  {\bibfnamefont {M.}~\bibnamefont {Krämer}},\ and\ \bibinfo {author}
  {\bibfnamefont {M.}~\bibnamefont {Helias}},\ }\href@noop {} {\bibinfo {title}
  {Critical feature learning in deep neural networks}} (\bibinfo {year}
  {2024}),\ \Eprint {https://arxiv.org/abs/2405.10761} {arXiv:2405.10761
  [cond-mat.dis-nn]} \BibitemShut {NoStop}%
\bibitem [{\citenamefont {Chizat}\ \emph {et~al.}(2019)\citenamefont {Chizat},
  \citenamefont {Oyallon},\ and\ \citenamefont {Bach}}]{NEURIPS2019_ae614c55}%
  \BibitemOpen
  \bibfield  {author} {\bibinfo {author} {\bibfnamefont {L.}~\bibnamefont
  {Chizat}}, \bibinfo {author} {\bibfnamefont {E.}~\bibnamefont {Oyallon}},\
  and\ \bibinfo {author} {\bibfnamefont {F.}~\bibnamefont {Bach}},\ }in\ \href
  {https://proceedings.neurips.cc/paper_files/paper/2019/file/ae614c557843b1df326cb29c57225459-Paper.pdf}
  {\emph {\bibinfo {booktitle} {Advances in Neural Information Processing
  Systems}}},\ Vol.~\bibinfo {volume} {32},\ \bibinfo {editor} {edited by\
  \bibinfo {editor} {\bibfnamefont {H.}~\bibnamefont {Wallach}}, \bibinfo
  {editor} {\bibfnamefont {H.}~\bibnamefont {Larochelle}}, \bibinfo {editor}
  {\bibfnamefont {A.}~\bibnamefont {Beygelzimer}}, \bibinfo {editor}
  {\bibfnamefont {F.}~\bibnamefont {d\textquotesingle Alch\'{e}-Buc}}, \bibinfo
  {editor} {\bibfnamefont {E.}~\bibnamefont {Fox}},\ and\ \bibinfo {editor}
  {\bibfnamefont {R.}~\bibnamefont {Garnett}}}\ (\bibinfo  {publisher} {Curran
  Associates, Inc.},\ \bibinfo {year} {2019})\BibitemShut {NoStop}%
\bibitem [{\citenamefont {Woodworth}\ \emph {et~al.}(2020)\citenamefont
  {Woodworth}, \citenamefont {Gunasekar}, \citenamefont {Lee}, \citenamefont
  {Moroshko}, \citenamefont {Savarese}, \citenamefont {Golan}, \citenamefont
  {Soudry},\ and\ \citenamefont {Srebro}}]{pmlr-v125-woodworth20a}%
  \BibitemOpen
  \bibfield  {author} {\bibinfo {author} {\bibfnamefont {B.}~\bibnamefont
  {Woodworth}}, \bibinfo {author} {\bibfnamefont {S.}~\bibnamefont
  {Gunasekar}}, \bibinfo {author} {\bibfnamefont {J.~D.}\ \bibnamefont {Lee}},
  \bibinfo {author} {\bibfnamefont {E.}~\bibnamefont {Moroshko}}, \bibinfo
  {author} {\bibfnamefont {P.}~\bibnamefont {Savarese}}, \bibinfo {author}
  {\bibfnamefont {I.}~\bibnamefont {Golan}}, \bibinfo {author} {\bibfnamefont
  {D.}~\bibnamefont {Soudry}},\ and\ \bibinfo {author} {\bibfnamefont
  {N.}~\bibnamefont {Srebro}},\ }in\ \href
  {https://proceedings.mlr.press/v125/woodworth20a.html} {\emph {\bibinfo
  {booktitle} {Proceedings of Thirty Third Conference on Learning Theory}}},\
  \bibinfo {series} {Proceedings of Machine Learning Research}, Vol.\ \bibinfo
  {volume} {125},\ \bibinfo {editor} {edited by\ \bibinfo {editor}
  {\bibfnamefont {J.}~\bibnamefont {Abernethy}}\ and\ \bibinfo {editor}
  {\bibfnamefont {S.}~\bibnamefont {Agarwal}}}\ (\bibinfo  {publisher} {PMLR},\
  \bibinfo {year} {2020})\ pp.\ \bibinfo {pages} {3635--3673}\BibitemShut
  {NoStop}%
\bibitem [{\citenamefont {Geiger}\ \emph {et~al.}(2020)\citenamefont {Geiger},
  \citenamefont {Spigler}, \citenamefont {Jacot},\ and\ \citenamefont
  {Wyart}}]{Geiger_2020}%
  \BibitemOpen
  \bibfield  {author} {\bibinfo {author} {\bibfnamefont {M.}~\bibnamefont
  {Geiger}}, \bibinfo {author} {\bibfnamefont {S.}~\bibnamefont {Spigler}},
  \bibinfo {author} {\bibfnamefont {A.}~\bibnamefont {Jacot}},\ and\ \bibinfo
  {author} {\bibfnamefont {M.}~\bibnamefont {Wyart}},\ }\href
  {https://doi.org/10.1088/1742-5468/abc4de} {\bibfield  {journal} {\bibinfo
  {journal} {Journal of Statistical Mechanics: Theory and Experiment}\ }\textbf
  {\bibinfo {volume} {2020}},\ \bibinfo {pages} {113301} (\bibinfo {year}
  {2020})}\BibitemShut {NoStop}%
\bibitem [{\citenamefont {Yaida}(2020)}]{pmlr-v107-yaida20a}%
  \BibitemOpen
  \bibfield  {author} {\bibinfo {author} {\bibfnamefont {S.}~\bibnamefont
  {Yaida}},\ }in\ \href {https://proceedings.mlr.press/v107/yaida20a.html}
  {\emph {\bibinfo {booktitle} {Proceedings of The First Mathematical and
  Scientific Machine Learning Conference}}},\ \bibinfo {series} {Proceedings of
  Machine Learning Research}, Vol.\ \bibinfo {volume} {107},\ \bibinfo {editor}
  {edited by\ \bibinfo {editor} {\bibfnamefont {J.}~\bibnamefont {Lu}}\ and\
  \bibinfo {editor} {\bibfnamefont {R.}~\bibnamefont {Ward}}}\ (\bibinfo
  {publisher} {PMLR},\ \bibinfo {year} {2020})\ pp.\ \bibinfo {pages}
  {165--192}\BibitemShut {NoStop}%
\bibitem [{\citenamefont {Dyer}\ and\ \citenamefont
  {Gur-Ari}(2020)}]{Dyer2020Asymptotics}%
  \BibitemOpen
  \bibfield  {author} {\bibinfo {author} {\bibfnamefont {E.}~\bibnamefont
  {Dyer}}\ and\ \bibinfo {author} {\bibfnamefont {G.}~\bibnamefont {Gur-Ari}},\
  }in\ \href {https://openreview.net/forum?id=S1gFvANKDS} {\emph {\bibinfo
  {booktitle} {International Conference on Learning Representations}}}\
  (\bibinfo {year} {2020})\BibitemShut {NoStop}%
\bibitem [{\citenamefont {Zavatone-Veth}\ \emph {et~al.}(2021)\citenamefont
  {Zavatone-Veth}, \citenamefont {Canatar}, \citenamefont {Ruben},\ and\
  \citenamefont {Pehlevan}}]{ZavatoneVeth21_NeurIPS_II}%
  \BibitemOpen
  \bibfield  {author} {\bibinfo {author} {\bibfnamefont {J.~A.}\ \bibnamefont
  {Zavatone-Veth}}, \bibinfo {author} {\bibfnamefont {A.}~\bibnamefont
  {Canatar}}, \bibinfo {author} {\bibfnamefont {B.}~\bibnamefont {Ruben}},\
  and\ \bibinfo {author} {\bibfnamefont {C.}~\bibnamefont {Pehlevan}},\ }in\
  \href {https://openreview.net/forum?id=1oRFmD0Fl-5} {\emph {\bibinfo
  {booktitle} {Advances in Neural Information Processing Systems}}},\ \bibinfo
  {editor} {edited by\ \bibinfo {editor} {\bibfnamefont {A.}~\bibnamefont
  {Beygelzimer}}, \bibinfo {editor} {\bibfnamefont {Y.}~\bibnamefont
  {Dauphin}}, \bibinfo {editor} {\bibfnamefont {P.}~\bibnamefont {Liang}},\
  and\ \bibinfo {editor} {\bibfnamefont {J.~W.}\ \bibnamefont {Vaughan}}}\
  (\bibinfo {year} {2021})\BibitemShut {NoStop}%
\bibitem [{\citenamefont {Roberts}\ \emph {et~al.}(2022)\citenamefont
  {Roberts}, \citenamefont {Yaida},\ and\ \citenamefont {Hanin}}]{Roberts22}%
  \BibitemOpen
  \bibfield  {author} {\bibinfo {author} {\bibfnamefont {D.~A.}\ \bibnamefont
  {Roberts}}, \bibinfo {author} {\bibfnamefont {S.}~\bibnamefont {Yaida}},\
  and\ \bibinfo {author} {\bibfnamefont {B.}~\bibnamefont {Hanin}},\ }\href
  {https://doi.org/10.1017/9781009023405} {\emph {\bibinfo {title} {The
  Principles of Deep Learning Theory}}}\ (\bibinfo  {publisher} {Cambridge
  University Press},\ \bibinfo {year} {2022})\BibitemShut {NoStop}%
\bibitem [{\citenamefont {Mei}\ \emph {et~al.}(2018)\citenamefont {Mei},
  \citenamefont {Montanari},\ and\ \citenamefont {Nguyen}}]{Mei18_E7665}%
  \BibitemOpen
  \bibfield  {author} {\bibinfo {author} {\bibfnamefont {S.}~\bibnamefont
  {Mei}}, \bibinfo {author} {\bibfnamefont {A.}~\bibnamefont {Montanari}},\
  and\ \bibinfo {author} {\bibfnamefont {P.-M.}\ \bibnamefont {Nguyen}},\
  }\href@noop {} {\bibfield  {journal} {\bibinfo  {journal} {Proceedings of the
  National Academy of Sciences}\ }\textbf {\bibinfo {volume} {115}},\ \bibinfo
  {pages} {E7665} (\bibinfo {year} {2018})}\BibitemShut {NoStop}%
\bibitem [{\citenamefont {Chizat}\ and\ \citenamefont
  {Bach}(2018)}]{NEURIPS2018_a1afc58c}%
  \BibitemOpen
  \bibfield  {author} {\bibinfo {author} {\bibfnamefont {L.}~\bibnamefont
  {Chizat}}\ and\ \bibinfo {author} {\bibfnamefont {F.}~\bibnamefont {Bach}},\
  }in\ \href
  {https://proceedings.neurips.cc/paper_files/paper/2018/file/a1afc58c6ca9540d057299ec3016d726-Paper.pdf}
  {\emph {\bibinfo {booktitle} {Advances in Neural Information Processing
  Systems}}},\ Vol.~\bibinfo {volume} {31},\ \bibinfo {editor} {edited by\
  \bibinfo {editor} {\bibfnamefont {S.}~\bibnamefont {Bengio}}, \bibinfo
  {editor} {\bibfnamefont {H.}~\bibnamefont {Wallach}}, \bibinfo {editor}
  {\bibfnamefont {H.}~\bibnamefont {Larochelle}}, \bibinfo {editor}
  {\bibfnamefont {K.}~\bibnamefont {Grauman}}, \bibinfo {editor} {\bibfnamefont
  {N.}~\bibnamefont {Cesa-Bianchi}},\ and\ \bibinfo {editor} {\bibfnamefont
  {R.}~\bibnamefont {Garnett}}}\ (\bibinfo  {publisher} {Curran Associates,
  Inc.},\ \bibinfo {year} {2018})\BibitemShut {NoStop}%
\bibitem [{\citenamefont {Rotskoff}\ and\ \citenamefont
  {Vanden-Eijnden}(2018)}]{NEURIPS2018_196f5641}%
  \BibitemOpen
  \bibfield  {author} {\bibinfo {author} {\bibfnamefont {G.}~\bibnamefont
  {Rotskoff}}\ and\ \bibinfo {author} {\bibfnamefont {E.}~\bibnamefont
  {Vanden-Eijnden}},\ }in\ \href
  {https://proceedings.neurips.cc/paper_files/paper/2018/file/196f5641aa9dc87067da4ff90fd81e7b-Paper.pdf}
  {\emph {\bibinfo {booktitle} {Advances in Neural Information Processing
  Systems}}},\ Vol.~\bibinfo {volume} {31},\ \bibinfo {editor} {edited by\
  \bibinfo {editor} {\bibfnamefont {S.}~\bibnamefont {Bengio}}, \bibinfo
  {editor} {\bibfnamefont {H.}~\bibnamefont {Wallach}}, \bibinfo {editor}
  {\bibfnamefont {H.}~\bibnamefont {Larochelle}}, \bibinfo {editor}
  {\bibfnamefont {K.}~\bibnamefont {Grauman}}, \bibinfo {editor} {\bibfnamefont
  {N.}~\bibnamefont {Cesa-Bianchi}},\ and\ \bibinfo {editor} {\bibfnamefont
  {R.}~\bibnamefont {Garnett}}}\ (\bibinfo  {publisher} {Curran Associates,
  Inc.},\ \bibinfo {year} {2018})\BibitemShut {NoStop}%
\bibitem [{\citenamefont {Sirignano}\ and\ \citenamefont
  {Spiliopoulos}(2020)}]{SIRIGNANO20201820}%
  \BibitemOpen
  \bibfield  {author} {\bibinfo {author} {\bibfnamefont {J.}~\bibnamefont
  {Sirignano}}\ and\ \bibinfo {author} {\bibfnamefont {K.}~\bibnamefont
  {Spiliopoulos}},\ }\href
  {https://doi.org/https://doi.org/10.1016/j.spa.2019.06.003} {\bibfield
  {journal} {\bibinfo  {journal} {Stochastic Processes and their Applications}\
  }\textbf {\bibinfo {volume} {130}},\ \bibinfo {pages} {1820} (\bibinfo {year}
  {2020})}\BibitemShut {NoStop}%
\bibitem [{\citenamefont {Yang}\ and\ \citenamefont
  {Hu}(2021)}]{pmlr-v139-yang21c}%
  \BibitemOpen
  \bibfield  {author} {\bibinfo {author} {\bibfnamefont {G.}~\bibnamefont
  {Yang}}\ and\ \bibinfo {author} {\bibfnamefont {E.~J.}\ \bibnamefont {Hu}},\
  }in\ \href {https://proceedings.mlr.press/v139/yang21c.html} {\emph {\bibinfo
  {booktitle} {Proceedings of the 38th International Conference on Machine
  Learning}}},\ \bibinfo {series} {Proceedings of Machine Learning Research},
  Vol.\ \bibinfo {volume} {139},\ \bibinfo {editor} {edited by\ \bibinfo
  {editor} {\bibfnamefont {M.}~\bibnamefont {Meila}}\ and\ \bibinfo {editor}
  {\bibfnamefont {T.}~\bibnamefont {Zhang}}}\ (\bibinfo  {publisher} {PMLR},\
  \bibinfo {year} {2021})\ pp.\ \bibinfo {pages} {11727--11737}\BibitemShut
  {NoStop}%
\bibitem [{\citenamefont {Bordelon}\ and\ \citenamefont
  {Pehlevan}(2022)}]{bordelon2022selfconsistent}%
  \BibitemOpen
  \bibfield  {author} {\bibinfo {author} {\bibfnamefont {B.}~\bibnamefont
  {Bordelon}}\ and\ \bibinfo {author} {\bibfnamefont {C.}~\bibnamefont
  {Pehlevan}},\ }in\ \href@noop {} {\emph {\bibinfo {booktitle} {Advances in
  Neural Information Processing Systems}}}\ (\bibinfo {year}
  {2022})\BibitemShut {NoStop}%
\bibitem [{\citenamefont {Bordelon}\ and\ \citenamefont
  {Pehlevan}(2023)}]{bordelon2023dynamics}%
  \BibitemOpen
  \bibfield  {author} {\bibinfo {author} {\bibfnamefont {B.}~\bibnamefont
  {Bordelon}}\ and\ \bibinfo {author} {\bibfnamefont {C.}~\bibnamefont
  {Pehlevan}},\ }in\ \href {https://openreview.net/forum?id=fKwG6grp8o} {\emph
  {\bibinfo {booktitle} {Thirty-seventh Conference on Neural Information
  Processing Systems}}}\ (\bibinfo {year} {2023})\BibitemShut {NoStop}%
\bibitem [{\citenamefont {Cortes}\ \emph {et~al.}(2012)\citenamefont {Cortes},
  \citenamefont {Mohri},\ and\ \citenamefont
  {Rostamizadeh}}]{JMLR:v13:cortes12a}%
  \BibitemOpen
  \bibfield  {author} {\bibinfo {author} {\bibfnamefont {C.}~\bibnamefont
  {Cortes}}, \bibinfo {author} {\bibfnamefont {M.}~\bibnamefont {Mohri}},\ and\
  \bibinfo {author} {\bibfnamefont {A.}~\bibnamefont {Rostamizadeh}},\ }\href
  {http://jmlr.org/papers/v13/cortes12a.html} {\bibfield  {journal} {\bibinfo
  {journal} {Journal of Machine Learning Research}\ }\textbf {\bibinfo {volume}
  {13}},\ \bibinfo {pages} {795} (\bibinfo {year} {2012})}\BibitemShut
  {NoStop}%
\bibitem [{\citenamefont {Kornblith}\ \emph {et~al.}(2019)\citenamefont
  {Kornblith}, \citenamefont {Norouzi}, \citenamefont {Lee},\ and\
  \citenamefont {Hinton}}]{pmlr-v97-kornblith19a}%
  \BibitemOpen
  \bibfield  {author} {\bibinfo {author} {\bibfnamefont {S.}~\bibnamefont
  {Kornblith}}, \bibinfo {author} {\bibfnamefont {M.}~\bibnamefont {Norouzi}},
  \bibinfo {author} {\bibfnamefont {H.}~\bibnamefont {Lee}},\ and\ \bibinfo
  {author} {\bibfnamefont {G.}~\bibnamefont {Hinton}},\ }in\ \href
  {https://proceedings.mlr.press/v97/kornblith19a.html} {\emph {\bibinfo
  {booktitle} {Proceedings of the 36th International Conference on Machine
  Learning}}},\ \bibinfo {series} {Proceedings of Machine Learning Research},
  Vol.~\bibinfo {volume} {97},\ \bibinfo {editor} {edited by\ \bibinfo {editor}
  {\bibfnamefont {K.}~\bibnamefont {Chaudhuri}}\ and\ \bibinfo {editor}
  {\bibfnamefont {R.}~\bibnamefont {Salakhutdinov}}}\ (\bibinfo  {publisher}
  {PMLR},\ \bibinfo {year} {2019})\ pp.\ \bibinfo {pages}
  {3519--3529}\BibitemShut {NoStop}%
\bibitem [{\citenamefont {Krauth}\ and\ \citenamefont
  {Mézard}(1989)}]{Krauth1989}%
  \BibitemOpen
  \bibfield  {author} {\bibinfo {author} {\bibfnamefont {W.}~\bibnamefont
  {Krauth}}\ and\ \bibinfo {author} {\bibfnamefont {M.}~\bibnamefont
  {Mézard}},\ }\href {https://doi.org/10.1051/jphys:0198900500200305700}
  {\bibfield  {journal} {\bibinfo  {journal} {Journal de Physique}\ }\textbf
  {\bibinfo {volume} {50}},\ \bibinfo {pages} {3057–3066} (\bibinfo {year}
  {1989})}\BibitemShut {NoStop}%
\bibitem [{\citenamefont {Seung}\ \emph {et~al.}(1992)\citenamefont {Seung},
  \citenamefont {Sompolinsky},\ and\ \citenamefont
  {Tishby}}]{PhysRevA.45.6056}%
  \BibitemOpen
  \bibfield  {author} {\bibinfo {author} {\bibfnamefont {H.~S.}\ \bibnamefont
  {Seung}}, \bibinfo {author} {\bibfnamefont {H.}~\bibnamefont {Sompolinsky}},\
  and\ \bibinfo {author} {\bibfnamefont {N.}~\bibnamefont {Tishby}},\ }\href
  {https://doi.org/10.1103/PhysRevA.45.6056} {\bibfield  {journal} {\bibinfo
  {journal} {Phys. Rev. A}\ }\textbf {\bibinfo {volume} {45}},\ \bibinfo
  {pages} {6056} (\bibinfo {year} {1992})}\BibitemShut {NoStop}%
\bibitem [{\citenamefont {Watkin}\ \emph {et~al.}(1993)\citenamefont {Watkin},
  \citenamefont {Rau},\ and\ \citenamefont {Biehl}}]{RevModPhys.65.499}%
  \BibitemOpen
  \bibfield  {author} {\bibinfo {author} {\bibfnamefont {T.~L.~H.}\
  \bibnamefont {Watkin}}, \bibinfo {author} {\bibfnamefont {A.}~\bibnamefont
  {Rau}},\ and\ \bibinfo {author} {\bibfnamefont {M.}~\bibnamefont {Biehl}},\
  }\href {https://doi.org/10.1103/RevModPhys.65.499} {\bibfield  {journal}
  {\bibinfo  {journal} {Rev. Mod. Phys.}\ }\textbf {\bibinfo {volume} {65}},\
  \bibinfo {pages} {499} (\bibinfo {year} {1993})}\BibitemShut {NoStop}%
\bibitem [{\citenamefont {Barkai}\ \emph {et~al.}(1992)\citenamefont {Barkai},
  \citenamefont {Hansel},\ and\ \citenamefont
  {Sompolinsky}}]{PhysRevA.45.4146}%
  \BibitemOpen
  \bibfield  {author} {\bibinfo {author} {\bibfnamefont {E.}~\bibnamefont
  {Barkai}}, \bibinfo {author} {\bibfnamefont {D.}~\bibnamefont {Hansel}},\
  and\ \bibinfo {author} {\bibfnamefont {H.}~\bibnamefont {Sompolinsky}},\
  }\href {https://doi.org/10.1103/PhysRevA.45.4146} {\bibfield  {journal}
  {\bibinfo  {journal} {Phys. Rev. A}\ }\textbf {\bibinfo {volume} {45}},\
  \bibinfo {pages} {4146} (\bibinfo {year} {1992})}\BibitemShut {NoStop}%
\bibitem [{\citenamefont {Engel}\ \emph {et~al.}(1992)\citenamefont {Engel},
  \citenamefont {K\"ohler}, \citenamefont {Tschepke}, \citenamefont
  {Vollmayr},\ and\ \citenamefont {Zippelius}}]{PhysRevA.45.7590}%
  \BibitemOpen
  \bibfield  {author} {\bibinfo {author} {\bibfnamefont {A.}~\bibnamefont
  {Engel}}, \bibinfo {author} {\bibfnamefont {H.~M.}\ \bibnamefont {K\"ohler}},
  \bibinfo {author} {\bibfnamefont {F.}~\bibnamefont {Tschepke}}, \bibinfo
  {author} {\bibfnamefont {H.}~\bibnamefont {Vollmayr}},\ and\ \bibinfo
  {author} {\bibfnamefont {A.}~\bibnamefont {Zippelius}},\ }\href
  {https://doi.org/10.1103/PhysRevA.45.7590} {\bibfield  {journal} {\bibinfo
  {journal} {Phys. Rev. A}\ }\textbf {\bibinfo {volume} {45}},\ \bibinfo
  {pages} {7590} (\bibinfo {year} {1992})}\BibitemShut {NoStop}%
\bibitem [{\citenamefont {Hou}\ \emph {et~al.}(2019)\citenamefont {Hou},
  \citenamefont {Wong},\ and\ \citenamefont {Huang}}]{Hou_2019}%
  \BibitemOpen
  \bibfield  {author} {\bibinfo {author} {\bibfnamefont {T.}~\bibnamefont
  {Hou}}, \bibinfo {author} {\bibfnamefont {K.~Y.~M.}\ \bibnamefont {Wong}},\
  and\ \bibinfo {author} {\bibfnamefont {H.}~\bibnamefont {Huang}},\ }\href
  {https://doi.org/10.1088/1751-8121/ab3f3f} {\bibfield  {journal} {\bibinfo
  {journal} {Journal of Physics A: Mathematical and Theoretical}\ }\textbf
  {\bibinfo {volume} {52}},\ \bibinfo {pages} {414001} (\bibinfo {year}
  {2019})}\BibitemShut {NoStop}%
\bibitem [{\citenamefont {Kunin}\ \emph {et~al.}(2021)\citenamefont {Kunin},
  \citenamefont {Sagastuy-Brena}, \citenamefont {Ganguli}, \citenamefont
  {Yamins},\ and\ \citenamefont {Tanaka}}]{kunin2021neural}%
  \BibitemOpen
  \bibfield  {author} {\bibinfo {author} {\bibfnamefont {D.}~\bibnamefont
  {Kunin}}, \bibinfo {author} {\bibfnamefont {J.}~\bibnamefont
  {Sagastuy-Brena}}, \bibinfo {author} {\bibfnamefont {S.}~\bibnamefont
  {Ganguli}}, \bibinfo {author} {\bibfnamefont {D.~L.}\ \bibnamefont
  {Yamins}},\ and\ \bibinfo {author} {\bibfnamefont {H.}~\bibnamefont
  {Tanaka}},\ }in\ \href {https://openreview.net/forum?id=q8qLAbQBupm} {\emph
  {\bibinfo {booktitle} {International Conference on Learning
  Representations}}}\ (\bibinfo {year} {2021})\BibitemShut {NoStop}%
\bibitem [{\citenamefont {Tanaka}\ and\ \citenamefont
  {Kunin}(2021)}]{NEURIPS2021_d76d8dee}%
  \BibitemOpen
  \bibfield  {author} {\bibinfo {author} {\bibfnamefont {H.}~\bibnamefont
  {Tanaka}}\ and\ \bibinfo {author} {\bibfnamefont {D.}~\bibnamefont {Kunin}},\
  }in\ \href
  {https://proceedings.neurips.cc/paper_files/paper/2021/file/d76d8deea9c19cc9aaf2237d2bf2f785-Paper.pdf}
  {\emph {\bibinfo {booktitle} {Advances in Neural Information Processing
  Systems}}},\ Vol.~\bibinfo {volume} {34},\ \bibinfo {editor} {edited by\
  \bibinfo {editor} {\bibfnamefont {M.}~\bibnamefont {Ranzato}}, \bibinfo
  {editor} {\bibfnamefont {A.}~\bibnamefont {Beygelzimer}}, \bibinfo {editor}
  {\bibfnamefont {Y.}~\bibnamefont {Dauphin}}, \bibinfo {editor} {\bibfnamefont
  {P.}~\bibnamefont {Liang}},\ and\ \bibinfo {editor} {\bibfnamefont {J.~W.}\
  \bibnamefont {Vaughan}}}\ (\bibinfo  {publisher} {Curran Associates, Inc.},\
  \bibinfo {year} {2021})\ pp.\ \bibinfo {pages} {25646--25660}\BibitemShut
  {NoStop}%
\bibitem [{\citenamefont {Rubin}\ \emph {et~al.}(2024)\citenamefont {Rubin},
  \citenamefont {Seroussi},\ and\ \citenamefont {Ringel}}]{rubin2024grokking}%
  \BibitemOpen
  \bibfield  {author} {\bibinfo {author} {\bibfnamefont {N.}~\bibnamefont
  {Rubin}}, \bibinfo {author} {\bibfnamefont {I.}~\bibnamefont {Seroussi}},\
  and\ \bibinfo {author} {\bibfnamefont {Z.}~\bibnamefont {Ringel}},\ }in\
  \href {https://openreview.net/forum?id=3ROGsTX3IR} {\emph {\bibinfo
  {booktitle} {The Twelfth International Conference on Learning
  Representations}}}\ (\bibinfo {year} {2024})\BibitemShut {NoStop}%
\bibitem [{\citenamefont {Brea}\ \emph {et~al.}(2019)\citenamefont {Brea},
  \citenamefont {Simsek}, \citenamefont {Illing},\ and\ \citenamefont
  {Gerstner}}]{brea2019weightspace}%
  \BibitemOpen
  \bibfield  {author} {\bibinfo {author} {\bibfnamefont {J.}~\bibnamefont
  {Brea}}, \bibinfo {author} {\bibfnamefont {B.}~\bibnamefont {Simsek}},
  \bibinfo {author} {\bibfnamefont {B.}~\bibnamefont {Illing}},\ and\ \bibinfo
  {author} {\bibfnamefont {W.}~\bibnamefont {Gerstner}},\ }\href@noop {}
  {\bibinfo {title} {Weight-space symmetry in deep networks gives rise to
  permutation saddles, connected by equal-loss valleys across the loss
  landscape}} (\bibinfo {year} {2019}),\ \Eprint
  {https://arxiv.org/abs/1907.02911} {arXiv:1907.02911 [cs.LG]} \BibitemShut
  {NoStop}%
\bibitem [{\citenamefont {Simsek}\ \emph {et~al.}(2021)\citenamefont {Simsek},
  \citenamefont {Ged}, \citenamefont {Jacot}, \citenamefont {Spadaro},
  \citenamefont {Hongler}, \citenamefont {Gerstner},\ and\ \citenamefont
  {Brea}}]{pmlr-v139-simsek21a}%
  \BibitemOpen
  \bibfield  {author} {\bibinfo {author} {\bibfnamefont {B.}~\bibnamefont
  {Simsek}}, \bibinfo {author} {\bibfnamefont {F.}~\bibnamefont {Ged}},
  \bibinfo {author} {\bibfnamefont {A.}~\bibnamefont {Jacot}}, \bibinfo
  {author} {\bibfnamefont {F.}~\bibnamefont {Spadaro}}, \bibinfo {author}
  {\bibfnamefont {C.}~\bibnamefont {Hongler}}, \bibinfo {author} {\bibfnamefont
  {W.}~\bibnamefont {Gerstner}},\ and\ \bibinfo {author} {\bibfnamefont
  {J.}~\bibnamefont {Brea}},\ }in\ \href
  {https://proceedings.mlr.press/v139/simsek21a.html} {\emph {\bibinfo
  {booktitle} {Proceedings of the 38th International Conference on Machine
  Learning}}},\ \bibinfo {series} {Proceedings of Machine Learning Research},
  Vol.\ \bibinfo {volume} {139},\ \bibinfo {editor} {edited by\ \bibinfo
  {editor} {\bibfnamefont {M.}~\bibnamefont {Meila}}\ and\ \bibinfo {editor}
  {\bibfnamefont {T.}~\bibnamefont {Zhang}}}\ (\bibinfo  {publisher} {PMLR},\
  \bibinfo {year} {2021})\ pp.\ \bibinfo {pages} {9722--9732}\BibitemShut
  {NoStop}%
\bibitem [{\citenamefont {Entezari}\ \emph {et~al.}(2022)\citenamefont
  {Entezari}, \citenamefont {Sedghi}, \citenamefont {Saukh},\ and\
  \citenamefont {Neyshabur}}]{entezari2022the}%
  \BibitemOpen
  \bibfield  {author} {\bibinfo {author} {\bibfnamefont {R.}~\bibnamefont
  {Entezari}}, \bibinfo {author} {\bibfnamefont {H.}~\bibnamefont {Sedghi}},
  \bibinfo {author} {\bibfnamefont {O.}~\bibnamefont {Saukh}},\ and\ \bibinfo
  {author} {\bibfnamefont {B.}~\bibnamefont {Neyshabur}},\ }in\ \href
  {https://openreview.net/forum?id=dNigytemkL} {\emph {\bibinfo {booktitle}
  {International Conference on Learning Representations}}}\ (\bibinfo {year}
  {2022})\BibitemShut {NoStop}%
\bibitem [{\citenamefont {Papyan}\ \emph {et~al.}(2020)\citenamefont {Papyan},
  \citenamefont {Han},\ and\ \citenamefont
  {Donoho}}]{doi:10.1073/pnas.2015509117}%
  \BibitemOpen
  \bibfield  {author} {\bibinfo {author} {\bibfnamefont {V.}~\bibnamefont
  {Papyan}}, \bibinfo {author} {\bibfnamefont {X.~Y.}\ \bibnamefont {Han}},\
  and\ \bibinfo {author} {\bibfnamefont {D.~L.}\ \bibnamefont {Donoho}},\
  }\href {https://doi.org/10.1073/pnas.2015509117} {\bibfield  {journal}
  {\bibinfo  {journal} {Proceedings of the National Academy of Sciences}\
  }\textbf {\bibinfo {volume} {117}},\ \bibinfo {pages} {24652} (\bibinfo
  {year} {2020})}\BibitemShut {NoStop}%
\bibitem [{\citenamefont {Han}\ \emph {et~al.}(2022)\citenamefont {Han},
  \citenamefont {Papyan},\ and\ \citenamefont {Donoho}}]{han2022neural}%
  \BibitemOpen
  \bibfield  {author} {\bibinfo {author} {\bibfnamefont {X.}~\bibnamefont
  {Han}}, \bibinfo {author} {\bibfnamefont {V.}~\bibnamefont {Papyan}},\ and\
  \bibinfo {author} {\bibfnamefont {D.~L.}\ \bibnamefont {Donoho}},\ }in\ \href
  {https://openreview.net/forum?id=w1UbdvWH_R3} {\emph {\bibinfo {booktitle}
  {International Conference on Learning Representations}}}\ (\bibinfo {year}
  {2022})\BibitemShut {NoStop}%
\bibitem [{\citenamefont {Fei-Fei}\ \emph {et~al.}(2006)\citenamefont
  {Fei-Fei}, \citenamefont {Fergus},\ and\ \citenamefont {Perona}}]{1597116}%
  \BibitemOpen
  \bibfield  {author} {\bibinfo {author} {\bibfnamefont {L.}~\bibnamefont
  {Fei-Fei}}, \bibinfo {author} {\bibfnamefont {R.}~\bibnamefont {Fergus}},\
  and\ \bibinfo {author} {\bibfnamefont {P.}~\bibnamefont {Perona}},\ }\href
  {https://doi.org/10.1109/TPAMI.2006.79} {\bibfield  {journal} {\bibinfo
  {journal} {IEEE Transactions on Pattern Analysis and Machine Intelligence}\
  }\textbf {\bibinfo {volume} {28}},\ \bibinfo {pages} {594} (\bibinfo {year}
  {2006})}\BibitemShut {NoStop}%
\bibitem [{\citenamefont {Vinyals}\ \emph {et~al.}(2016)\citenamefont
  {Vinyals}, \citenamefont {Blundell}, \citenamefont {Lillicrap}, \citenamefont
  {kavukcuoglu},\ and\ \citenamefont {Wierstra}}]{NIPS2016_90e13578}%
  \BibitemOpen
  \bibfield  {author} {\bibinfo {author} {\bibfnamefont {O.}~\bibnamefont
  {Vinyals}}, \bibinfo {author} {\bibfnamefont {C.}~\bibnamefont {Blundell}},
  \bibinfo {author} {\bibfnamefont {T.}~\bibnamefont {Lillicrap}}, \bibinfo
  {author} {\bibfnamefont {k.}~\bibnamefont {kavukcuoglu}},\ and\ \bibinfo
  {author} {\bibfnamefont {D.}~\bibnamefont {Wierstra}},\ }in\ \href
  {https://proceedings.neurips.cc/paper_files/paper/2016/file/90e1357833654983612fb05e3ec9148c-Paper.pdf}
  {\emph {\bibinfo {booktitle} {Advances in Neural Information Processing
  Systems}}},\ Vol.~\bibinfo {volume} {29},\ \bibinfo {editor} {edited by\
  \bibinfo {editor} {\bibfnamefont {D.}~\bibnamefont {Lee}}, \bibinfo {editor}
  {\bibfnamefont {M.}~\bibnamefont {Sugiyama}}, \bibinfo {editor}
  {\bibfnamefont {U.}~\bibnamefont {Luxburg}}, \bibinfo {editor} {\bibfnamefont
  {I.}~\bibnamefont {Guyon}},\ and\ \bibinfo {editor} {\bibfnamefont
  {R.}~\bibnamefont {Garnett}}}\ (\bibinfo  {publisher} {Curran Associates,
  Inc.},\ \bibinfo {year} {2016})\BibitemShut {NoStop}%
\bibitem [{\citenamefont {Snell}\ \emph {et~al.}(2017)\citenamefont {Snell},
  \citenamefont {Swersky},\ and\ \citenamefont {Zemel}}]{NIPS2017_cb8da676}%
  \BibitemOpen
  \bibfield  {author} {\bibinfo {author} {\bibfnamefont {J.}~\bibnamefont
  {Snell}}, \bibinfo {author} {\bibfnamefont {K.}~\bibnamefont {Swersky}},\
  and\ \bibinfo {author} {\bibfnamefont {R.}~\bibnamefont {Zemel}},\ }in\ \href
  {https://proceedings.neurips.cc/paper_files/paper/2017/file/cb8da6767461f2812ae4290eac7cbc42-Paper.pdf}
  {\emph {\bibinfo {booktitle} {Advances in Neural Information Processing
  Systems}}},\ Vol.~\bibinfo {volume} {30},\ \bibinfo {editor} {edited by\
  \bibinfo {editor} {\bibfnamefont {I.}~\bibnamefont {Guyon}}, \bibinfo
  {editor} {\bibfnamefont {U.~V.}\ \bibnamefont {Luxburg}}, \bibinfo {editor}
  {\bibfnamefont {S.}~\bibnamefont {Bengio}}, \bibinfo {editor} {\bibfnamefont
  {H.}~\bibnamefont {Wallach}}, \bibinfo {editor} {\bibfnamefont
  {R.}~\bibnamefont {Fergus}}, \bibinfo {editor} {\bibfnamefont
  {S.}~\bibnamefont {Vishwanathan}},\ and\ \bibinfo {editor} {\bibfnamefont
  {R.}~\bibnamefont {Garnett}}}\ (\bibinfo  {publisher} {Curran Associates,
  Inc.},\ \bibinfo {year} {2017})\BibitemShut {NoStop}%
\bibitem [{\citenamefont {Sorscher}\ \emph {et~al.}(2022)\citenamefont
  {Sorscher}, \citenamefont {Ganguli},\ and\ \citenamefont
  {Sompolinsky}}]{doi:10.1073/pnas.2200800119}%
  \BibitemOpen
  \bibfield  {author} {\bibinfo {author} {\bibfnamefont {B.}~\bibnamefont
  {Sorscher}}, \bibinfo {author} {\bibfnamefont {S.}~\bibnamefont {Ganguli}},\
  and\ \bibinfo {author} {\bibfnamefont {H.}~\bibnamefont {Sompolinsky}},\
  }\href {https://doi.org/10.1073/pnas.2200800119} {\bibfield  {journal}
  {\bibinfo  {journal} {Proceedings of the National Academy of Sciences}\
  }\textbf {\bibinfo {volume} {119}},\ \bibinfo {pages} {e2200800119} (\bibinfo
  {year} {2022})}\BibitemShut {NoStop}%
\bibitem [{\citenamefont {Galanti}\ \emph {et~al.}(2022)\citenamefont
  {Galanti}, \citenamefont {Gy{\"o}rgy},\ and\ \citenamefont
  {Hutter}}]{galanti2022on}%
  \BibitemOpen
  \bibfield  {author} {\bibinfo {author} {\bibfnamefont {T.}~\bibnamefont
  {Galanti}}, \bibinfo {author} {\bibfnamefont {A.}~\bibnamefont
  {Gy{\"o}rgy}},\ and\ \bibinfo {author} {\bibfnamefont {M.}~\bibnamefont
  {Hutter}},\ }in\ \href {https://openreview.net/forum?id=SwIp410B6aQ} {\emph
  {\bibinfo {booktitle} {International Conference on Learning
  Representations}}}\ (\bibinfo {year} {2022})\BibitemShut {NoStop}%
\bibitem [{\citenamefont {Jacot}\ \emph {et~al.}(2018)\citenamefont {Jacot},
  \citenamefont {Gabriel},\ and\ \citenamefont
  {Hongler}}]{NEURIPS2018_5a4be1fa}%
  \BibitemOpen
  \bibfield  {author} {\bibinfo {author} {\bibfnamefont {A.}~\bibnamefont
  {Jacot}}, \bibinfo {author} {\bibfnamefont {F.}~\bibnamefont {Gabriel}},\
  and\ \bibinfo {author} {\bibfnamefont {C.}~\bibnamefont {Hongler}},\ }in\
  \href
  {https://proceedings.neurips.cc/paper_files/paper/2018/file/5a4be1fa34e62bb8a6ec6b91d2462f5a-Paper.pdf}
  {\emph {\bibinfo {booktitle} {Advances in Neural Information Processing
  Systems}}},\ Vol.~\bibinfo {volume} {31},\ \bibinfo {editor} {edited by\
  \bibinfo {editor} {\bibfnamefont {S.}~\bibnamefont {Bengio}}, \bibinfo
  {editor} {\bibfnamefont {H.}~\bibnamefont {Wallach}}, \bibinfo {editor}
  {\bibfnamefont {H.}~\bibnamefont {Larochelle}}, \bibinfo {editor}
  {\bibfnamefont {K.}~\bibnamefont {Grauman}}, \bibinfo {editor} {\bibfnamefont
  {N.}~\bibnamefont {Cesa-Bianchi}},\ and\ \bibinfo {editor} {\bibfnamefont
  {R.}~\bibnamefont {Garnett}}}\ (\bibinfo  {publisher} {Curran Associates,
  Inc.},\ \bibinfo {year} {2018})\BibitemShut {NoStop}%
\bibitem [{\citenamefont {Gupta}\ and\ \citenamefont
  {Nagar}(1999)}]{Gupta1999}%
  \BibitemOpen
  \bibfield  {author} {\bibinfo {author} {\bibfnamefont {A.~K.}\ \bibnamefont
  {Gupta}}\ and\ \bibinfo {author} {\bibfnamefont {D.~K.}\ \bibnamefont
  {Nagar}},\ }\href@noop {} {\emph {\bibinfo {title} {Matrix Variate
  Distributions}}},\ Monographs and Surveys in Pure and Applied Mathematics\
  (\bibinfo  {publisher} {Chapman \& Hall/CRC},\ \bibinfo {address}
  {Philadelphia, PA},\ \bibinfo {year} {1999})\BibitemShut {NoStop}%
\bibitem [{\citenamefont {Fedoryuk}(1977)}]{fedoryuk1977saddle}%
  \BibitemOpen
  \bibfield  {author} {\bibinfo {author} {\bibfnamefont {M.}~\bibnamefont
  {Fedoryuk}},\ }\href@noop {} {\bibinfo {title} {The saddle-point method}}
  (\bibinfo {year} {1977})\BibitemShut {NoStop}%
\bibitem [{\citenamefont {Fedoryuk}(1989)}]{Fedoryuk1989}%
  \BibitemOpen
  \bibfield  {author} {\bibinfo {author} {\bibfnamefont {M.~V.}\ \bibnamefont
  {Fedoryuk}},\ }\bibinfo {title} {Asymptotic methods in analysis},\ in\ \href
  {https://doi.org/10.1007/978-3-642-61310-4_2} {\emph {\bibinfo {booktitle}
  {Encyclopaedia of Mathematical Sciences}}}\ (\bibinfo  {publisher} {Springer
  Berlin Heidelberg},\ \bibinfo {year} {1989})\ p.\ \bibinfo {pages}
  {83–191}\BibitemShut {NoStop}%
\bibitem [{\citenamefont {Shun}\ and\ \citenamefont
  {McCullagh}(1995)}]{33dbad4f-5b73-3bc9-ac01-477e2b81fa2a}%
  \BibitemOpen
  \bibfield  {author} {\bibinfo {author} {\bibfnamefont {Z.}~\bibnamefont
  {Shun}}\ and\ \bibinfo {author} {\bibfnamefont {P.}~\bibnamefont
  {McCullagh}},\ }\href@noop {} {\bibfield  {journal} {\bibinfo  {journal}
  {Journal of the Royal Statistical Society. Series B (Methodological)}\
  }\textbf {\bibinfo {volume} {57}},\ \bibinfo {pages} {749} (\bibinfo {year}
  {1995})}\BibitemShut {NoStop}%
\bibitem [{\citenamefont {Spokoiny}(2023)}]{doi:10.1137/22M1495688}%
  \BibitemOpen
  \bibfield  {author} {\bibinfo {author} {\bibfnamefont {V.}~\bibnamefont
  {Spokoiny}},\ }\href {https://doi.org/10.1137/22M1495688} {\bibfield
  {journal} {\bibinfo  {journal} {SIAM/ASA Journal on Uncertainty
  Quantification}\ }\textbf {\bibinfo {volume} {11}},\ \bibinfo {pages} {1044}
  (\bibinfo {year} {2023})}\BibitemShut {NoStop}%
\bibitem [{\citenamefont {Avidan}\ \emph {et~al.}(2023)\citenamefont {Avidan},
  \citenamefont {Li},\ and\ \citenamefont
  {Sompolinsky}}]{avidan2023connecting}%
  \BibitemOpen
  \bibfield  {author} {\bibinfo {author} {\bibfnamefont {Y.}~\bibnamefont
  {Avidan}}, \bibinfo {author} {\bibfnamefont {Q.}~\bibnamefont {Li}},\ and\
  \bibinfo {author} {\bibfnamefont {H.}~\bibnamefont {Sompolinsky}},\
  }\href@noop {} {\bibinfo {title} {Connecting ntk and nngp: A unified
  theoretical framework for neural network learning dynamics in the kernel
  regime}} (\bibinfo {year} {2023}),\ \Eprint
  {https://arxiv.org/abs/2309.04522} {arXiv:2309.04522} \BibitemShut {NoStop}%
\bibitem [{\citenamefont {Owen}(1980)}]{doi:10.1080/03610918008812164}%
  \BibitemOpen
  \bibfield  {author} {\bibinfo {author} {\bibfnamefont {D.~B.}\ \bibnamefont
  {Owen}},\ }\href {https://doi.org/10.1080/03610918008812164} {\bibfield
  {journal} {\bibinfo  {journal} {Communications in Statistics - Simulation and
  Computation}\ }\textbf {\bibinfo {volume} {9}},\ \bibinfo {pages} {389}
  (\bibinfo {year} {1980})}\BibitemShut {NoStop}%
\bibitem [{\citenamefont {van Meegen}\ and\ \citenamefont {van
  Albada}(2021)}]{PhysRevResearch.3.043077}%
  \BibitemOpen
  \bibfield  {author} {\bibinfo {author} {\bibfnamefont {A.}~\bibnamefont {van
  Meegen}}\ and\ \bibinfo {author} {\bibfnamefont {S.~J.}\ \bibnamefont {van
  Albada}},\ }\href {https://doi.org/10.1103/PhysRevResearch.3.043077}
  {\bibfield  {journal} {\bibinfo  {journal} {Phys. Rev. Res.}\ }\textbf
  {\bibinfo {volume} {3}},\ \bibinfo {pages} {043077} (\bibinfo {year}
  {2021})}\BibitemShut {NoStop}%
\bibitem [{\citenamefont {Harris}\ \emph {et~al.}(2020)\citenamefont {Harris},
  \citenamefont {Millman}, \citenamefont {van~der Walt}, \citenamefont
  {Gommers}, \citenamefont {Virtanen}, \citenamefont {Cournapeau},
  \citenamefont {Wieser}, \citenamefont {Taylor}, \citenamefont {Berg},
  \citenamefont {Smith}, \citenamefont {Kern}, \citenamefont {Picus},
  \citenamefont {Hoyer}, \citenamefont {van Kerkwijk}, \citenamefont {Brett},
  \citenamefont {Haldane}, \citenamefont {Fernández~del Río}, \citenamefont
  {Wiebe}, \citenamefont {Peterson}, \citenamefont {Gérard-Marchant},
  \citenamefont {Sheppard}, \citenamefont {Reddy}, \citenamefont {Weckesser},
  \citenamefont {Abbasi}, \citenamefont {Gohlke},\ and\ \citenamefont
  {Oliphant}}]{Harris20_357}%
  \BibitemOpen
  \bibfield  {author} {\bibinfo {author} {\bibfnamefont {C.~R.}\ \bibnamefont
  {Harris}}, \bibinfo {author} {\bibfnamefont {K.~J.}\ \bibnamefont {Millman}},
  \bibinfo {author} {\bibfnamefont {S.~J.}\ \bibnamefont {van~der Walt}},
  \bibinfo {author} {\bibfnamefont {R.}~\bibnamefont {Gommers}}, \bibinfo
  {author} {\bibfnamefont {P.}~\bibnamefont {Virtanen}}, \bibinfo {author}
  {\bibfnamefont {D.}~\bibnamefont {Cournapeau}}, \bibinfo {author}
  {\bibfnamefont {E.}~\bibnamefont {Wieser}}, \bibinfo {author} {\bibfnamefont
  {J.}~\bibnamefont {Taylor}}, \bibinfo {author} {\bibfnamefont
  {S.}~\bibnamefont {Berg}}, \bibinfo {author} {\bibfnamefont {N.~J.}\
  \bibnamefont {Smith}}, \bibinfo {author} {\bibfnamefont {R.}~\bibnamefont
  {Kern}}, \bibinfo {author} {\bibfnamefont {M.}~\bibnamefont {Picus}},
  \bibinfo {author} {\bibfnamefont {S.}~\bibnamefont {Hoyer}}, \bibinfo
  {author} {\bibfnamefont {M.~H.}\ \bibnamefont {van Kerkwijk}}, \bibinfo
  {author} {\bibfnamefont {M.}~\bibnamefont {Brett}}, \bibinfo {author}
  {\bibfnamefont {A.}~\bibnamefont {Haldane}}, \bibinfo {author} {\bibfnamefont
  {J.}~\bibnamefont {Fernández~del Río}}, \bibinfo {author} {\bibfnamefont
  {M.}~\bibnamefont {Wiebe}}, \bibinfo {author} {\bibfnamefont
  {P.}~\bibnamefont {Peterson}}, \bibinfo {author} {\bibfnamefont
  {P.}~\bibnamefont {Gérard-Marchant}}, \bibinfo {author} {\bibfnamefont
  {K.}~\bibnamefont {Sheppard}}, \bibinfo {author} {\bibfnamefont
  {T.}~\bibnamefont {Reddy}}, \bibinfo {author} {\bibfnamefont
  {W.}~\bibnamefont {Weckesser}}, \bibinfo {author} {\bibfnamefont
  {H.}~\bibnamefont {Abbasi}}, \bibinfo {author} {\bibfnamefont
  {C.}~\bibnamefont {Gohlke}},\ and\ \bibinfo {author} {\bibfnamefont {T.~E.}\
  \bibnamefont {Oliphant}},\ }\href {https://doi.org/10.1038/s41586-020-2649-2}
  {\bibfield  {journal} {\bibinfo  {journal} {Nature}\ }\textbf {\bibinfo
  {volume} {585}},\ \bibinfo {pages} {357–362} (\bibinfo {year}
  {2020})}\BibitemShut {NoStop}%
\bibitem [{\citenamefont {Virtanen}\ \emph {et~al.}(2020)\citenamefont
  {Virtanen}, \citenamefont {Gommers}, \citenamefont {Oliphant}, \citenamefont
  {Haberland}, \citenamefont {Reddy}, \citenamefont {Cournapeau}, \citenamefont
  {Burovski}, \citenamefont {Peterson}, \citenamefont {Weckesser},
  \citenamefont {Bright}, \citenamefont {{van der Walt}}, \citenamefont
  {Brett}, \citenamefont {Wilson}, \citenamefont {Millman}, \citenamefont
  {Mayorov}, \citenamefont {Nelson}, \citenamefont {Jones}, \citenamefont
  {Kern}, \citenamefont {Larson}, \citenamefont {Carey}, \citenamefont {Polat},
  \citenamefont {Feng}, \citenamefont {Moore}, \citenamefont {{VanderPlas}},
  \citenamefont {Laxalde}, \citenamefont {Perktold}, \citenamefont {Cimrman},
  \citenamefont {Henriksen}, \citenamefont {Quintero}, \citenamefont {Harris},
  \citenamefont {Archibald}, \citenamefont {Ribeiro}, \citenamefont
  {Pedregosa}, \citenamefont {{van Mulbregt}},\ and\ \citenamefont {{SciPy 1.0
  Contributors}}}]{Virtanen20_261}%
  \BibitemOpen
  \bibfield  {author} {\bibinfo {author} {\bibfnamefont {P.}~\bibnamefont
  {Virtanen}}, \bibinfo {author} {\bibfnamefont {R.}~\bibnamefont {Gommers}},
  \bibinfo {author} {\bibfnamefont {T.~E.}\ \bibnamefont {Oliphant}}, \bibinfo
  {author} {\bibfnamefont {M.}~\bibnamefont {Haberland}}, \bibinfo {author}
  {\bibfnamefont {T.}~\bibnamefont {Reddy}}, \bibinfo {author} {\bibfnamefont
  {D.}~\bibnamefont {Cournapeau}}, \bibinfo {author} {\bibfnamefont
  {E.}~\bibnamefont {Burovski}}, \bibinfo {author} {\bibfnamefont
  {P.}~\bibnamefont {Peterson}}, \bibinfo {author} {\bibfnamefont
  {W.}~\bibnamefont {Weckesser}}, \bibinfo {author} {\bibfnamefont
  {J.}~\bibnamefont {Bright}}, \bibinfo {author} {\bibfnamefont {S.~J.}\
  \bibnamefont {{van der Walt}}}, \bibinfo {author} {\bibfnamefont
  {M.}~\bibnamefont {Brett}}, \bibinfo {author} {\bibfnamefont
  {J.}~\bibnamefont {Wilson}}, \bibinfo {author} {\bibfnamefont {K.~J.}\
  \bibnamefont {Millman}}, \bibinfo {author} {\bibfnamefont {N.}~\bibnamefont
  {Mayorov}}, \bibinfo {author} {\bibfnamefont {A.~R.~J.}\ \bibnamefont
  {Nelson}}, \bibinfo {author} {\bibfnamefont {E.}~\bibnamefont {Jones}},
  \bibinfo {author} {\bibfnamefont {R.}~\bibnamefont {Kern}}, \bibinfo {author}
  {\bibfnamefont {E.}~\bibnamefont {Larson}}, \bibinfo {author} {\bibfnamefont
  {C.~J.}\ \bibnamefont {Carey}}, \bibinfo {author} {\bibfnamefont
  {{\.I}.}~\bibnamefont {Polat}}, \bibinfo {author} {\bibfnamefont
  {Y.}~\bibnamefont {Feng}}, \bibinfo {author} {\bibfnamefont {E.~W.}\
  \bibnamefont {Moore}}, \bibinfo {author} {\bibfnamefont {J.}~\bibnamefont
  {{VanderPlas}}}, \bibinfo {author} {\bibfnamefont {D.}~\bibnamefont
  {Laxalde}}, \bibinfo {author} {\bibfnamefont {J.}~\bibnamefont {Perktold}},
  \bibinfo {author} {\bibfnamefont {R.}~\bibnamefont {Cimrman}}, \bibinfo
  {author} {\bibfnamefont {I.}~\bibnamefont {Henriksen}}, \bibinfo {author}
  {\bibfnamefont {E.~A.}\ \bibnamefont {Quintero}}, \bibinfo {author}
  {\bibfnamefont {C.~R.}\ \bibnamefont {Harris}}, \bibinfo {author}
  {\bibfnamefont {A.~M.}\ \bibnamefont {Archibald}}, \bibinfo {author}
  {\bibfnamefont {A.~H.}\ \bibnamefont {Ribeiro}}, \bibinfo {author}
  {\bibfnamefont {F.}~\bibnamefont {Pedregosa}}, \bibinfo {author}
  {\bibfnamefont {P.}~\bibnamefont {{van Mulbregt}}},\ and\ \bibinfo {author}
  {\bibnamefont {{SciPy 1.0 Contributors}}},\ }\href
  {https://doi.org/10.1038/s41592-019-0686-2} {\bibfield  {journal} {\bibinfo
  {journal} {Nature Methods}\ }\textbf {\bibinfo {volume} {17}},\ \bibinfo
  {pages} {261} (\bibinfo {year} {2020})}\BibitemShut {NoStop}%
\bibitem [{\citenamefont {Betancourt}(2017)}]{Betancourt17_arXiv}%
  \BibitemOpen
  \bibfield  {author} {\bibinfo {author} {\bibfnamefont {M.}~\bibnamefont
  {Betancourt}},\ }\href@noop {} {\bibinfo {title} {A conceptual introduction
  to hamiltonian monte carlo}} (\bibinfo {year} {2017})\BibitemShut {NoStop}%
\bibitem [{\citenamefont {Izmailov}\ \emph {et~al.}(2021)\citenamefont
  {Izmailov}, \citenamefont {Vikram}, \citenamefont {Hoffman},\ and\
  \citenamefont {Wilson}}]{pmlr-v139-izmailov21a}%
  \BibitemOpen
  \bibfield  {author} {\bibinfo {author} {\bibfnamefont {P.}~\bibnamefont
  {Izmailov}}, \bibinfo {author} {\bibfnamefont {S.}~\bibnamefont {Vikram}},
  \bibinfo {author} {\bibfnamefont {M.~D.}\ \bibnamefont {Hoffman}},\ and\
  \bibinfo {author} {\bibfnamefont {A.~G.~G.}\ \bibnamefont {Wilson}},\ }in\
  \href {https://proceedings.mlr.press/v139/izmailov21a.html} {\emph {\bibinfo
  {booktitle} {Proceedings of the 38th International Conference on Machine
  Learning}}},\ \bibinfo {series} {Proceedings of Machine Learning Research},
  Vol.\ \bibinfo {volume} {139},\ \bibinfo {editor} {edited by\ \bibinfo
  {editor} {\bibfnamefont {M.}~\bibnamefont {Meila}}\ and\ \bibinfo {editor}
  {\bibfnamefont {T.}~\bibnamefont {Zhang}}}\ (\bibinfo  {publisher} {PMLR},\
  \bibinfo {year} {2021})\ pp.\ \bibinfo {pages} {4629--4640}\BibitemShut
  {NoStop}%
\bibitem [{\citenamefont {Phan}\ \emph {et~al.}(2019)\citenamefont {Phan},
  \citenamefont {Pradhan},\ and\ \citenamefont
  {Jankowiak}}]{phan2019composable}%
  \BibitemOpen
  \bibfield  {author} {\bibinfo {author} {\bibfnamefont {D.}~\bibnamefont
  {Phan}}, \bibinfo {author} {\bibfnamefont {N.}~\bibnamefont {Pradhan}},\ and\
  \bibinfo {author} {\bibfnamefont {M.}~\bibnamefont {Jankowiak}},\ }\href@noop
  {} {\bibfield  {journal} {\bibinfo  {journal} {arXiv preprint
  arXiv:1912.11554}\ } (\bibinfo {year} {2019})}\BibitemShut {NoStop}%
\bibitem [{\citenamefont {Bingham}\ \emph {et~al.}(2019)\citenamefont
  {Bingham}, \citenamefont {Chen}, \citenamefont {Jankowiak}, \citenamefont
  {Obermeyer}, \citenamefont {Pradhan}, \citenamefont {Karaletsos},
  \citenamefont {Singh}, \citenamefont {Szerlip}, \citenamefont {Horsfall},\
  and\ \citenamefont {Goodman}}]{bingham2019pyro}%
  \BibitemOpen
  \bibfield  {author} {\bibinfo {author} {\bibfnamefont {E.}~\bibnamefont
  {Bingham}}, \bibinfo {author} {\bibfnamefont {J.~P.}\ \bibnamefont {Chen}},
  \bibinfo {author} {\bibfnamefont {M.}~\bibnamefont {Jankowiak}}, \bibinfo
  {author} {\bibfnamefont {F.}~\bibnamefont {Obermeyer}}, \bibinfo {author}
  {\bibfnamefont {N.}~\bibnamefont {Pradhan}}, \bibinfo {author} {\bibfnamefont
  {T.}~\bibnamefont {Karaletsos}}, \bibinfo {author} {\bibfnamefont
  {R.}~\bibnamefont {Singh}}, \bibinfo {author} {\bibfnamefont {P.~A.}\
  \bibnamefont {Szerlip}}, \bibinfo {author} {\bibfnamefont {P.}~\bibnamefont
  {Horsfall}},\ and\ \bibinfo {author} {\bibfnamefont {N.~D.}\ \bibnamefont
  {Goodman}},\ }\href {http://jmlr.org/papers/v20/18-403.html} {\bibfield
  {journal} {\bibinfo  {journal} {J. Mach. Learn. Res.}\ }\textbf {\bibinfo
  {volume} {20}},\ \bibinfo {pages} {28:1} (\bibinfo {year}
  {2019})}\BibitemShut {NoStop}%
\bibitem [{\citenamefont {Bradbury}\ \emph {et~al.}(2018)\citenamefont
  {Bradbury}, \citenamefont {Frostig}, \citenamefont {Hawkins}, \citenamefont
  {Johnson}, \citenamefont {Leary}, \citenamefont {Maclaurin}, \citenamefont
  {Necula}, \citenamefont {Paszke}, \citenamefont {Vander{P}las}, \citenamefont
  {Wanderman-{M}ilne},\ and\ \citenamefont {Zhang}}]{jax2018github}%
  \BibitemOpen
  \bibfield  {author} {\bibinfo {author} {\bibfnamefont {J.}~\bibnamefont
  {Bradbury}}, \bibinfo {author} {\bibfnamefont {R.}~\bibnamefont {Frostig}},
  \bibinfo {author} {\bibfnamefont {P.}~\bibnamefont {Hawkins}}, \bibinfo
  {author} {\bibfnamefont {M.~J.}\ \bibnamefont {Johnson}}, \bibinfo {author}
  {\bibfnamefont {C.}~\bibnamefont {Leary}}, \bibinfo {author} {\bibfnamefont
  {D.}~\bibnamefont {Maclaurin}}, \bibinfo {author} {\bibfnamefont
  {G.}~\bibnamefont {Necula}}, \bibinfo {author} {\bibfnamefont
  {A.}~\bibnamefont {Paszke}}, \bibinfo {author} {\bibfnamefont
  {J.}~\bibnamefont {Vander{P}las}}, \bibinfo {author} {\bibfnamefont
  {S.}~\bibnamefont {Wanderman-{M}ilne}},\ and\ \bibinfo {author}
  {\bibfnamefont {Q.}~\bibnamefont {Zhang}},\ }\href
  {http://github.com/google/jax} {\bibinfo {title} {{JAX}: composable
  transformations of {P}ython+{N}um{P}y programs}} (\bibinfo {year}
  {2018})\BibitemShut {NoStop}%
\bibitem [{\citenamefont {Salvatier}\ \emph {et~al.}(2016)\citenamefont
  {Salvatier}, \citenamefont {Wiecki},\ and\ \citenamefont
  {Fonnesbeck}}]{Salvatier2016}%
  \BibitemOpen
  \bibfield  {author} {\bibinfo {author} {\bibfnamefont {J.}~\bibnamefont
  {Salvatier}}, \bibinfo {author} {\bibfnamefont {T.~V.}\ \bibnamefont
  {Wiecki}},\ and\ \bibinfo {author} {\bibfnamefont {C.}~\bibnamefont
  {Fonnesbeck}},\ }\href {https://doi.org/10.7717/peerj-cs.55} {\bibfield
  {journal} {\bibinfo  {journal} {{PeerJ} Computer Science}\ }\textbf {\bibinfo
  {volume} {2}},\ \bibinfo {pages} {e55} (\bibinfo {year} {2016})}\BibitemShut
  {NoStop}%
\bibitem [{\citenamefont {Blondel}\ \emph {et~al.}(2021)\citenamefont
  {Blondel}, \citenamefont {Berthet}, \citenamefont {Cuturi}, \citenamefont
  {Frostig}, \citenamefont {Hoyer}, \citenamefont {Llinares-L{\'o}pez},
  \citenamefont {Pedregosa},\ and\ \citenamefont
  {Vert}}]{jaxopt_implicit_diff}%
  \BibitemOpen
  \bibfield  {author} {\bibinfo {author} {\bibfnamefont {M.}~\bibnamefont
  {Blondel}}, \bibinfo {author} {\bibfnamefont {Q.}~\bibnamefont {Berthet}},
  \bibinfo {author} {\bibfnamefont {M.}~\bibnamefont {Cuturi}}, \bibinfo
  {author} {\bibfnamefont {R.}~\bibnamefont {Frostig}}, \bibinfo {author}
  {\bibfnamefont {S.}~\bibnamefont {Hoyer}}, \bibinfo {author} {\bibfnamefont
  {F.}~\bibnamefont {Llinares-L{\'o}pez}}, \bibinfo {author} {\bibfnamefont
  {F.}~\bibnamefont {Pedregosa}},\ and\ \bibinfo {author} {\bibfnamefont
  {J.-P.}\ \bibnamefont {Vert}},\ }\href@noop {} {\bibfield  {journal}
  {\bibinfo  {journal} {arXiv preprint arXiv:2105.15183}\ } (\bibinfo {year}
  {2021})}\BibitemShut {NoStop}%
\bibitem [{\citenamefont {Cho}\ and\ \citenamefont
  {Saul}(2009)}]{NIPS2009_5751ec3e}%
  \BibitemOpen
  \bibfield  {author} {\bibinfo {author} {\bibfnamefont {Y.}~\bibnamefont
  {Cho}}\ and\ \bibinfo {author} {\bibfnamefont {L.}~\bibnamefont {Saul}},\
  }in\ \href
  {https://proceedings.neurips.cc/paper_files/paper/2009/file/5751ec3e9a4feab575962e78e006250d-Paper.pdf}
  {\emph {\bibinfo {booktitle} {Advances in Neural Information Processing
  Systems}}},\ Vol.~\bibinfo {volume} {22},\ \bibinfo {editor} {edited by\
  \bibinfo {editor} {\bibfnamefont {Y.}~\bibnamefont {Bengio}}, \bibinfo
  {editor} {\bibfnamefont {D.}~\bibnamefont {Schuurmans}}, \bibinfo {editor}
  {\bibfnamefont {J.}~\bibnamefont {Lafferty}}, \bibinfo {editor}
  {\bibfnamefont {C.}~\bibnamefont {Williams}},\ and\ \bibinfo {editor}
  {\bibfnamefont {A.}~\bibnamefont {Culotta}}}\ (\bibinfo  {publisher} {Curran
  Associates, Inc.},\ \bibinfo {year} {2009})\BibitemShut {NoStop}%
\end{thebibliography}
\end{document}